\documentclass[11pt, a4paper, twocolumn, confidential, goog]{google}

\usepackage[authoryear, sort&compress, round]{natbib}
\usepackage{textcomp, xspace}
\usepackage{pbox}
\usepackage{booktabs}
\usepackage{graphicx}
\usepackage{rotating}
\usepackage{tcolorbox}
\usepackage[capitalise]{cleveref}
\usepackage{multirow}

\newcommand{\framework}{\textsc{TRACE}\xspace}

\uselogo{} 

\title{Do LLMs Really Need 10+ Thoughts for ``Find the Time 1000 Days Later''? Towards Structural Understanding of LLM Overthinking}
\correspondingauthor{xlfzhang@\{umich.edu, google.com\}.\\ $^{+}$Work done during XFZ's internship at GDM.}

\reportnumber{} % Leave blank if n/a

\author[1,+]{Xinliang Frederick Zhang}
\author[2,*]{Anhad Mohananey}
\author[2,*]{Alexandra Chronopoulou}
\author[2]{Pinelopi Papalampidi}
\author[2]{\\Somit Gupta}
\author[2]{Tsendsuren Munkhdalai}
\author[1]{Lu Wang}
\author[2]{Shyam Upadhyay}

\affil[*]{Equal contributions}
\affil[1]{University of Michigan}
\affil[2]{\thepa{}{}}

\begin{abstract}
Models employing long chain-of-thought (CoT) reasoning have shown superior performance on complex reasoning tasks. Yet, this capability introduces a critical and often overlooked inefficiency---\textbf{overthinking}---models often engage in unnecessarily extensive reasoning even for \textit{simple} queries, incurring significant computations without accuracy improvements. 
While prior work has explored solutions to mitigate overthinking, a fundamental gap remains in our understanding of its underlying causes. Most existing analyses are limited to superficial, profiling-based observations, failing to delve into \textit{LLMs' inner workings}. 
This study introduces a systematic, fine-grained analyzer of LLMs' thought process to bridge the gap, \framework. We first benchmark the overthinking issue, confirming that long-thinking models are five to twenty times slower on simple tasks with no substantial gains. We then use \framework to first decompose the thought process into minimally complete \textit{sub-thoughts}. Next, by inferring discourse relationships among sub-thoughts, we construct granular thought progression graphs and subsequently identify common thinking patterns for topically similar queries.
Our analysis reveals two major patterns for open-weight thinking models---\textit{Explorer} and \textit{Late Landing}.  This finding provides evidence that \textit{over-verification} and \textit{over-exploration} are the primary drivers of overthinking in LLMs. Grounded in thought structures, we propose a utility-based definition of overthinking, which moves beyond length-based metrics. This revised definition offers a more insightful understanding of LLMs' thought progression, as well as practical guidelines for principled overthinking management.

\end{abstract}

\begin{document}

\maketitle

\section{Introduction}

There is a recent interest in thinking models that generate long-chain-of-thought (CoT) responses without manual prompts~\citep{DBLP:journals/corr/abs-2501-09686, DBLP:journals/corr/abs-2502-17419}. However, current state-of-the-art thinking models~\citep{gemini2.5, gpt-o1, qwq, r1} often engage in excessive thinking even for simple inquiries, leading to computational inefficiency and unnecessary reasoning~\citep{overthink_survey}. While various methods have been proposed to address this, such as using RL training with length-based penalties~\citep{DBLP:journals/corr/abs-2502-04463, DBLP:journals/corr/abs-2503-04697}, compressing reasoning~\citep{DBLP:journals/corr/abs-2405-14838}, or specifying response length in prompts~\citep{han-etal-2025-token, DBLP:journals/corr/abs-2502-18600}, a fundamental knowledge gap remains in our understanding of the underlying causes behind LLM overthinking. Worse still, most of existing analyses are limited to generic profiling and only superficially examine how thinking length impacts model performance~\citep{DBLP:journals/corr/abs-2412-21187, qwen3, DBLP:journals/corr/abs-2505-13379}. In fact, a deeper analysis of LLMs' internal mechanisms---specifically, their thought structures and the minimum thinking effort sufficient for effective performance---is conspicuously absent.

To this end, we first systematically benchmark overthinking of LLMs, via head-to-head comparison between \textit{non-thinking} and \textit{thinking} modes.\footnote{\textit{Non-thinking} mode provides near-instant responses; \textit{thinking} mode takes longer time to reason before answering.} Our results show that for \textit{simple queries} like ``7+2'' or ``date of 1000 days after today'', thinking models naively spend \textit{five to twenty times} longer at inference than non-thinking counterparts, with little or even no improvement in performance. This finding is inline with the literature~\citep{DBLP:journals/corr/abs-2412-21187} but we uncover the finding at scale, across 14 thinking LLMs~\citep{qwen3, r1} and 6 data domains~\citep[i.a.,][]{miao-etal-2020-diverse, gsm8k, tan-etal-2023-towards, simpleqa}.

Next, we dig deeper into the LLMs' thought process to understand the ``why'' and ``how'' of overthinking via our proposed analyzer (\Cref{fig:framework_overview}), \framework, a \textbf{T}hought-process \textbf{R}econstruction and \textbf{A}utomated \textbf{C}lustering \textbf{E}ngine. Concretely, we decompose the thought process into smaller, independent, minimally complete sub-thoughts, and then evaluate the accuracy and helpfulness of each sub-thought. Next, we leverage LLM-as-rater~\citep{DBLP:journals/ftir/RobertsonZ09, gemini2.5} to infer discourse relationships among sub-thoughts. The outputs allow us to construct thought progression graphs for individual queries, and furthermore, identify general thought patterns for \textit{topically similar queries}. Our analysis reveals two primary thought progression patterns: the \textbf{Explorer} and \textbf{Late Landing}. The \textit{Explorer} pattern distributes the probability of correctness across multiple potential answers. It may find the correct answer early, but it continues to explore alternatives. Conversely, the \textit{Late Landing} pattern follows a convergent reasoning path where the probability of correctness is highly concentrated on the final, distinct answer. 
This further suggests \textbf{over-verification} and \textbf{over-exploration} as key drivers of overthinking. 
Based on the thought structures, we propose a utility-based definition of overthinking---\textit{continuation of the thought process after the marginal return drops below a predefined threshold ($\epsilon$)}.
This revised definition moves beyond mere length-based metrics, and transforms overthinking from a vague notion into a quantifiable and actionable metric. Concretely, it not only pinpoints the onset of overthinking within the thought process, but also informs practical heuristics for real-time overthinking management using thought progression-induced characteristics.
Overall, \framework deepens the understanding of LLM overthinking more structurally.

\section{Related Work}

\paragraph{Thinking Models.}
Large Language Models (LLMs), built on the transformer architecture~\citep{DBLP:conf/nips/VaswaniSPUJGKP17}, represent a transformative step toward Artificial General Intelligence~\citep[AGI;][]{DBLP:journals/corr/abs-2501-03151}.
Recently, the rise of thinking models\footnote{They are also commonly referred to as \textit{reasoning models}.}~\citep{DBLP:journals/corr/abs-2501-09686, gpt-o1, r1} has significantly enhanced  performance with their \textbf{system-2 thinking} competence~\citep{system2, DBLP:journals/corr/abs-2502-17419}, 
Their remarkable performance on complex tasks, such as challenging mathematical contests~\citep{DBLP:conf/nips/HendrycksBKABTS21, DBLP:conf/iclr/LightmanKBEBLLS24, LuongLockhart2025}, competitive programming tasks~\citep{DBLP:journals/corr/abs-2404-10952, DBLP:conf/iclr/JainHGLYZWSSS25} and personalization tasks~\citep{DBLP:journals/corr/abs-2507-04607}, is largely attributed to their capacity for generating extensive reasoning traces~\citep{ DBLP:journals/corr/abs-2502-17419, DBLP:journals/corr/abs-2503-09567}. This capability is built upon the foundational technique of chain-of-thought (CoT) prompting~\citep{DBLP:conf/nips/KojimaGRMI22, DBLP:conf/nips/Wei0SBIXCLZ22}, which encourages models to articulate intermediate steps before answering. 
By expending more tokens to explore various problem-solving paths, engaging in self-correction, and performing verification, these models achieve state-of-the-art (SOTA) results in challenging domains~\citep{DBLP:journals/corr/abs-2408-03314, DBLP:journals/corr/abs-2501-19393, DBLP:journals/corr/abs-2507-12759}.

\paragraph{LLM Overthinking.}

While the ``longer is better'' paradigm~\citep{jin-etal-2024-impact, DBLP:journals/corr/abs-2503-18069}, a test-time scaling approach~\citep{DBLP:journals/corr/abs-2408-03314} that empowers system-2 thinking, has boosted model accuracy on complex tasks, recent research reveals a critical \textit{inefficiency} issue in thinking LLMs: \textbf{overthinking}~\citep{DBLP:journals/corr/abs-2412-21187, overthink_survey}. 
This phenomenon is characterized by models expending \textit{unnecessarily excessive compute} on queries, especially simple ones and those with an evident answer, for marginal improvements or even negative returns. At its worst, overthinking causes performance degradation~\citep{DBLP:journals/corr/abs-2410-21333}, where a model may abandon a correct initial/intermediate answer for  a wrong one eventually. 
Although the benefits of extended reasoning on difficult problems are clear, the trade-offs across the full spectrum of task difficulties remain \textit{underexplored}, particularly for \textit{simple queries},\footnote{Though ``simple'' is cognitively hard to define, we treat queries solvable by bright middle school students as \textit{simple}.} where models' verbosity makes them more susceptible to overthinking~\citep{DBLP:journals/corr/abs-2412-21187, pu_terminator}.
To this end, we are the first to comprehensively benchmark LLM overthinking tendencies on \textit{simple queries}. Our evaluation spans both \textit{horizontally}---across a diverse range of tasks and domains---and \textit{vertically}, by varying the intrinsic difficulty of the problems.

\begin{figure*}[t]
    \centering
    \includegraphics[width=1.0\textwidth]{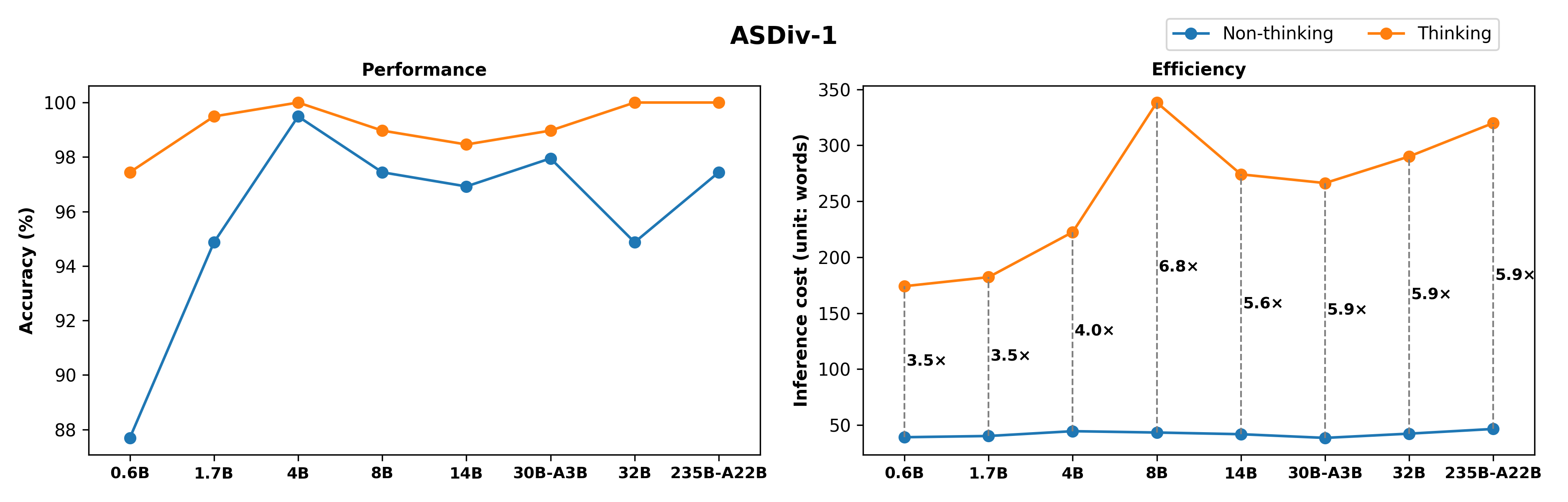}
    \caption{Performance and inference-time efficiency trends of \texttt{Qwen3} models at different scales on simple math reasoning. We find that additional thinking becomes ineffective once the model scale is above  the threshold of 4B. Plots for other tasks such as temporal and logical reasoning, as well as knowledge recall, are in \Cref{appx:horizontal}.}
    \label{fig:perf_eff_trend_asdiv}
\end{figure*}

Meanwhile, despite a variety of strategies have been proposed to address the overthinking issue~\citep{DBLP:journals/corr/abs-2502-04463, DBLP:journals/corr/abs-2503-04697, han-etal-2025-token, DBLP:journals/corr/abs-2502-18600}, a deep exploration of the underlying drivers of overthinking remains largely absent in the literature. Existing analyses are often superficial, focusing on coarse, length-based metrics. 
For instance, some studies measure overthinking in a monolithic way, defining it based on the token distance to the first correct answer and treating subsequent tokens as overthinking~\citep{DBLP:journals/corr/abs-2412-21187}; Others evaluate whether the shortest response among over-sampled generations is the most cost-effective~\citep{pu_terminator, DBLP:journals/corr/abs-2505-17813}; More recent work, inspired by thinking mode fusion~\citep{qwen3}, quantifies overthinking as the generation of excessive thinking tokens on simple queries without yielding performance improvements over the non-thinking mode~\citep{aggarwal2025optimalthinkingbench}.
Regardless of implementation variants, existing approaches remain confined to length-based profiling, without examining the underlying \textit{thought evolution}.

In contrast, our work provides a fine-grained analysis at the sub-thought level. Specifically, we investigate the \textbf{internal dynamics of the thought process}, examining how different answers are proposed throughout the thought process and, more importantly, how later sub-thoughts relate to earlier ones. To the best of our knowledge, we are the first to conduct such a structural analysis to understand \textit{why and how} overthinking occurs in LLMs, from its inner workings perspective.

\section{Benchmark LLM Overthinking}
\label{sec:benchmark}
To investigate thinking LLMs' inefficiency, we systematically benchmark the phenomenon of overthinking on \textbf{simple queries}~\citep{DBLP:journals/corr/abs-2506-06941}, where LLMs expend computational resources on thought tokens that yield marginal or no performance gains. While prior work has identified this issue, its scope has been largely confined to reasoning tasks related to STEM~\citep{aggarwal2025optimalthinkingbench}, especially mathematical reasoning~\citep{DBLP:journals/corr/abs-2412-21187, DBLP:journals/corr/abs-2504-20708}. Our research provides a more comprehensive analysis by evaluating overthinking tendencies both \textit{horizontally} (across various domains) and \textit{vertically} (varying task difficulties). 
 This dual-axis approach provides the first large-scale characterization of the trade-off between reasoning length and model performance, clarifying \textit{when and why thinking is beneficial}.

\paragraph{Defining Overthinking.}
Inline with the Following~\citep{DBLP:journals/corr/abs-2412-21187, overthink_survey, aggarwal2025optimalthinkingbench}, we introduce an \textit{initial} definition of overthinking based on the \textbf{generation length}, allowing us to quantify the overthinking extent.

\begin{tcolorbox}[colback=white, colframe=black, fonttitle=\bfseries, title=Length-based Overthinking Definition]
\textit{Overthinking is the generation of additional thought tokens that do not contribute to a performance gain.} 
\end{tcolorbox}

Formally, we measure this as the difference in the total generation length between the thinking and non-thinking modes of \textit{the same evaluated model} for the subset of questions that are \textit{already answered correctly} in the non-thinking mode.

\subsection{Horizontal Analysis}
\label{sec:horizontal}

Our \textit{horizontal} analysis surveys a wide range of domains and tasks---ranging from simple reasoning (e.g., grade-level math problems) to fact-retrieval tasks with minimal reasoning requirements (e.g., machine reading comprehension). The goal is to \textit{establish a baseline by identifying settings where extended reasoning is unlikely to offer benefits}.

\paragraph{Experimental Setup.}
\label{experimental_setup}
To ensure our findings are robust, we benchmark a suite of 14 models trained via reinforcement learning (RL) or supervised fine-tuning via distillation. Specifically, we sample responses from a wide range of models, including the \textit{Qwen3 model family} \citep[from 0.6B to 235B parameters;][]{qwen3} and \textit{Deepseek-R1 distilled models}~\citep{r1} based on Qwen2.5~\citep{qwen2.5} and Llama-3~\citep{llama3} backbones (from 1.5B to 70B parameters). For the Qwen3 models, we generate responses in both \textit{thinking} and \textit{non-thinking} modes to create a controlled comparison. See the full list of backbone models at~\Cref{app:backbones}.

Our experiments focus on \textbf{simple queries} for which LLMs are more likely to exhibit overthinking~\citep{DBLP:journals/corr/abs-2506-06941}. These queries span two main domains, sourced from six datasets (datasets statistics are summarized in \cref{tab:evaluation_datasets}).
\begin{itemize}[nosep]
    \item \textbf{Simple Reasoning}: Tasks requiring basic logical steps, such as grade-school math word problems~\citep{miao-etal-2020-diverse}, date arithmetic~\citep{tan-etal-2023-towards}, and logic grid puzzles~\citep{DBLP:journals/corr/abs-2502-01100}.
    \item \textbf{Knowledge Recall}: Tasks that necessitate recalling information with minimal reasoning: discerning unanswerable questions~\citep{rajpurkar-etal-2018-know}, locating facts from long contexts~\citep{kamradt2023needle}, and producing factually correct responses~\citep{simpleqa}.
\end{itemize}

To ensure deterministic and reproducible outputs, we employ a \textbf{greedy decoding} strategy with temperature set to 0 and top-p set to 0. The maximum token generation limit was set to 8$k$ tokens for thinking LLMs (or 16$k$ for more complex temporal reasoning tasks) to allow for exhaustive exploration of the thought process, and 2$k$ for the non-thinking mode.

\paragraph{Preliminary Results.}
Benchmark performances are shown in~\Cref{fig:perf_eff_trend_asdiv} and \Cref{tab:benchmark} (full results).
For simple reasoning tasks, the performance gains from enabling a model's thinking capabilities are most significant for models with fewer than 4--8 billion parameters (\Cref{fig:perf_eff_trend_asdiv}). 
Beyond this size threshold, the performance gap between thinking and non-thinking modes diminishes, mostly approaching zero. This demonstrates that additional thinking ceases to be effective and thus renders overthinking once the model size is beyond a threshold. Conversely, for knowledge recall tasks that involve a minimal reasoning workload, enabling thinking provides negligible benefits, irrespective of task difficulty (\Cref{fig:perf_eff_trend_squad} to \ref{fig:perf_eff_trend_simpleqa}). 

Although these analyses focus on simple queries, the results already suggest that the necessity for long-form thinking is not dictated by the \textit{overall task complexity}.\footnote{The literature commonly uses task complexity as a signal to decide the expected reasoning efforts~\citep{pu_terminator}.} Rather, it depends on the \textit{expected reasoning workload},\footnote{Workload refers to amount of reasoning efforts, e.g., intermediate stepwise inference, required to reach a correct answer.} which is further confirmed in \cref{sec:vertical} by varying reasoning efforts.

\begin{table}[t]
\centering
\resizebox{0.45\textwidth}{!}{
\begin{tabular}{@{}lrr@{}}
\toprule
\textbf{Difficulty Level} & \textbf{Performance} & \textbf{Efficiency} \\
\midrule
ASDiv-1         & 97.44/100.00  & 46.5/320.0 (255.8)      \\
ASDiv-2         & 92.94/95.88   & 54.4/348.5 (274.5)      \\
ASDiv-3         & 89.00/96.75 & 59.7/429.8 (349.7)   \\
ASDiv-4         & 83.39/93.36   & 74.8/553.1 (450.8)      \\
ASDiv-5         & 76.03/90.41   & 127.3/701.5 (553.9)     \\
GSM8k           & 74.75/91.50 & 118.1/1$,$021.7 (889.7) \\
\bottomrule
\end{tabular}
}
\caption{Accuracy performance and inference efficiency (unit: words) of \texttt{Qwen3-235B-A22B} for vertical analysis (math reasoning). Values are reported as non-thinking/thinking, with thought lengths within parentheses. Full results see~\Cref{tab:benchmark_math}.} 
\vspace{-4mm}
\label{tab:qwen235b_performance_math}
\end{table}

\subsection{Vertical Analysis}
\label{sec:vertical}
Our \textit{vertical} analysis narrows the focus to two domains (mathematical and temporal reasoning) where task difficulty can be systematically controlled. The aim here is to \textit{identify the range of which extended reasoning might have become beneficial, and to distinguish it from settings where it offers little to no advantage}.

\paragraph{Experimental Setup.}
We adopt the same evaluation setup as in~\Cref{sec:horizontal}, testing 14 models with greedy decoding.  
For data domains, we focus on mathematical and temporal reasoning. Specifically, for mathematical reasoning, we use ASDiv~\citep{miao-etal-2020-diverse} at grade 1–5 levels, along with GSM8k~\citep{gsm8k}, which is considered more challenging than grade-5 ASDiv and solvable by bright middle-school students. 
For temporal reasoning, the original data provided in ~\citet{tan-etal-2023-towards} corresponds to level-1 difficulty. We then progressively increase query complexity using the procedure described in~\Cref{appx:temp_data_creation}, resulting in five difficulty levels. Sample examples for each level are shown in~\Cref{tab:temporal_sample}.

\paragraph{Preliminary Results.}
Our vertical study shows that \textit{extended reasoning only pays off within narrow boundaries}---outside of which it quickly devolves into overthinking.

\begin{table}[t]
\centering
\resizebox{0.45\textwidth}{!}{
\begin{tabular}{@{}lrr@{}}
\toprule
\textbf{Difficulty Level} & \textbf{Performance} & \textbf{Efficiency} \\
\midrule
Temporal-L1 & 95.87/99.41 & 71.2/261.0 (162.0)    \\
Temporal-L2 & 95.52/97.61 & 114.2/679.7 (534.7)   \\
Temporal-L3 & 32.24/52.54 & 309.2/2$,$485.6 (2$,$255.8) \\
Temporal-L4 & 14.63/47.76 & 334.6/2$,$843.8 (2$,$625.4) \\
Temporal-L5 & 8.66/45.37  & 323.9/3$,$336.0 (3$,$078.9) \\
\bottomrule
\end{tabular}
}
\caption{Accuracy performance and inference efficiency (unit: words) of \texttt{Qwen3-235B-A22B} for vertical analysis (temporal reasoning). Full results see~\Cref{tab:benchmark_date}.} 
\vspace{-4mm}
\label{tab:qwen235b_performance_date}
\end{table}

For \textit{mathematical reasoning} (\Cref{tab:qwen235b_performance_math}), thinking improves accuracy as tasks get harder and maintains its performance above $90\%$: from negligible gains at ASDiv-1/2 to $15$ points at GSM8k. Yet this comes at a steep cost: solving GSM8k requires over $10\times$ more thought tokens, and still $80\%$ of that extra compute produces no measurable gain. In other words, \textit{even where thinking helps, the majority (e.g., $80\%$) of computation is wasted}.
For \textit{temporal reasoning} (\Cref{tab:qwen235b_performance_date}), at L1–L2, where queries involve manageable duration representations, non-thinking models already achieve near-perfect accuracy, so extra reasoning adds little help. Beyond L3, where tasks demand day-level counting over hundreds or thousands of days, robust handling of leap years, and potential confusion with Julian calendar system, thinking performance collapses despite \textit{huge reasoning workload}. While enabling thinking does help, boosting accuracy up to roughly $50\%$, the improvement quickly saturates. This reflects a ceiling imposed by the model’s \textbf{representational capacity}---its ability to internally encode and manipulate the structural rules required for the task. Past this capacity, \textit{more thinking becomes pure overthinking and additional reasoning cannot bridge the gap}.

In summary, thinking is valuable only in a \textit{narrow middle ground}: trivial tasks waste it, and tasks beyond the model’s representational capacity nullify it. While math domain highlights the first case, non-math domains such as temporal reasoning exhibits the second, revealing the true landscape of overthinking that math domain-intense studies alone cannot capture~\citep[i.a.,][]{DBLP:journals/corr/abs-2412-21187}.

\begin{figure*}[t]
    \centering
    \includegraphics[width=1.0\textwidth]{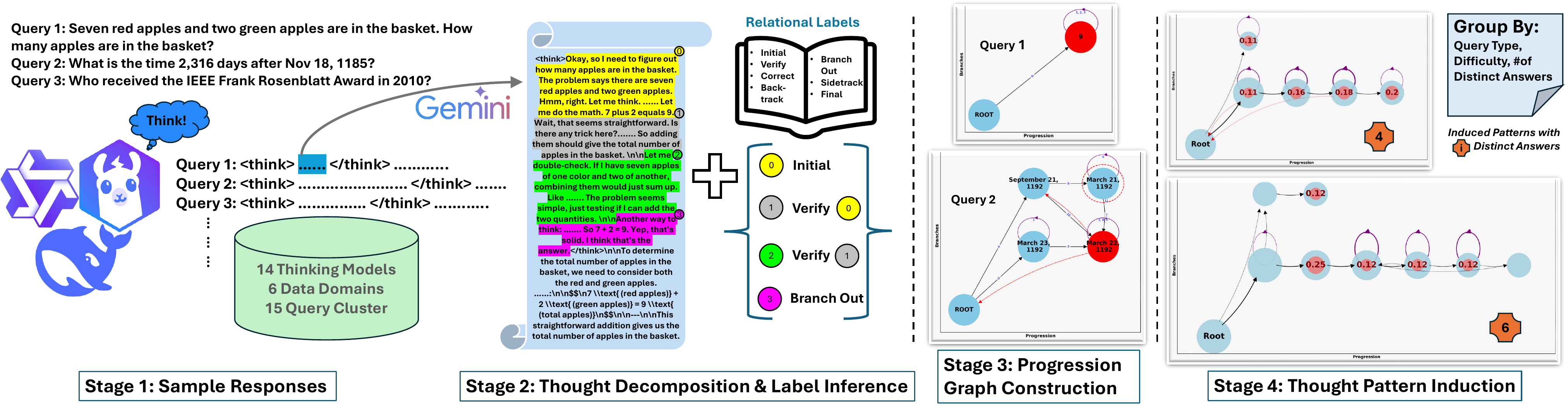}
    \vspace{-4mm}
    \caption{Overview of our proposed analyzer (\framework) to study the inner workings of an LLM's thought process. It contains four main stages (detailed in~\Cref{sec:analysis_framework}): \textit{Response Sampling},  \textit{Thought Decomposition \& Label Inference}, \textit{ Progression Graph Construction}, and \textit{Thought Pattern Induction}.}
    \label{fig:framework_overview}
\end{figure*}

\section{Analysis Framework of Inner Workings}
\label{sec:analysis_framework}

Figure~\ref{fig:framework_overview} introduces the overview of our proposed framework,~\framework, to study the inner workings of an LLM's thought process. \framework, a \textbf{T}hought-process \textbf{R}econstruction and \textbf{A}utomated \textbf{C}lustering \textbf{E}ngine, contains four main stages: \textit{(1) Response Sampling}, where we generate responses in auto-regressive manner from various models; \textit{(2) Thought Decomposition \& Label Inference}, where we break down the reasoning and infer the relationships between sub-thoughts; \textit{(3) Progression Graph Construction}, where we represent the reasoning flow in a graph structure; and \textit{(4) Thought Pattern Induction}, where we aggregate individual graphs to discover generalized thought patterns.

For stage 1, we follow the same setting as in~\Cref{sec:horizontal} to sample responses from 4 large thinking models: \texttt{Qwen3-30B-A3B}, \texttt{Qwen3-32B}, \texttt{R1-Distill-Llama-70B} and \texttt{Qwen3-235B-A22B}.\footnote{We focus on large models ($>20B$) because smaller ones are unable to produce meaningful outputs for complex tasks. \texttt{R1-Distill-Owen-32B} is not suited because it lacks the long-thinking ability required to explore thinking structure.}

\paragraph{Stage 2: Thought Decomposition \& Label Inference.}
Once responses are collected, we utilize \texttt{gemini-2.5-pro}~\citep{gemini2.5} to systematically decompose each thought process into sequential \textbf{sub-thoughts} and infer the functional relationship between them. Specifically, we define a sub-thought as a text segment satisfying three strict criteria---self-contained, complete and answer-bearing (detailed in \Cref{appx:sub_thought})---and leverage sub-thought transition markers~\citep{DBLP:journals/corr/abs-2504-20708} to help decide the sub-thought boundaries. 

\textbf{Label Inference:} While prior literature has informally described model behaviors~\citep{DBLP:journals/corr/abs-2503-01307}, these descriptions often lack formal definitions. We establish a comprehensive and formally defined set of thought-to-thought relational labels to systematically categorize the transitions in the thought process. We have unary labels including \textit{Initial} (first attempt, planning) and \textit{Final} (final answer delivery); For the intermediate sub-thoughts, we devise binary labels: \textit{Verification} (confirm the preceding thought's correctness), \textit{Correction} (rectify the preceding thought), \textit{Backtrack} (reverts to an earlier path) and \textit{Branching Out} (explore a new approach). \textit{Sidetrack} is a unique label for rambling---a digressive, seemingly unrelated tangent that often adds no value to the thinking. Detailed definitions of each label are included in~\Cref{appx:inference_label}.

We have provided the complete instruction in~\Cref{fig:prompt_p1} and~\ref{fig:prompt_p2} for reproducibility purpose. In addition to~\Cref{fig:framework_overview} (stage 2), we provide another parsed output in~\Cref{fig:gsm8k_parsed_example_part1} to~\ref{fig:gsm8k_parsed_example_part2}.

\begin{figure}[t]
    \centering
    \includegraphics[width=0.5\textwidth]{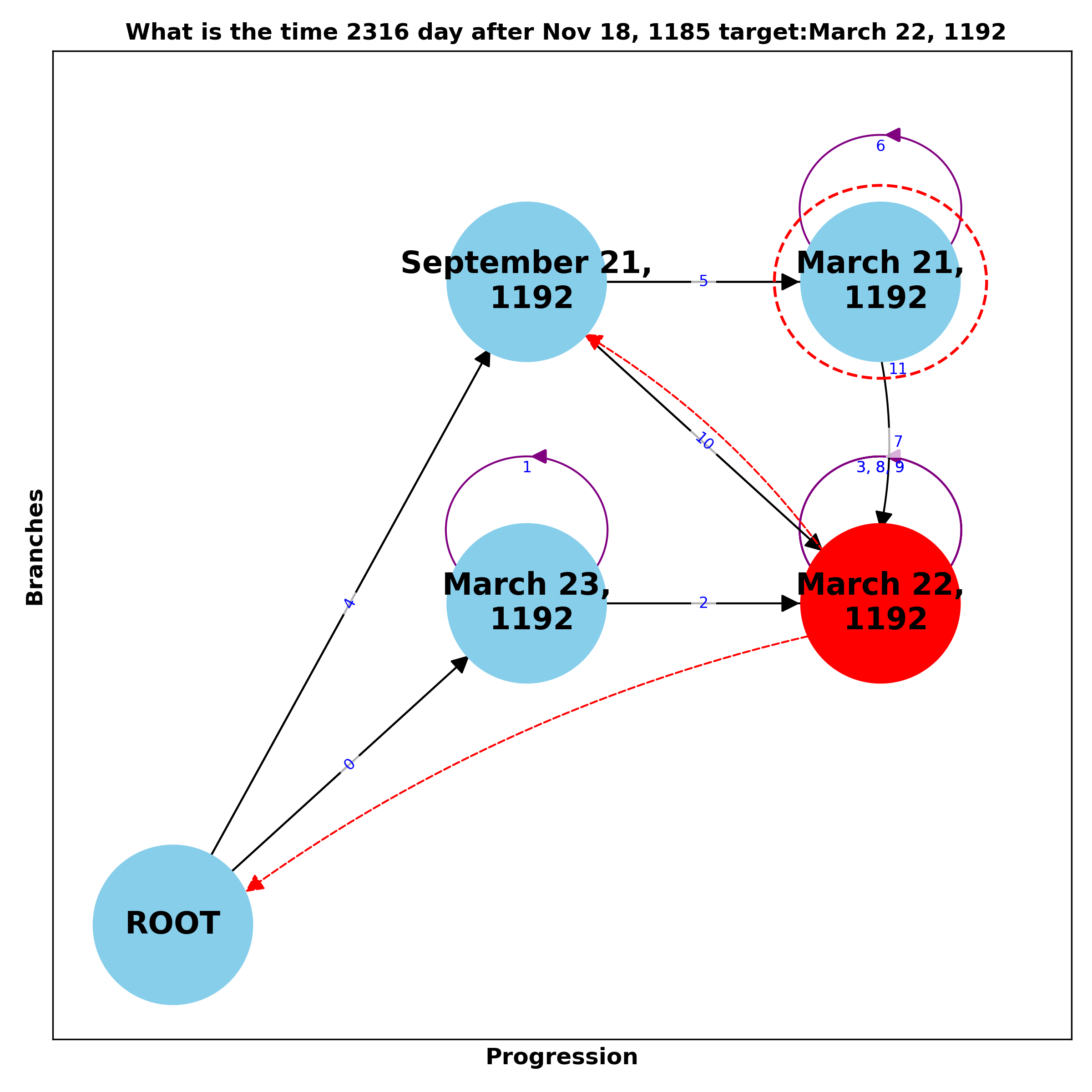}
    \caption{Individual thought progression graph of \texttt{Qwen3-235B-A22B} model on a sampled date arithmetic (temporal-L3) query. \textcolor{red}{Red bubble} denotes the ground-truth answer, while the \textcolor{red}{red dashed circle} denotes the final delivered answer.}
    \label{fig:progression_3}
\end{figure}

\paragraph{Stage 3: Progression Graph Construction.}
To formally represent the reasoning trajectory, we construct a \textbf{thought progression graph} (e.g., \Cref{fig:progression_3}) for each individual thinking trace. In this graph, each node corresponds to a distinct answer proposed by the LLM during its thought process. A directed edge connects two nodes, representing the inferred relational label (e.g., \textit{verification}) between the corresponding sub-thoughts. 
To facilitate systematic analysis, we project these graphs onto a 2D coordinate system, with a designated root node at (0,0) representing the query. Unary labels are encoded as follows: \textit{initial} is marked as a directed edge from the root to the first explored answer at (1,1), and the \textit{final} node (if applicable) is highlighted with a red dashed circle. For visualization, \textit{branching out} is converted into either \textit{verification} or \textit{correction}, depending on whether the newly explored method produces the same or a different answer from the preceding node.

Regarding the layout, each new distinct answer is placed one step to the right ($x$-value $+$ 1). When \textit{backtracking} occurs and leads to exploration of a new answer, the new node is instead placed on an elevated row ($y$-value $+$ 1). This design choice serves two purposes: (1) it visually distinguishes the new, alternative path from prior linear progression, preventing overlaps, and (2) it emphasizes the \textbf{search depth} of the thinking process, with higher rows representing deeper layers of exploration prompted by prior \textit{abandoned} path.

\paragraph{Stage 4: Thought Pattern Induction (Clustering).}
The final stage of \framework aims to induce generalized and interpretable reasoning patterns (e.g., \Cref{fig:explorer_pattern_group5}) for a collection of topically similar individual progression graphs~\citep{chambers-jurafsky-2008-unsupervised, jin-etal-2022-event, zhang-etal-2024-narrative}.\footnote{Following the literature, we group responses by similar prompts, not responses to the identical prompt.} This is accomplished via a three-step process:
\begin{enumerate}[nosep]
    \item \textbf{Group-Based Aggregation}: Graphs are first grouped by shared characteristics, i.e., \textit{query type}, \textit{query difficulty}, and \textit{the number of distinct answers proposed in the thought process}. Within each group, we perform a trivial alignment of the graphs using their coordinate representations and aggregate them. During aggregation, node weights are determined by their presence count, and edge weights are determined by their frequency.
    \item \textbf{Thresholding}: We apply a heuristic threshold (e.g., 0.3) to the aggregated graph to prune infrequent nodes and edges. This filtering step removes noise and highlights the dominant reasoning paths.
    \item \textbf{Ground-Truth Analysis}: Finally, we link the resulting patterns with the ground-truth answers. This allows us to analyze the distribution of ground-truth answers across different reasoning structures (e.g., \textit{Late Landing} vs. \textit{Explorer};~\Cref{sec:thought_patterns}) and understand how different models arrive at their conclusions.
\end{enumerate}

\section{Findings}
\label{sec:findings}
By applying our \framework, we can systematically break down a model's thought process into a quantifiable progression. \Cref{fig:progression_3} presents one individual progression graph on a sample response.\footnote{Additional graphs are presented in \Cref{appx:individual}.}
We then conduct a comprehensive analysis based on the \textit{aggregation instead of individuals}, as the latter may display noisy or mixed behaviors.
More importantly, our analysis reveals two predominant patterns of thought progression, when the model generates \textit{at least 3 intermediate answers while thinking}.\footnote{For cases with $\le2$ distinct answers, the progression is typically linear without branches. See \Cref{fig:explorer_pattern_group2} and \ref{fig:late_lander_pattern_group1_gsm8k}.} These patterns, which we term \textbf{Explorer} and \textbf{Late Landing}, reflect \textit{intrinsic dynamics of models}---rather than properties of individual prompts---and represent different manifestations of their reasoning and tendencies to overthink.

\paragraph{Novel Definition of Overthinking}
Based on the thought patterns identified in~\Cref{sec:thought_patterns} and the associated utility tracing (detailed in~\Cref{sec:utility_tracing}), we provide a refined definition for overthinking based on the decomposed \textit{thought structure} (\Cref{fig:progression_3}).
 
\begin{tcolorbox}[colback=white, colframe=black, fonttitle=\bfseries, title=Structure-based Overthinking Definition]
\textit{Overthinking is the continuation of thought beyond the point where the marginal \textbf{return} ($\Delta$Performance / $\Delta$Thought) drops below a predefined threshold $\epsilon$.} 
\end{tcolorbox}

This redefinition formally identifies overthinking as reasoning that continues after the onset of \textit{significant diminishing returns}, a critical threshold that we term the \textbf{convergence point}. Typically, \textit{$\Delta$Thought} corresponds to a one–sub-thought increment. Meanwhile, \Cref{sec:overthink_manage} details approximating the \textit{return} via graph-induced characteristics.

\subsection{Thought Progression Patterns}
\label{sec:thought_patterns}
The two identified patterns are distinguished by \textit{how the probability of arriving at the correct answer is distributed} throughout the thought process. 

\begin{figure}[t]
    \centering
    \includegraphics[width=0.5\textwidth]{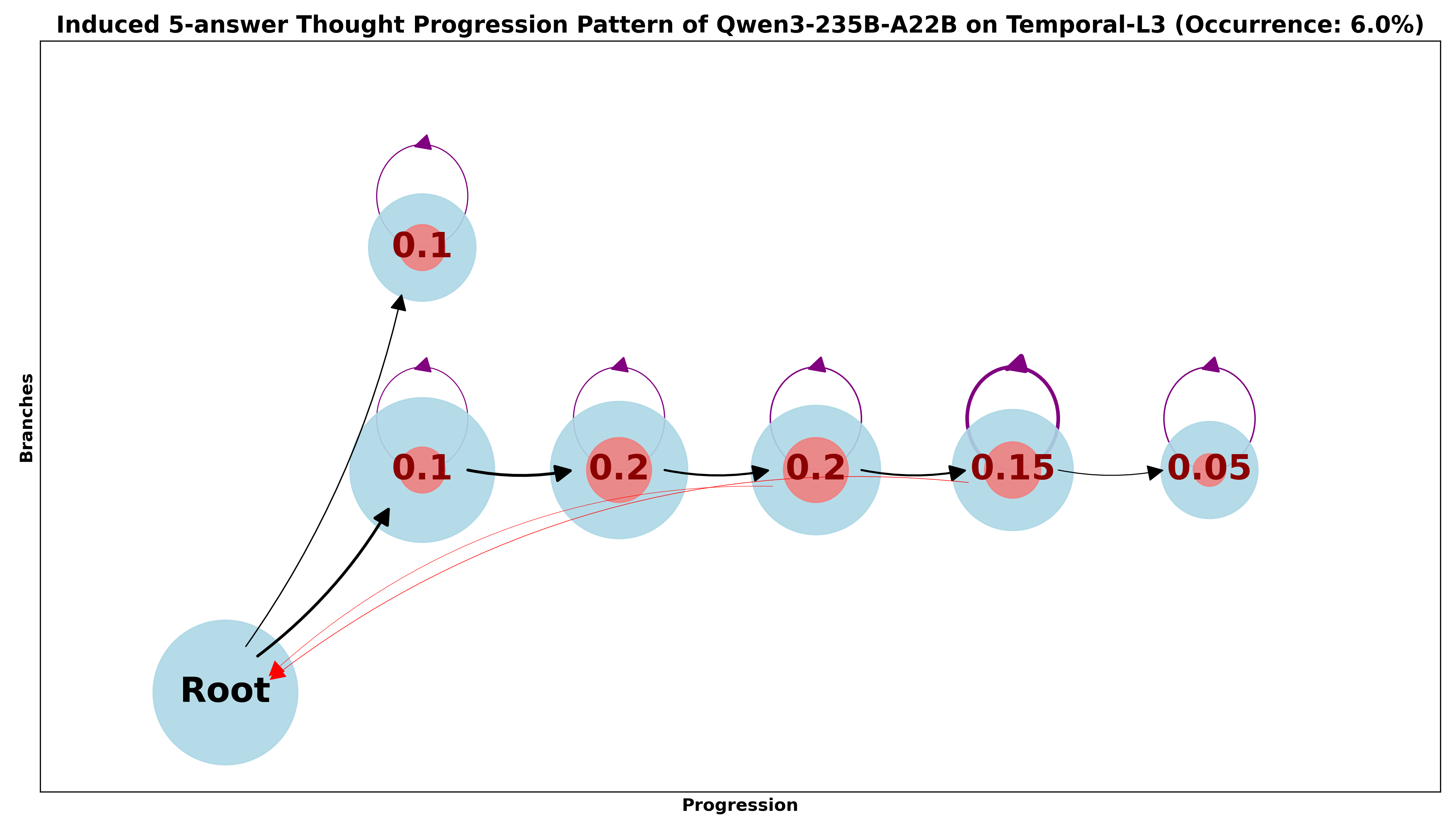}
    \caption{The typical \textbf{Explorer} thought progression pattern (5 distinct answer case). The size of the blue nodes indicates the visit frequency, while the size of the red nodes (and associated values) indicates the probability of the ground truth being present at that node. Due to the exploratory behavior, \textit{multiple reasoning branches emerge and the correct answer can be discovered at any stage of the thought process}. Edge thickness indicates the edge frequency, and \textcolor{red}{red dashed curve} denotes the occurrence of backtracking, where the model abandon its current reasoning path. More plots in \Cref{appx:more_plots_explorer}.}
    \label{fig:explorer_pattern_group5}
\end{figure}

\paragraph{Explorer.} This pattern (\Cref{fig:explorer_pattern_group5}) is characterized by the ground-truth answer being spread out across almost all nodes (i.e., distinct answers) in the thought process. In this mode, nearly every distinct intermediate answer has a non-trivial probability of being correct. A potential side effect of this exploratory behavior is that a correct answer might be discovered early in the process. 
As a result, this pattern of overthinking manifests as \textit{excessive exploration}, where the model evaluates numerous alternatives, often leading it to refute its earlier conclusions, a behavior dubbed \textit{backtrack}. 
While this exploration can cause the model to discard correct answers prematurely, returning to an earlier explored answer after backtracking significantly boosts the answer's credibility, which is analogous to reaching the same answer via two independent, parallel reasoning paths.

\texttt{Qwen3-235B-A22B}, the largest evaluated model, exhibits this \textit{exploratory} pattern. Note, the pattern is tied to models, irrespective of input prompts in general (see \Cref{appx:more_plots_explorer} for plots on more tasks).

\begin{figure}[t]
    \centering
    \includegraphics[width=0.5\textwidth]{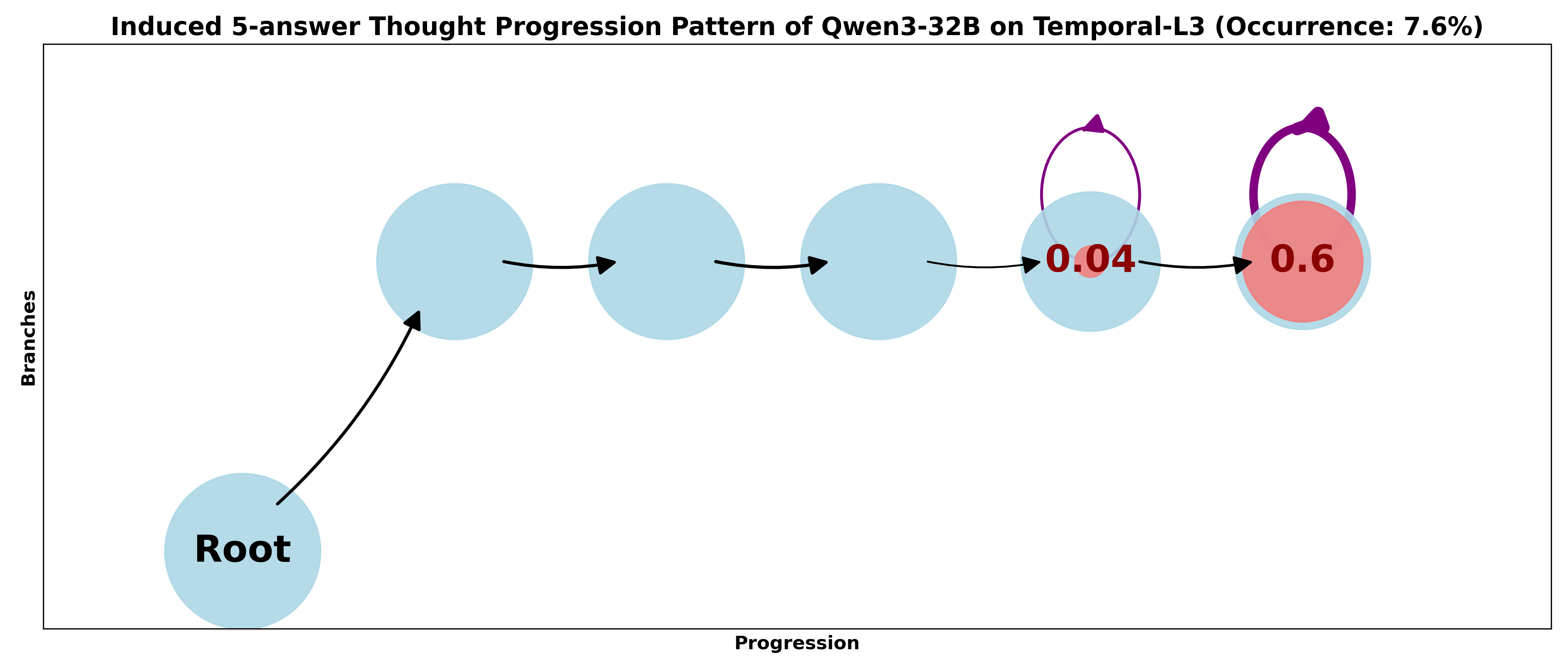}
    \caption{The typical \textbf{Late Landing} thought progression pattern (5 distinct answer case). The model follows a more linear path, with the probability of the ground-truth answer (indicated by the red node size and value) being highly concentrated at the terminal stage of the thought process. Towards the end, \textit{the model engages in over-verification, marked by a thick self-loop, to increase its confidence}. More plots in~\Cref{appx:more_plots_late_landing}.}
    \label{fig:late_lander_pattern_group5}
\end{figure}

\paragraph{Late Landing.} This pattern follows a ``convergent'' reasoning trajectory. As the thought process unfolds, the model gets closer to the correct answer, and the correctness probability increasingly concentrates towards the final proposal. 
As evinced in \Cref{fig:late_lander_pattern_group5}, early answers have extremely low correctness probabilities, while the final node overwhelmingly captures the correct answer. This behavior reflects a process of \textit{sequential-esque correction}, in contrast to the branching exploration observed in the Explorer pattern.
 Overthinking in this pattern manifests as \textit{excessive usage of self-verification}:\footnote{The \textbf{Explorer} also exhibits over-usage of verification, but second to over-exploration.} the model engages in an unnecessarily long chain of validating an already-correct answer, aiming for overly cautious confidence rather than stopping once the confidence is adequate.

Most open-weight thinking models, such as \texttt{R1-Distill-Llama-70B}, \texttt{Qwen3-30B-A3B}, and \texttt{Qwen3-32B}, exhibit this \textit{over-verification} pattern.

\subsection{Utility Tracing}
\label{sec:utility_tracing}
To quantify overthinking more structurally, we look into \textit{utility tracing} by investigating the evolving performance as a function of the number of sub-thoughts. This analysis reveals distinct utility curves for our two previously identified thought patterns. Note, each sub-thought is analyzed for both \textbf{correctness} (whether the sub-thought  contains the correct answer itself) and \textbf{helpfulness} (whether the thoughts can lead to a correct answer in the \textit{final response}, i.e., post-\texttt{</think>} part shown in~\cref{fig:framework_overview}).\footnote{To evaluate helpfulness, we wrap prior sub-thoughts in \texttt{<think>} tokens, prompt the model to output responses only.}

\begin{figure}[t]
    \centering
    \includegraphics[width=0.5\textwidth]{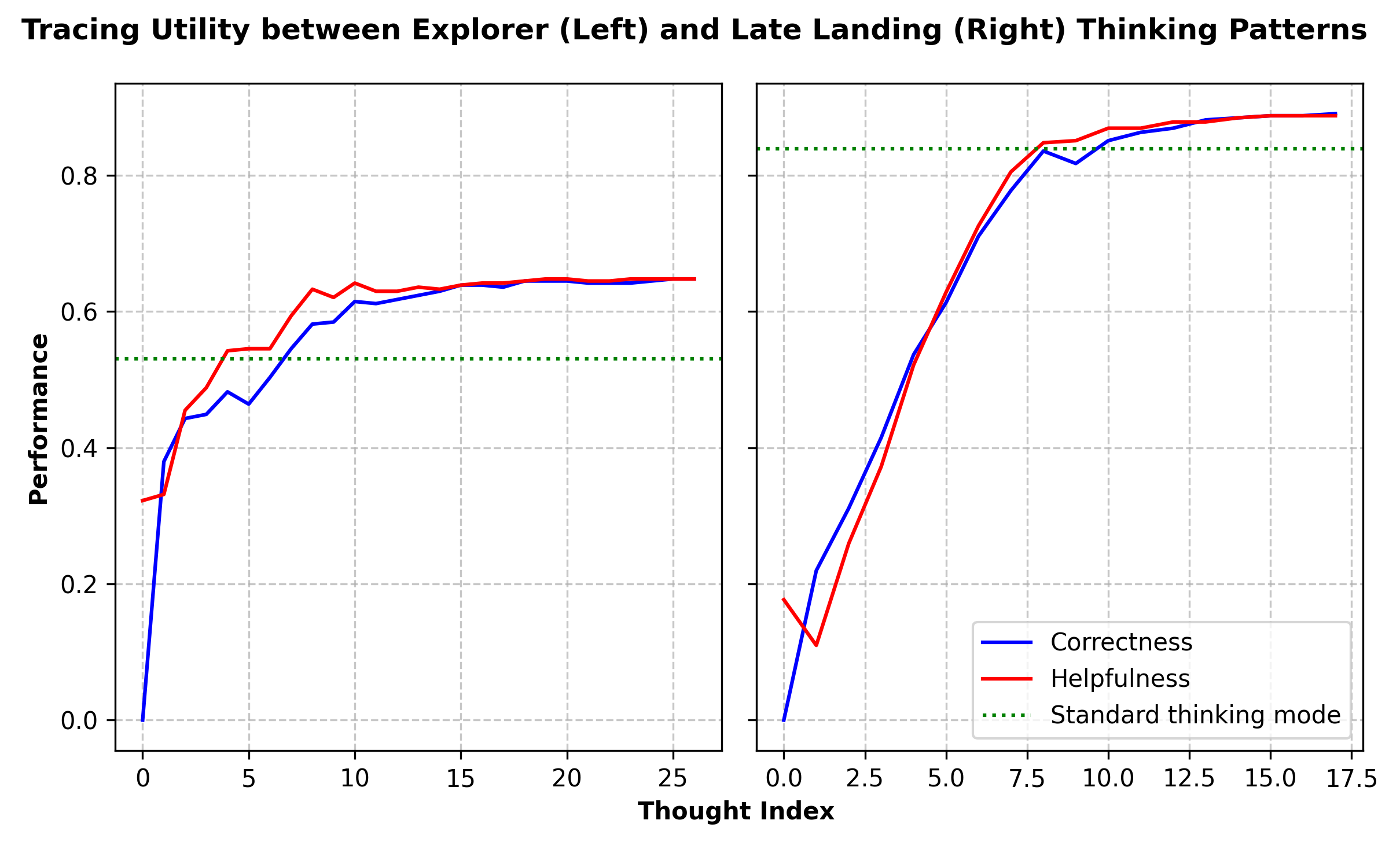}
    \vspace{-4mm}
    \caption{Tracing utilities on \textit{Temporal-L3} task between \textbf{Explorer} (\texttt{Qwen3-235B-A22B}) and \textbf{Late Landing} (e.g.,\texttt{Qwen3-30B}) thinking patterns. For \textbf{Explorer}, performance is volatile and peaks early, showing that further reasoning provides diminishing or even negative returns; while for \textbf{Late Landing}, it rises steadily to a plateau, and overthinking manifests as redundant steps after this convergence point. \textit{Thinking mode} denotes the model performance when no intervention is applied.}
    \label{fig:utility_evolution}
\end{figure}

As shown in \Cref{fig:utility_evolution}, the \textbf{Explorer} pattern displays volatile performance that often peaks early, which means excessively extended reasoning can yield diminishing or negative returns. Conversely, the \textbf{Late Landing} pattern shows a steady initial  performance increase (i.e., consistent slope); this progression suddenly converges to a stable plateau after a substantial number of sub-thoughts, where overthinking manifests as redundant steps (e.g., unnecessary verifications) past the \textit{convergence point}.

\subsection{Towards Overthinking Management}
\label{sec:overthink_manage}
We present a case study on the \textit{Temporal-L3} task to illustrate how our structure-based definition can be used to detect and manage overthinking in practice.

Based on the utility evolution of~\Cref{fig:utility_evolution}, we empirically set $\epsilon=0$ for \texttt{Qwen3-235B-A22B} and $\epsilon=1$ for \texttt{Qwen3-32B}.\footnote{We ignore the initial descent for one sub-thought case.} For both models, the convergence point is the completion of the \textit{eighth sub-thought}, classifying any additional thinking beyond this point as overthinking. Concretely, \Cref{fig:utility_evolution} shows that the introduction of an additional thought reduces performance ($63.25\rightarrow62.05$) or provides negligible benefit ($84.76\rightarrow85.06$). Notably, performances at this convergence point already surpasses the standard thinking baselines ($52.87$ and $83.84$, respectively).\footnote{4 samples were excluded due to \texttt{gemini-2.5-pro} parsing errors, accounting for minor discrepancies with \Cref{tab:qwen235b_performance_date}.}
This highlights that our redefinition detects convergence point, and thus identifies overthinking at a fine-grained, sub-thought level, offering a more robust measurement of overthinking than the prior sample-level, length-based definition. In fact, the prior definition can be viewed as a special case of our novel redefinition, where \textit{$\Delta$ Thought} is the \textit{difference in total length} between thinking and non-thinking outputs.

Further, to extend this redefinition to real-world inference settings without ground-truths, we introduce two heuristics derived from the behavioral patterns observed in~\Cref{sec:thought_patterns}. They serve as proxies for estimating the convergence point.

\begin{itemize}[nosep]
\item \textbf{Self-looping}: terminate once the model performs $k$ consecutive self-verifications after proposing an answer (i.e., visit a node).
\item \textbf{Backtrack}: terminate if and only if the model revisits a previously proposed answer \textit{as the direct result of a backtrack action}.
\end{itemize}

We start with using self-looping heuristic only ($K=2$), and we achieve $62.23$ and $68.90$ accuracies for \texttt{Qwen3-235B-A22B} and  \texttt{Qwen3-32B}, respectively, while average output length is halved (from $2{,}722 \rightarrow 1{,}315$ and $4{,}000 \rightarrow 1{,}874$). Model-specific tuning yields further improvements. For the Explorer-type \texttt{Qwen3-235B-A22B}, adding backtracking preserves accuracy while reducing length to $1{,}100$ words (nearly $60\%$ efficiency savings). For the Late-Landing-type \texttt{Qwen3-32B}, setting $k=3$ slightly encourages verification, raising accuracy to $80.18$ ($3$ points below the thinking mode) while cutting the inference cost by $40\%$ ($4k \rightarrow 2{,}463$).

Overall, this case study showcases that our structure-based redefinition not only pinpoints overthinking in post-hoc analysis, but also enables practical heuristics for managing overthinking at real-time inference without access to ground truths.

\section{Conclusion}
\vspace{-2mm}
In this work, we present \framework, a fine-grained analyzer that reconstructs LLM reasoning into sub-thoughts and progression graphs, uncovering the structural evolution of (over)-thinking.
Our large-scale preliminary study shows that models waste 5–20× more compute on simple tasks without accuracy gains, driven primarily by \textit{over-verification} and \textit{over-exploration}.
Based on these insights, we propose a utility-based redefinition of overthinking, grounded in thought structures that moves beyond length-based metrics.
Together, our \framework and findings provide a sharper lens into LLMs’ internal decision-making and overthinking management.

\section*{Acknowledgment}
We appreciate Chris Dyer and Ankur Taly for taking their time to provide insightful reviews of this manuscript.
We also thank Tejasi Latkar, Vikash Sehwag, Jonathan Lai, Shane Gu, Le Hou and members from Google DeepMind for their valuable feedback during the preparation of this work.

\clearpage

\setcounter{table}{0}
\setcounter{figure}{0}
\renewcommand{\thefigure}{A\arabic{figure}}
\renewcommand{\thetable}{A\arabic{table}}

\appendix
\section{Models Used in This Work}
\label{app:backbones}
This work utilized a diverse set of large language models (LLMs), especially those with long thinkiong capabilities. The models were primarily chosen to cover a wide range of parameter sizes, from smaller, more efficient models to very large, high-performing ones.

The complete list of models used in this study is provided below:
\begin{itemize}[nosep]
    \item \textbf{Qwen3 models (supports both thinking and non-thinking modes)}:
    \begin{itemize}[nosep]
        \item \texttt{Qwen3-0.6B}
        \item \texttt{Qwen3-1.7B}
        \item \texttt{Qwen3-4B}
        \item \texttt{Qwen3-8B}
        \item \texttt{Qwen3-14B}
        \item \texttt{Qwen3-30B-A3B}
        \item \texttt{Qwen3-32B}
        \item \texttt{Qwen3-235B-A22B}
    \end{itemize}
    \item \textbf{R1-distilled thinking models}:
    \begin{itemize}[nosep]
        \item \texttt{DeepSeek-R1-Distill-Owen-1.5B}
        \item \texttt{DeepSeek-R1-Distill-Owen-7B}
        \item \texttt{DeepSeek-R1-Distill-Owen-8B}
        \item \texttt{DeepSeek-R1-Distill-Owen-14B}
        \item \texttt{DeepSeek-R1-Distill-Owen-32B}
        \item \texttt{DeepSeek-R1-Distill-Llama-70B}
    \end{itemize}
\end{itemize}

We use vLLM~\citep{kwon2023efficient} to deploy aforementioned models.

\section{Data Used in This Work}
\label{app:data}
This work utilizes 6 data domains, covering 15 query clusters, for systematic evaluations. The six domains and associated datasets used in the horizontal analysis are displayed in~\Cref{tab:evaluation_datasets}.

The complete list of data domains and query clusters being evaluated in this study is provided below:
\bigskip
\begin{itemize}[nosep]
    \item \textbf{Knowledge Recall}:
    \begin{itemize}[nosep]
    \item Short-context Machine Reasoning Comprehension domain: SQuAD2.0~\citep{rajpurkar-etal-2018-know}.\footnote{\url{https://huggingface.co/datasets/rajpurkar/squad_v2}. The dataset is distributed under the CC BY-SA 4.0 license.}
    \item Long-context Fact Retrieval domain: NAIH~\citep{kamradt2023needle}.\footnote{\url{https://github.com/gkamradt/LLMTest_NeedleInAHaystack}. The dataset is distributed under the MIT license.}
    \item Factuality Question Answering domain: SimpleQA~\citep{simpleqa}.\footnote{\url{https://openai.com/index/introducing-simpleqa/}. The dataset is distributed under the MIT license.}
    \end{itemize}
    \item \textbf{Reasoning}:
    \begin{itemize}[nosep]
    \item Mathematical Reasoning domain (sorted by difficulty): ASDiv-grade1~\citep{miao-etal-2020-diverse}, ASDiv-grade2~\citep{miao-etal-2020-diverse}, ASDiv-grade3~\citep{miao-etal-2020-diverse}, ASDiv-grade4~\citep{miao-etal-2020-diverse}, ASDiv-grade5~\citep{miao-etal-2020-diverse},\footnote{\url{https://github.com/chaochun/nlu-asdiv-dataset}. The dataset is distributed under the CC BY-NC 4.0 license.} and GSM8k~\citep{gsm8k}.\footnote{\url{https://huggingface.co/datasets/openai/gsm8k}. The dataset is distributed under the MIT license.}
    \item Temporal Reasoning domain: Date arithmetic~\citep[a.k.a., Temporal-L1;][]{tan-etal-2023-towards}.\footnote{\url{https://huggingface.co/datasets/tonytan48/TempReason}. This dataset is released on a public GitHub.} Temporal-L2 to L5 are derived from~\citet{tan-etal-2023-towards} following our designed curation guideline (\Cref{appx:temp_data_creation}).
    \item Logical Reasoning domain: Zebra Logic~\citep[Easy;][]{DBLP:journals/corr/abs-2502-01100}.\footnote{\url{https://huggingface.co/blog/yuchenlin/zebra-logic}. The dataset is distributed under the CC BY 4.0 license.}
    \end{itemize}
\end{itemize}

Note, the usage of each dataset is consistent with its intended use, and these datasets present no more than minimal risk.

\begin{table*}[t]
\centering
\resizebox{1.0\textwidth}{!}{%
\begin{tabular}{lrrrrrr}
\toprule
\textbf{Category} & \multicolumn{3}{c}{\textbf{Knowledge Recall}} & \multicolumn{3}{c}{\textbf{Reasoning}} \\  \cmidrule(lr){2-4}   \cmidrule(lr){5-7}
\textbf{Domain} & Short-ctx MRC & Long-ctx Fact Retrieval & Factuality QA & Math & Temporal & Logical \\
\textbf{Dataset} & SQuAD 2.0 & NIAH & SimpleQA & ASDIV & Date Arithmetic & Zebra Logic \\
\textbf{Size} & 420 (210) & 352 (132) & 400 & 195 & 400 & 240 \\
\bottomrule
\end{tabular}
}
\caption{Datasets used for benchmarking in horizontal analysis. We include six datasets spanning two major task domains---knowledge recall and reasoning. The number within parentheses denote the portion size of \textit{non-answerable} queries given the provided context. }
\label{tab:evaluation_datasets}
\end{table*}

\section{More Plots for Horizontal Study}
\label{appx:horizontal}
\begin{figure*}[t]
    \centering
    \includegraphics[width=0.9\textwidth]{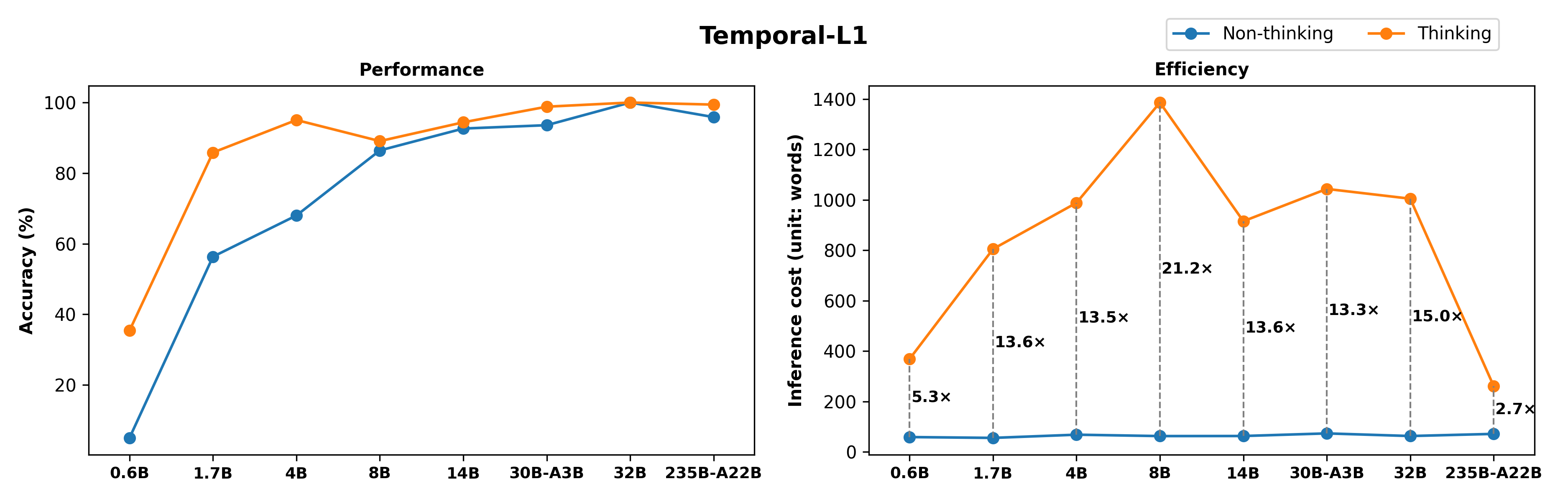}
    \caption{Performance and inference-time efficiency trends of \texttt{Qwen3} models at different scales on simple temporal reasoning.}
    \label{fig:perf_eff_trend_date}
    \vspace{-4mm}
\end{figure*}

\begin{figure*}[]
    \centering
    \includegraphics[width=0.9\textwidth]{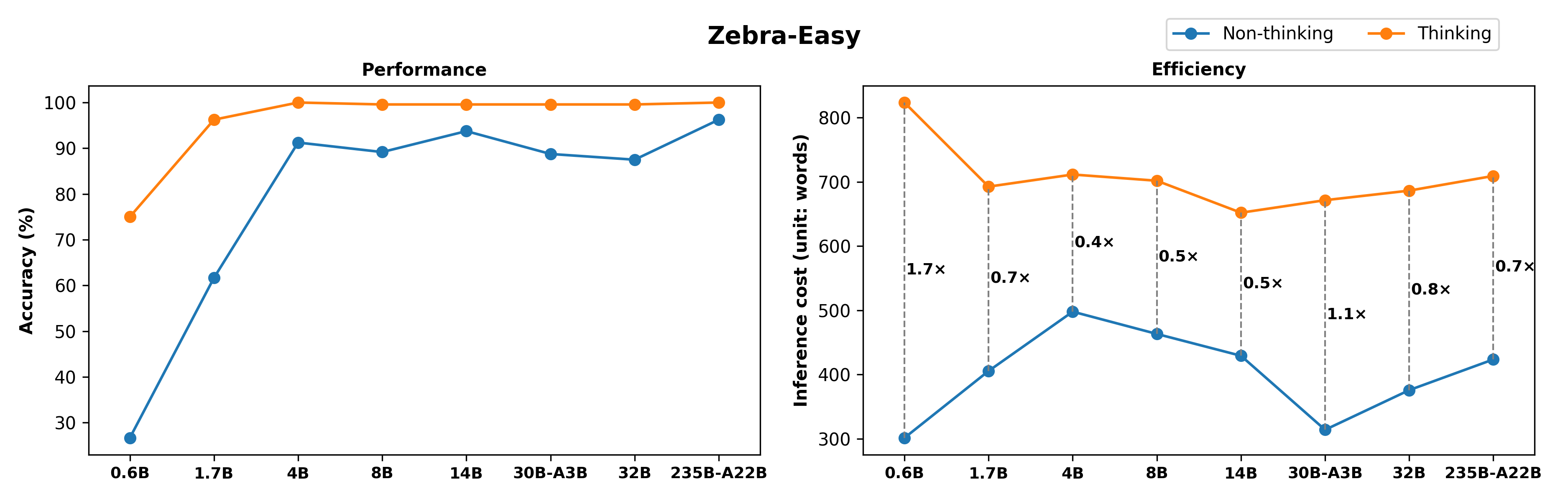}
    \caption{Performance and inference-time efficiency trends of \texttt{Qwen3} models at different scales on simple logic reasoning.}
    \label{fig:perf_eff_trend_logic}
    \vspace{-4mm}
\end{figure*}

\begin{figure*}[]
    \centering
    \includegraphics[width=0.9\textwidth]{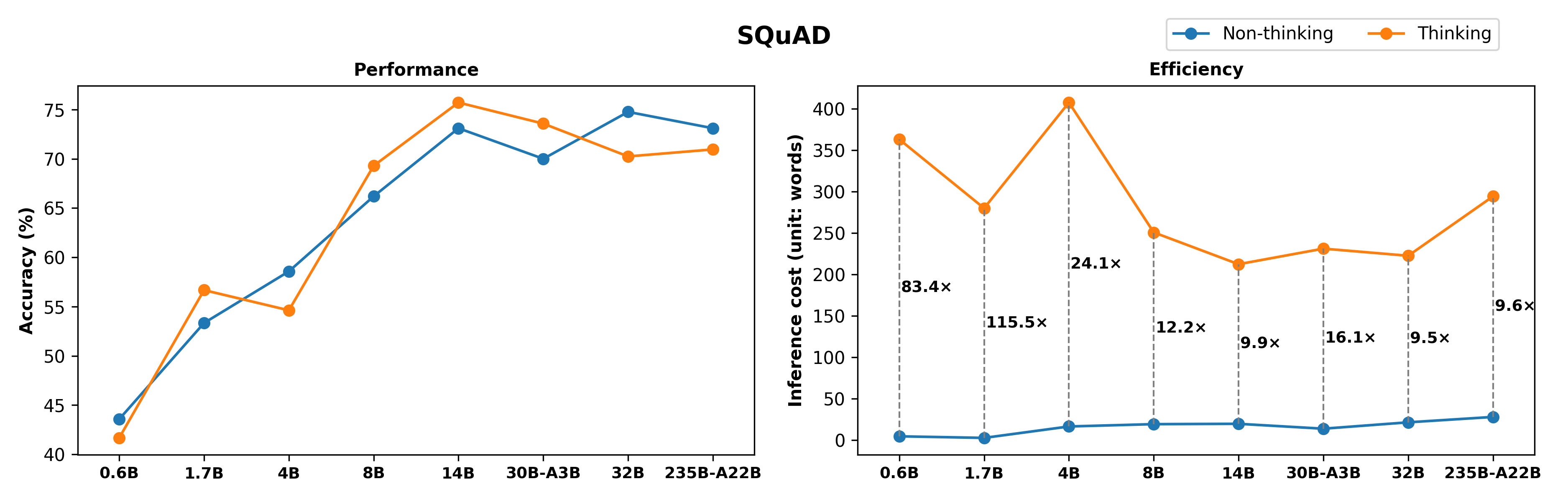}
    \caption{Performance and inference-time efficiency trends of \texttt{Qwen3} models at different scales on simple short-context knowledge recall (machine reading comprehension).}
    \label{fig:perf_eff_trend_squad}
    \vspace{-4mm}
\end{figure*}

\begin{figure*}[]
    \centering
    \includegraphics[width=0.9\textwidth]{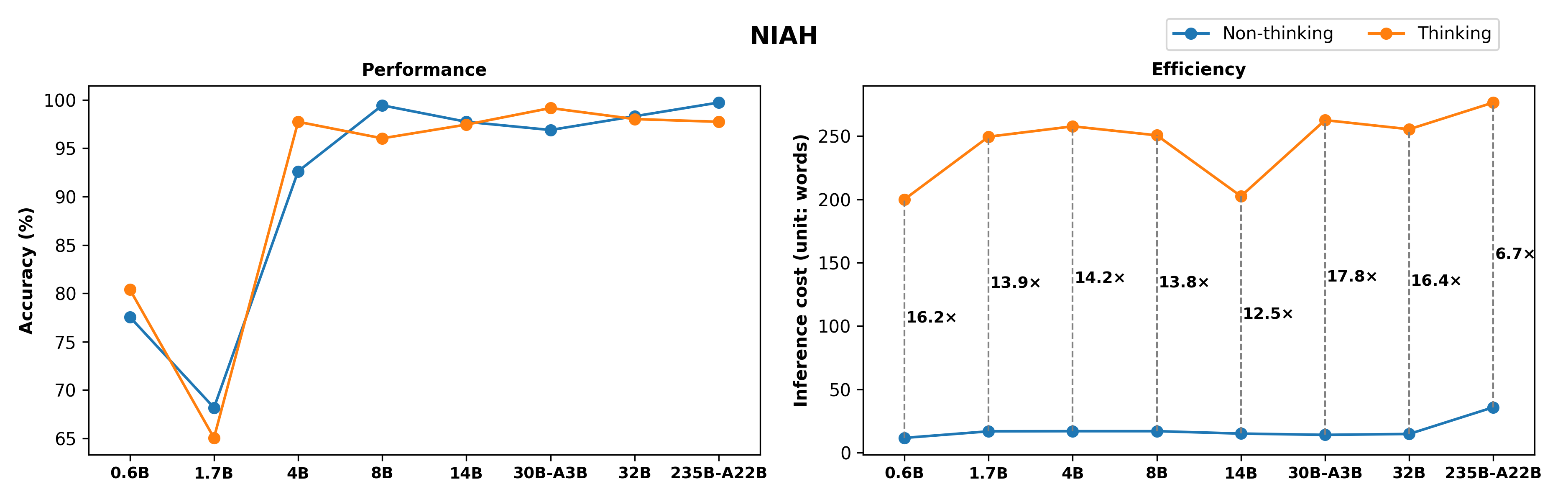}
    \caption{Performance and inference-time efficiency trends of \texttt{Qwen3} models at different scales on simple long-context knowledge recall (fact retrieval).}
    \label{fig:perf_eff_trend_niah}
    \vspace{-4mm}
\end{figure*}

\begin{figure*}[]
    \centering
    \includegraphics[width=0.9\textwidth]{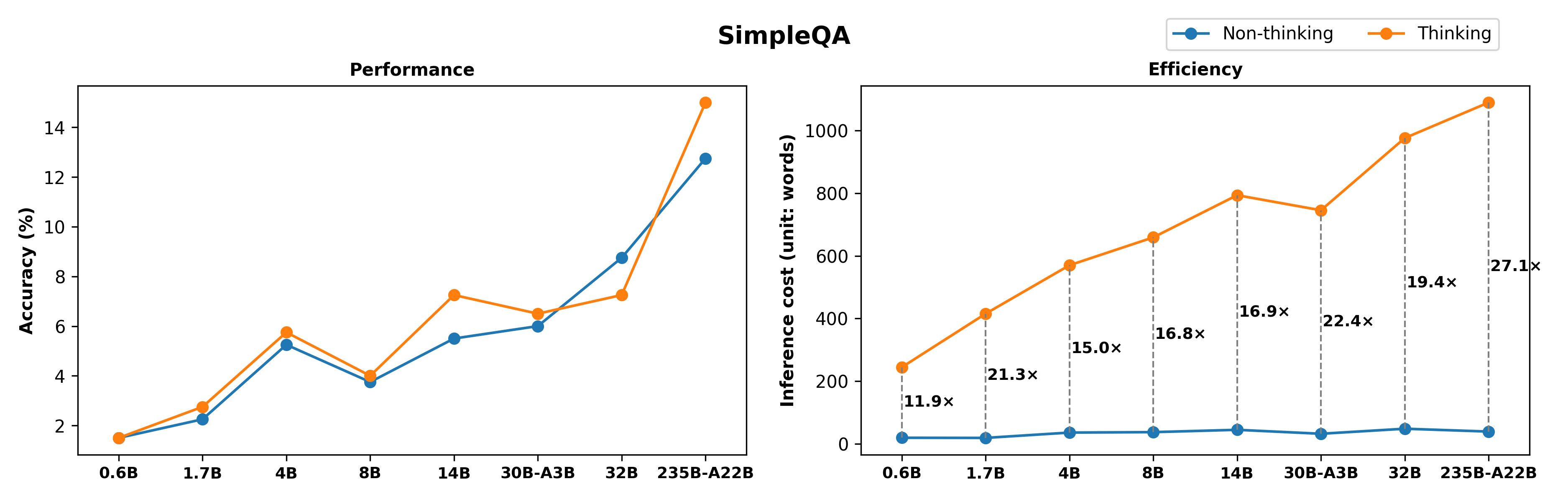}
    \caption{Performance and inference-time efficiency trends of \texttt{Qwen3} models at different scales on a factuality QA task.}
    \label{fig:perf_eff_trend_simpleqa}
    \vspace{-4mm}
\end{figure*}

\Cref{fig:perf_eff_trend_date}, \ref{fig:perf_eff_trend_logic}, \ref{fig:perf_eff_trend_squad}, \ref{fig:perf_eff_trend_niah}, and \ref{fig:perf_eff_trend_simpleqa} display the performance and inference-time efficiency trends of \texttt{Qwen3} models on temporal reasoning, logical reasoning, Short-context MRC, Long-context Fact Retrieval, and Factuality QA tasks, respectively.

\section{Multi-level Temporal Reasoning (Date Arithmetic) Data Curation Process}
\label{appx:temp_data_creation}
A date arithmetic question from the original temporal reasoning dataset~\citep{tan-etal-2023-towards} is like ``\textit{What is the time 10 year and 1 month after Feb, 1068}'', which contains two important components: starting time and duration. 

We perform systematic adjustments as follows to gradually increase the task complexity: 
\begin{itemize}[nosep]
    \item From level-1 to level-2, we add \textit{day} information to both starting time and duration.
    \item From level-2 to level-3, we convert the duration from the standard representation \textit{(year, month and day)} to day-only representation, i.e., 10 year, 0 month and 21 day is equivalent to $3,674$ days if the starting date is Feb 26, 1068.
    \item For level-4 and level-5, we double and triple the duration, respectively. In order to determine target dates, we use online date calculator to ensure the process flawless. 
\end{itemize}

We show sample examples in~\Cref{tab:temporal_sample} for each difficulty level. 

\begin{table*}[t]
\centering
\resizebox{0.9\textwidth}{!}{%
\begin{tabular}{llr} \toprule
Difficulty Level & Question                                                        & \multicolumn{1}{l}{Target} \\ \midrule
Level-1          & What is the time 10 year and 1 month after Feb, 1068            & Mar, 1078                  \\
Level-2          & What is the time 10 year, 0 month and 21 day after Feb 26, 1068 & Mar 19, 1078               \\
Level-3          & What is the time 3674 day after Feb 26, 1068                    & Mar 19, 1078               \\
Level-4          & What is the time 7348 day after Feb 26, 1068                    & Apr 9, 1088                \\
Level-5          & What is the time 11022 day after Feb 26, 1068                   & May 1, 1098               \\  \bottomrule
\end{tabular}
}
\caption{Sample examples for different difficulty levels of the temporal reasoning task.}
\label{tab:temporal_sample}
\end{table*}

\section{Full Benchmark Results of \Cref{sec:benchmark}}
\Cref{tab:benchmark}, \Cref{tab:benchmark_math} and \Cref{tab:benchmark_date} display the full benchmark results using the \texttt{Qwen3} model family.

\Cref{tab:benchmark_r1}, \Cref{tab:benchmark_math_r1} and \Cref{tab:benchmark_date_r1} display the full benchmark results using the \texttt{DeepSeek-R1-Distill} model family.

\begin{table*}[]
\centering
\resizebox{1\textwidth}{!}{
\begin{tabular}{@{}lcccccccccccc@{}}
\toprule
\multirow{2}{*}{Model} & \multicolumn{2}{c}{SQuAD} & \multicolumn{2}{c}{NIAH} & \multicolumn{2}{c}{SimpleQA} & \multicolumn{2}{c}{ASDiv-1} & \multicolumn{2}{c}{Date Arithmetic (L1)} & \multicolumn{2}{c}{Zebra (Easy)} \\ 
\cmidrule(lr){2-3} \cmidrule(lr){4-5} \cmidrule(lr){6-7} \cmidrule(lr){8-9} \cmidrule(lr){10-11} \cmidrule(lr){12-13} & 
Performance & Efficiency & Performance & Efficiency & Performance & Efficiency & Performance & Efficiency & Performance & Efficiency & Performance & Efficiency\\
\midrule
Qwen3-0.6B & 43.57/41.67 & 4.3/362.8(360.6) & 77.56/80.40 & 11.6/200.0(192.4) & 1.50/1.50 & 19.0/244.3(232.5) & 87.69/97.44 & 39.1/174.1(148.8) & 5.01/35.40 & 58.6/368.5(332.4) & 26.67/75.00 & 301.3/823.6(642.6) \\
Qwen3-1.7B & 53.33/56.67 & 2.4/279.7(275.9) & 68.18/65.06 & 16.8/249.5(227.3) & 2.25/2.75 & 18.6/415.3(384.9) & 94.87/99.49 & 40.2/182.2(148.3) & 56.34/85.84 & 55.3/804.8(720.3) & 61.67/96.25 & 405.5/692.6(488.5) \\
Qwen3-4B & 58.57/64.52 & 16.2/407.3(382.7) & 92.61/97.73 & 16.9/257.7(229.0) & 5.25/5.75 & 35.6/570.0(501.2) & 99.49/100.00 & 44.5/222.5(178.6) & 91.74/97.05 & 68.0/988.9(880.7) & 91.25/100.00 & 497.9/711.5(486.2) \\
Qwen3-8B & 66.19/69.29 & 19.0/250.5(226.6) & 99.43/96.02 & 16.9/250.6(219.7) & 3.75/4.00 & 37.0/659.0(562.9) & 97.44/98.97 & 43.3/338.3(267.4) & 86.43/89.97 & 62.5/1386.0(1265.5) & 89.17/99.58 & 463.2/701.8(479.0) \\
Qwen3-14B & 73.10/75.71 & 19.4/212.1(203.8) & 97.73/97.44 & 15.0/202.5(180.5) & 5.50/7.25 & 44.4/793.7(682.8) & 96.92/98.46 & 41.8/274.1(229.8) & 92.63/94.40 & 62.8/915.0(792.1) & 93.75/99.58 & 429.2/652.0(428.6) \\
Qwen3-30B-A3B & 70.00/73.57 & 13.5/230.9(204.1) & 96.88/99.15 & 14.0/262.6(233.0) & 6.00/6.50 & 31.8/745.6(644.5) & 97.95/98.97 & 38.5/266.3(218.9) & 98.53/98.82 & 73.1/1043.2(925.4) & 88.75/99.58 & 313.9/671.5(451.4) \\
Qwen3-32B & 74.76/70.24 & 21.2/222.3(201.3) & 98.30/98.01 & 14.7/255.4(230.9) & 8.75/7.25 & 47.9/976.2(797.0) & 94.87/100.00 & 42.3/290.1(221.9) & 100.00/100.00 & 62.7/1004.4(883.3) & 87.50/99.58 & 375.7/686.3(459.6) \\
Qwen3-235B-A22B & 73.10/70.95 & 27.7/294.0(254.8) & 99.72/97.73 & 35.8/276.5(247.9) & 12.75/15.00 & 38.8/1089.7(953.2) & 97.44/100.00 & 46.5/320.0(255.8) & 95.87/99.41 & 71.2/261.0(162.0) & 96.25/100.00 & 423.5/907.3(678.1) \\
\bottomrule
\end{tabular}
}
\caption{Complete accuracy performance and inference efficiency (unit: words) of \texttt{Qwen3} model family on horizontal analysis. For both performance and efficiency metrics, The first number is for the non-thinking mode, while the second is for the thinking mode. The numbers in parentheses indicate the thought lengths. Results of \texttt{DeepSeek-R1-Distill} see~\Cref{tab:benchmark_r1}.}
\label{tab:benchmark}
\end{table*}

% Required packages in your LaTeX preamble:
% \usepackage{booktabs}
% \usepackage{graphicx}

\begin{table*}[]
\centering
\resizebox{\textwidth}{!}{
\begin{tabular}{@{}lcccccccccccc@{}}
\toprule
\multirow{2}{*}{Model} & \multicolumn{2}{c}{ASDiv-1} & \multicolumn{2}{c}{ASDiv-2} & \multicolumn{2}{c}{ASDiv-3} & \multicolumn{2}{c}{ASDiv-4} & \multicolumn{2}{c}{ASDiv-5} & \multicolumn{2}{c}{GSM8k} \\
\cmidrule(lr){2-3} \cmidrule(lr){4-5} \cmidrule(lr){6-7} \cmidrule(lr){8-9} \cmidrule(lr){10-11} \cmidrule(lr){12-13}
 & Performance & Efficiency & Performance & Efficiency & Performance & Efficiency & Performance & Efficiency & Performance & Efficiency & Performance & Efficiency\\
\midrule
Qwen3-0.6B & 87.69/97.44 & 39.1/174.1(148.8) & 86.18/91.76 & 45.0/242.3(216.7) & 77.75/90.25 & 45.7/274.2(245.4) & 70.43/80.73 & 53.0/312.0(278.4) & 55.48/82.19 & 110.0/412.0(343.9) & 50.75/74.75 & 84.9/582.4(529.4) \\
Qwen3-1.7B & 94.87/99.49 & 40.2/182.2(148.3) & 87.65/94.71 & 45.3/228.9(192.4) & 85.75/93.75 & 53.3/254.5(212.9) & 77.74/91.69 & 58.9/330.9(275.8) & 68.49/88.36 & 108.8/490.6(392.8) & 65.25/87.00 & 105.6/555.3(472.7) \\
Qwen3-4B & 99.49/100.00 & 44.5/222.5(178.6) & 95.00/95.88 & 51.4/286.6(233.5) & 91.00/96.50 & 55.7/346.9(281.5) & 83.06/95.35 & 69.4/454.4(368.8) & 77.40/90.41 & 121.4/695.9(556.9) & 76.00/91.75 & 112.0/786.1(667.6) \\
Qwen3-8B & 97.44/98.97 & 43.3/338.3(267.4) & 95.00/95.88 & 51.8/362.1(282.8) & 92.25/95.50 & 56.1/398.4(310.5) & 79.40/94.02 & 67.7/511.1(404.0) & 76.03/90.41 & 127.6/770.6(588.5) & 80.00/92.00 & 119.7/818.0(681.6) \\
Qwen3-14B & 96.92/98.46 & 41.8/274.1(229.8) & 93.82/96.18 & 50.3/313.3(257.5) & 88.00/96.25 & 52.8/351.8(283.4) & 81.06/97.01 & 71.3/499.2(400.5) & 73.29/90.41 & 134.3/829.1(664.7) & 74.25/93.50 & 110.8/781.1(654.1) \\
Qwen3-30B-A3B & 97.95/98.97 & 38.5/266.3(218.9) & 94.71/96.18 & 43.5/296.9(239.8) & 93.25/96.25 & 48.5/333.7(265.7) & 84.72/95.35 & 61.7/422.1(330.8) & 75.34/89.73 & 131.9/769.7(604.6) & 79.50/94.00 & 112.7/771.3(639.8) \\
Qwen3-32B & 94.87/100.00 & 42.3/290.1(221.9) & 90.59/96.18 & 52.1/346.7(259.9) & 84.50/96.25 & 58.4/394.1(302.7) & 78.74/94.35 & 77.3/515.4(399.6) & 68.49/86.99 & 170.2/903.9(719.4) & 65.25/92.25 & 118.2/764.2(622.7) \\
Qwen3-235B-A22B & 97.44/100.00 & 46.5/320.0(255.8) & 92.94/95.88 & 54.4/348.5(274.5) & 89.00/96.75 & 59.7/429.8(349.7) & 83.39/93.36 & 74.8/553.1(450.8) & 76.03/90.41 & 127.3/701.5(553.9) & 74.75/91.50 & 118.1/1021.7(889.7) \\
\bottomrule
\end{tabular}
}
\caption{Complete accuracy performance and inference efficiency (unit: words) of \texttt{Qwen3} model family on vertical analysis (math reasoning). For both performance and efficiency metrics, The first number is for the non-thinking mode, while the second is for the thinking mode. The number in parentheses indicates the thought length. Results of \texttt{DeepSeek-R1-Distill} see~\Cref{tab:benchmark_math_r1}.}
\label{tab:benchmark_math}
\end{table*}

\begin{table*}[]
\centering
\resizebox{\textwidth}{!}{
\begin{tabular}{@{}lcccccccccc@{}}
\toprule
\multirow{2}{*}{Model} & \multicolumn{2}{c}{Temporal-L1} & \multicolumn{2}{c}{Temporal-L2} & \multicolumn{2}{c}{Temporal} & \multicolumn{2}{c}{Temporal-L4} & \multicolumn{2}{c}{Temporal-L5} \\
\cmidrule(lr){2-3} \cmidrule(lr){4-5} \cmidrule(lr){6-7} \cmidrule(lr){8-9} \cmidrule(lr){10-11}
 & Performance & Efficiency & Performance & Efficiency & Performance & Efficiency & Performance & Efficiency & Performance & Efficiency\\
\midrule
Qwen3-4B & 91.74/97.05 & 68.0/988.9(880.7) & 74.03/88.96 & 92.1/1748.1(1608.6) & 2.69/59.70 & 246.2/3933.3(3708.7) & 0.90/61.49 & 281.7/2200.7(1944.7) & 0.60/53.43 & 178.5/5134.5(4878.0) \\
Qwen3-8B & 86.43/89.97 & 62.5/1386.0(1265.5) & 77.31/90.75 & 84.2/2046.1(1900.1) & 8.96/81.79 & 352.9/4212.5(3927.5) & 3.58/80.60 & 335.2/3748.2(3503.6) & 2.39/82.69 & 372.0/3807.9(3502.1) \\
Qwen3-14B & 92.63/94.40 & 62.8/915.0(792.1) & 81.79/98.51 & 85.3/1924.3(1739.0) & 8.36/88.06 & 318.7/3408.5(3161.4) & 6.87/87.76 & 319.3/3823.8(3568.2) & 0.90/85.67 & 307.7/2993.3(2706.3) \\
Qwen3-30B-A3B & 98.53/98.82 & 73.1/1043.2(925.4) & 97.01/98.21 & 110.7/2306.3(2147.4) & 20.90/89.55 & 332.6/3684.5(3421.2) & 17.01/89.85 & 340.8/3624.8(3369.5) & 14.03/85.07 & 356.5/3601.8(3334.9) \\
Qwen3-32B & 100.0/100.0 & 62.7/1004.4(883.3) & 93.13/95.52 & 93.2/2198.1(2014.5) & 17.61/82.69 & 377.1/3992.6(3714.7) & 12.24/80.60 & 335.1/4001.9(3738.7) & 7.46/80.30 & 334.8/4217.6(3963.2) \\
Qwen3-235B-A22B & 95.87/99.41 & 71.2/261.0(162.0) & 95.52/97.61 & 114.2/679.7(534.7) & 32.24/52.54 & 309.2/2485.6(2255.8) & 14.63/47.76 & 334.6/2843.8(2625.4) & 8.66/45.37 & 323.9/3336.0(3078.9) \\
\bottomrule
\end{tabular}
}
\caption{Complete accuracy performance and inference efficiency (unit: words) of \texttt{Qwen3} model family on vertical analysis (temporal reasoning). For both performance and efficiency metrics, The first number is for the non-thinking mode, while the second is for the thinking mode. The number in parentheses indicates the thought length. Results of \texttt{DeepSeek-R1-Distill} see~\Cref{tab:benchmark_date_r1}.}
\label{tab:benchmark_date}
\end{table*}

\begin{table*}[]
\centering
\resizebox{1\textwidth}{!}{
\begin{tabular}{@{}lcccccccccccc@{}}
\toprule
\multirow{2}{*}{Model} & \multicolumn{2}{c}{SQuAD} & \multicolumn{2}{c}{NIAH} & \multicolumn{2}{c}{SimpleQA} & \multicolumn{2}{c}{ASDiv-1} & \multicolumn{2}{c}{Date Arithmetic (L1)} & \multicolumn{2}{c}{Zebra (Easy)} \\ 
\cmidrule(lr){2-3} \cmidrule(lr){4-5} \cmidrule(lr){6-7} \cmidrule(lr){8-9} \cmidrule(lr){10-11} \cmidrule(lr){12-13}
 & Performance & Efficiency & Performance & Efficiency & Performance & Efficiency & Performance & Efficiency & Performance & Efficiency & Performance & Efficiency \\ 
\midrule
DeepSeek-R1-Distill-Qwen-1.5B & 35.48 & 317.0 (310.8) & 5.11 & 2818.2 (2792.7) & 0.00 & - (-) & 65.13 & 280.2 (253.4) & 31.86 & 331.1 (315.8) & 12.92 & 653.0 (470.7) \\
DeepSeek-R1-Distill-Qwen-7B   & 47.38 & 221.0 (210.6) & 7.10 & 858.0 (835.7)   & 0.00 & - (-) & 93.85 & 119.6 (57.1)  & 33.92 & 146.2 (58.3)  & 69.58 & 855.6 (617.1) \\
DeepSeek-R1-Distill-Llama-8B  & 51.67 & 232.9 (224.9) & 65.06 & 388.7 (371.5)  & 0.25 & 606.0 (583.0) & 93.33 & 135.9 (64.7) & 69.03 & 216.3 (70.9)  & 90.42 & 901.9 (667.7) \\
DeepSeek-R1-Distill-Qwen-14B  & 63.10 & 228.5 (207.1) & 82.67 & 287.2 (262.9)  & 7.00 & 500.3 (480.9) & 94.36 & 120.9 (59.2) & 88.79 & 167.6 (65.3)  & 95.42 & 778.7 (562.0) \\
DeepSeek-R1-Distill-Qwen-32B  & 71.90 & 217.8 (198.1) & 88.35 & 254.4 (234.3)  & 7.25 & 397.1 (377.8) & 95.90 & 129.3 (61.3) & 98.82 & 129.8 (60.1)  & 98.75 & 720.7 (497.6) \\
DeepSeek-R1-Distill-Llama-70B & 69.29 & 235.6 (215.5) & 93.75 & 213.6 (198.0)  & 20.00 & 401.6 (382.7) & 95.90 & 121.4 (55.8) & 93.22 & 243.9 (203.4) & 98.75 & 733.1 (519.9) \\
\bottomrule
\end{tabular}
}
\caption{Complete accuracy performance and inference efficiency (unit: words) of \texttt{DeepSeek-R1-Distill} model family on horizontal analysis. The numbers in parentheses indicate the thought lengths.}
\label{tab:benchmark_r1}
\end{table*}

\begin{table*}[]
\centering
\resizebox{\textwidth}{!}{
\begin{tabular}{@{}lcccccccccccc@{}}
\toprule
\multirow{2}{*}{Model} & \multicolumn{2}{c}{ASDiv-1} & \multicolumn{2}{c}{ASDiv-2} & \multicolumn{2}{c}{ASDiv-3} & \multicolumn{2}{c}{ASDiv-4} & \multicolumn{2}{c}{ASDiv-5} & \multicolumn{2}{c}{GSM8k} \\
\cmidrule(lr){2-3} \cmidrule(lr){4-5} \cmidrule(lr){6-7} \cmidrule(lr){8-9} \cmidrule(lr){10-11} \cmidrule(lr){12-13}
 & Performance & Efficiency & Performance & Efficiency & Performance & Efficiency & Performance & Efficiency & Performance & Efficiency & Performance & Efficiency \\
\midrule
DeepSeek-R1-Distill-Qwen-1.5B & 65.13 & 280.2 (253.4) & 49.12 & 278.6 (245.0) & 50.75 & 298.0 (260.3) & 33.22 & 321.6 (287.9) & 39.04 & 392.1 (357.7) & 27.00 & 372.2 (302.9) \\
DeepSeek-R1-Distill-Qwen-7B   & 93.85 & 119.6 (57.1)  & 91.47 & 134.1 (61.4)  & 81.25 & 146.6 (67.4)  & 77.74 & 170.2 (76.9)  & 75.34 & 236.2 (101.8) & 72.25 & 239.1 (98.5)  \\
DeepSeek-R1-Distill-Llama-8B  & 93.33 & 135.9 (64.7)  & 90.29 & 147.0 (68.4)  & 86.00 & 161.6 (76.4)  & 77.41 & 173.3 (80.7)  & 71.71 & 252.3 (114.4) & 71.50 & 249.1 (111.9) \\
DeepSeek-R1-Distill-Qwen-14B  & 94.36 & 120.9 (59.2)  & 91.18 & 133.7 (62.7)  & 85.75 & 157.5 (72.1)  & 77.74 & 173.5 (78.4)  & 80.14 & 234.8 (101.2) & 83.00 & 246.0 (106.6)  \\
DeepSeek-R1-Distill-Qwen-32B  & 95.90 & 129.3 (61.3)  & 93.24 & 145.6 (70.1)  & 89.50 & 164.3 (73.5)  & 87.38 & 181.5 (80.9)  & 84.93 & 258.3 (115.4) & 88.00 & 254.5 (106.2)  \\
DeepSeek-R1-Distill-Llama-70B & 95.90 & 121.4 (55.8)  & 92.65 & 136.6 (62.4)  & 86.50 & 154.2 (67.7)  & 84.39 & 174.0 (74.6)  & 74.66 & 252.3 (105.2) & 79.50 & 263.7 (105.9)  \\
\bottomrule
\end{tabular}
}
\caption{Complete accuracy performance and inference efficiency (unit: words) of \texttt{DeepSeek-R1-Distill} model family on vertical analysis (math reasoning). The numbers in parentheses indicate the thought lengths.}
\label{tab:benchmark_math_r1}
\end{table*}

\begin{table*}[]
\centering
\resizebox{\textwidth}{!}{
\begin{tabular}{@{}lcccccccccc@{}}
\toprule
\multirow{2}{*}{Model} & \multicolumn{2}{c}{Temporal-L1} & \multicolumn{2}{c}{Temporal-L2} & \multicolumn{2}{c}{Temporal-L3} & \multicolumn{2}{c}{Temporal-L4} & \multicolumn{2}{c}{Temporal-L5} \\
\cmidrule(lr){2-3} \cmidrule(lr){4-5} \cmidrule(lr){6-7} \cmidrule(lr){8-9} \cmidrule(lr){10-11}
 & Performance & Efficiency & Performance & Efficiency & Performance & Efficiency & Performance & Efficiency & Performance & Efficiency \\
\midrule
DeepSeek-R1-Distill-Qwen-1.5B & 31.86 & 331.1 (315.8) & 11.94 & 431.5 (413.3) & 0.00 & - (-) & 0.00 & - (-) & 0.00 & - (-) \\
DeepSeek-R1-Distill-Qwen-7B   & 33.92 & 146.2 (58.3)  & 67.76 & 182.9 (64.9)  & 2.39 & 453.9 (127.8) & 1.79 & 1866.2 (155.0) & 1.19 & 546.8 (125.0) \\
DeepSeek-R1-Distill-Llama-8B  & 69.03 & 216.3 (70.9)  & 59.10 & 265.0 (92.7)  & 1.49 & 419.4 (171.0) & 0.30 & 332.0 (187.0) & 1.19 & 315.5 (125.0) \\
DeepSeek-R1-Distill-Qwen-14B  & 88.79 & 167.6 (65.3)  & 55.22 & 209.8 (96.1)  & 4.18 & 486.0 (167.7) & 3.58 & 424.3 (150.8) & 2.99 & 402.0 (156.9) \\
DeepSeek-R1-Distill-Qwen-32B  & 98.82 & 129.8 (60.1)  & 91.34 & 153.2 (72.0)  & 3.28 & 444.1 (173.8) & 4.78 & 509.3 (165.9) & 2.69 & 444.7 (144.1) \\
DeepSeek-R1-Distill-Llama-70B & 93.22 & 243.9 (203.4) & 91.04 & 381.9 (307.0) & 36.42 & 1098.3 (999.3) & 24.18 & 1411.3 (1333.8) & 23.88 & 1382.6 (1295.7) \\
\bottomrule
\end{tabular}
}
\caption{Complete accuracy performance and inference efficiency (unit: words) of \texttt{DeepSeek-R1-Distill} model family on vertical analysis (temporal reasoning). The numbers in parentheses indicate the thought lengths.}
\label{tab:benchmark_date_r1}
\end{table*}

\section{Sub-thought Decomposition Details}

\subsection{Sub-thought Criteria}
\label{appx:sub_thought}
We define a sub-thought according to three strict criteria:
\begin{itemize}[nosep]
    \item \textbf{Self-contained}: It must be independently comprehensible without heavy reliance on other sub-thoughts.
    \item \textbf{Complete}: It must represent a full logical step, typically including an intent, an action or calculation, and a conclusion.
    \item \textbf{Answer-bearing}: It must explicitly state a proposed answer to the query.
\end{itemize}

\subsection{Discourse Label Details}
\label{appx:inference_label}
\begin{itemize}[nosep]
    \item \textbf{Initial}: The first attempt at solving the problem, which may include an outline of the overall plan. Each thought process has only one initial sub-thought.
    \item \textbf{Verification}: A confirmation of the correctness of the \textit{immediately preceding} sub-thought, without any changes or corrections.
    \item \textbf{Correction}: A modification or rectification of the \textit{immediately preceding} sub-thought.
    \item \textbf{Backtrack}: The abandonment of the current line of reasoning to revert to an earlier (but not immediately preceding) sub-thought and explore a different path.
    \item \textbf{Sidetrack}: An exploration of supplementary information that is not directly part of an alternative approach to the main question.
    \item \textbf{Branching Out}: An exploration of an alternative method or a different aspect of the problem, as opposed to merely adding information (sidetrack).
    \item \textbf{Final}: The sub-thought that delivers the final proposed answer.
\end{itemize}

\section{More Plots for Individual Thought Progression}
\label{appx:individual}
This section presents selected individual thought progression plots.

\begin{figure}[t]
    \centering
    \includegraphics[width=0.4\textwidth]{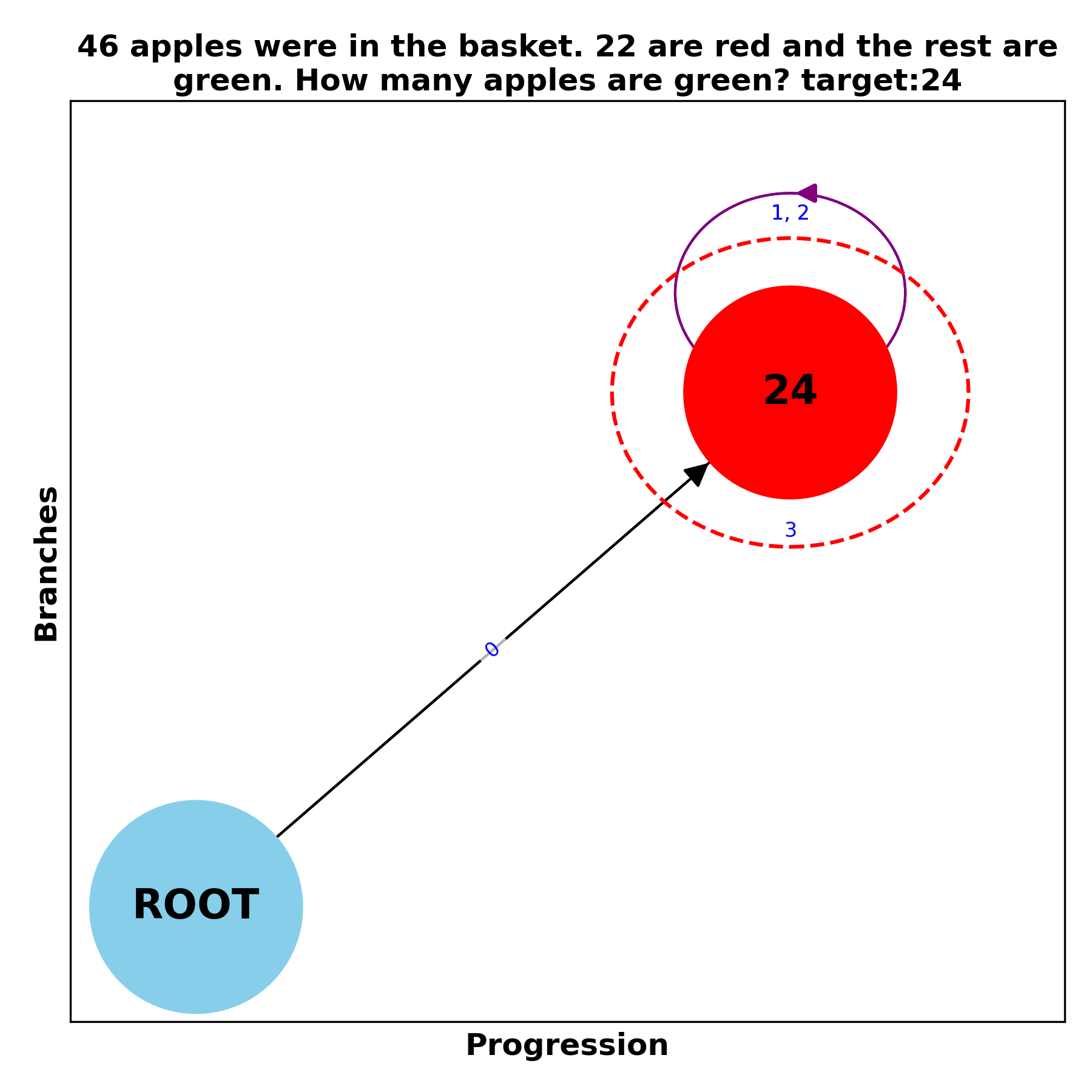}
    \caption{Individual thought progression graph of \texttt{Qwen3-235B-A22B} model on a sampled grade-1 arithmetic query.}
    \label{fig:progression_1}
\end{figure}

\Cref{fig:progression_1} displays the thought progression of the \texttt{Qwen3-235B-A22B} model on a grade-1 arithmetic query.

\begin{figure}[t]
    \centering
    \includegraphics[width=0.4\textwidth]{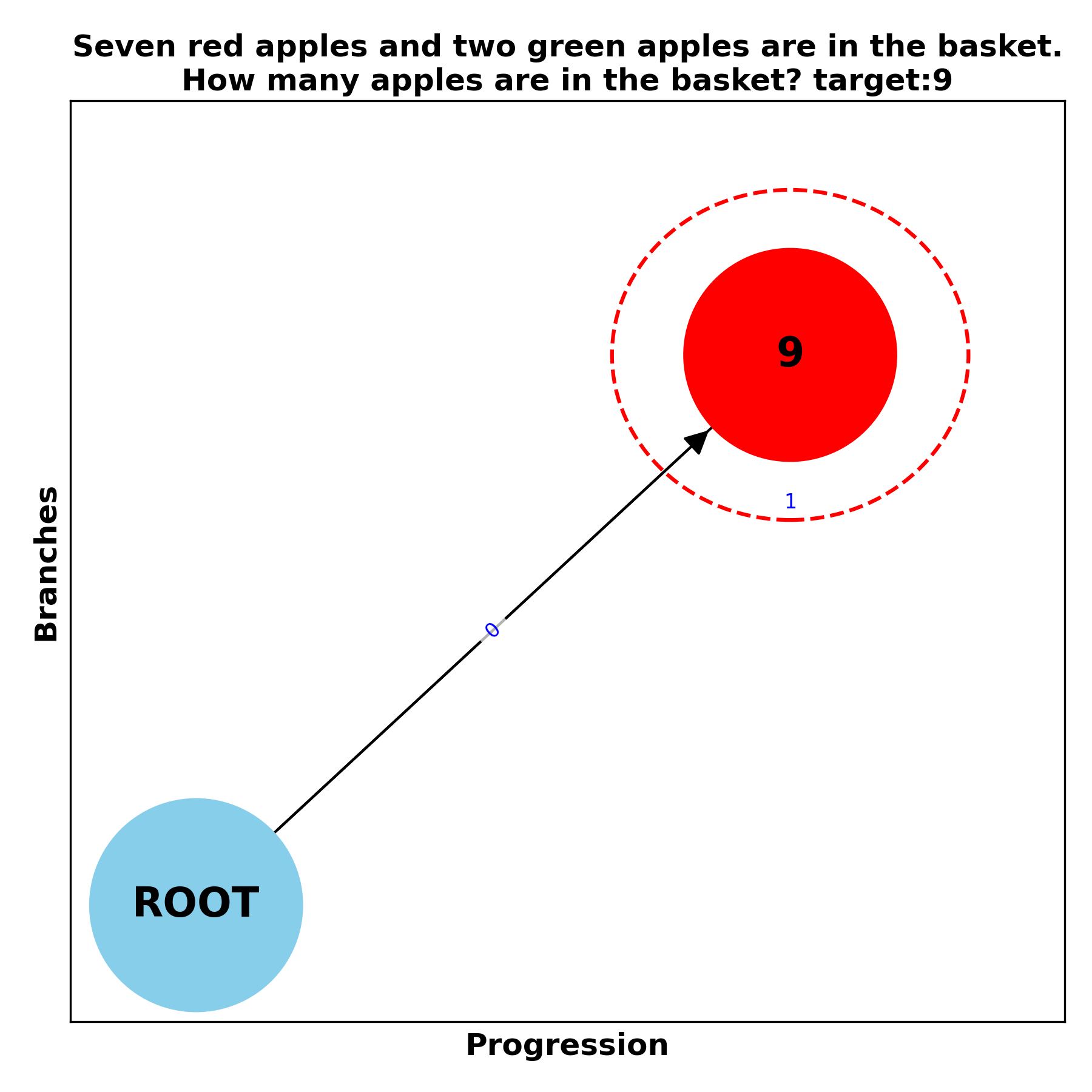}
    \caption{Individual thought progression graph of \texttt{R1-Distill-Llama-70B} model on a sampled grade-1 arithmetic query.}
    \label{fig:progression_2}
\end{figure}

\Cref{fig:progression_2} displays the thought progression of the \texttt{R1-Distill-Llama-70B} model on a grade-1 arithmetic query.

\begin{figure}[t]
    \centering
    \includegraphics[width=0.5\textwidth]{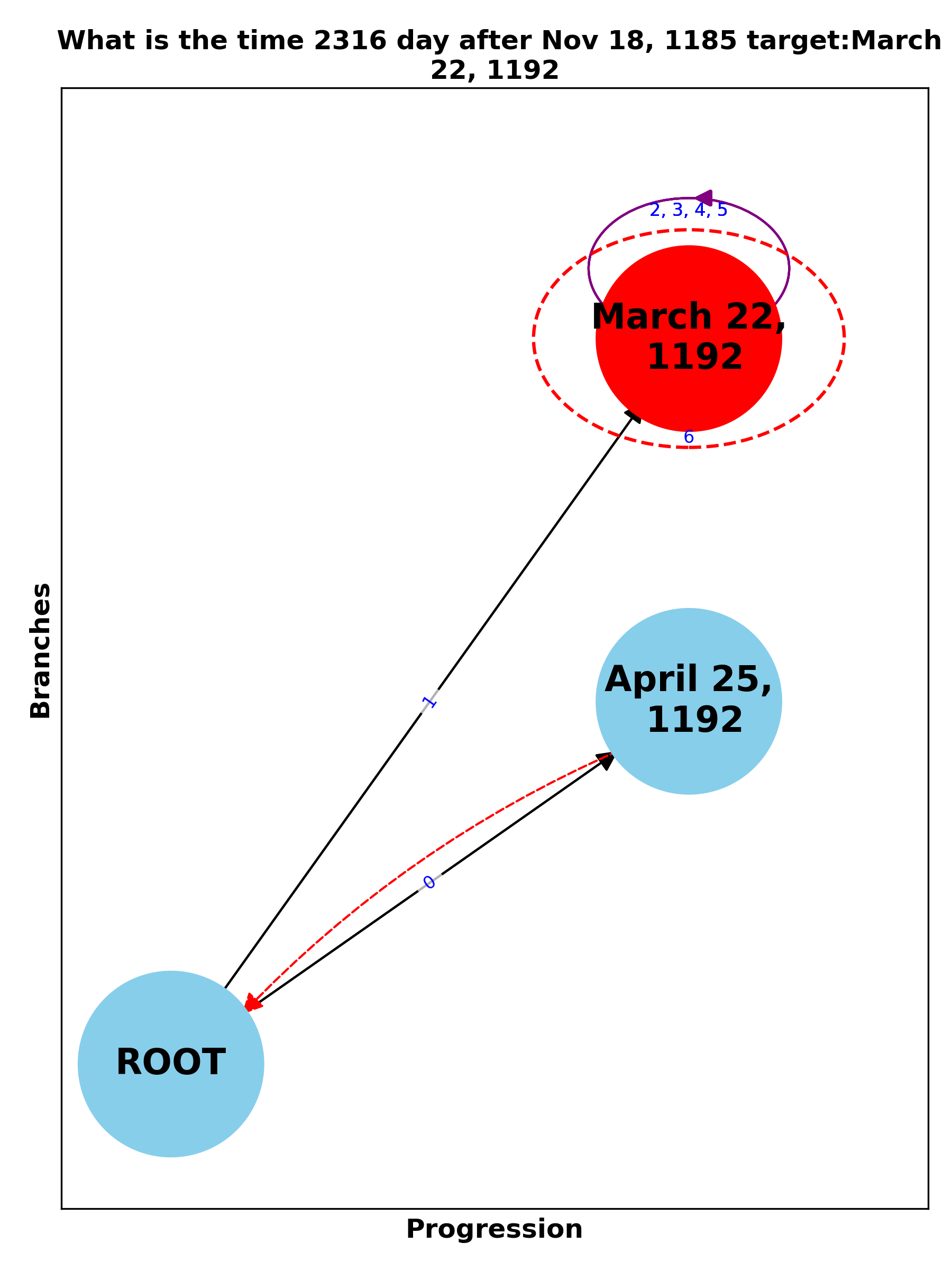}
    \caption{Individual thought progression graph of \texttt{Qwen3-32B} model on a sampled date arithmetic (temporal-L3) query.}
    \label{fig:progression_4}
\end{figure}

\Cref{fig:progression_4} displays the thought progression of the \texttt{Qwen3-32B} model on a date arithmetic (temporal-L3) query.

\begin{figure}[t]
    \centering
    \includegraphics[width=0.5\textwidth]{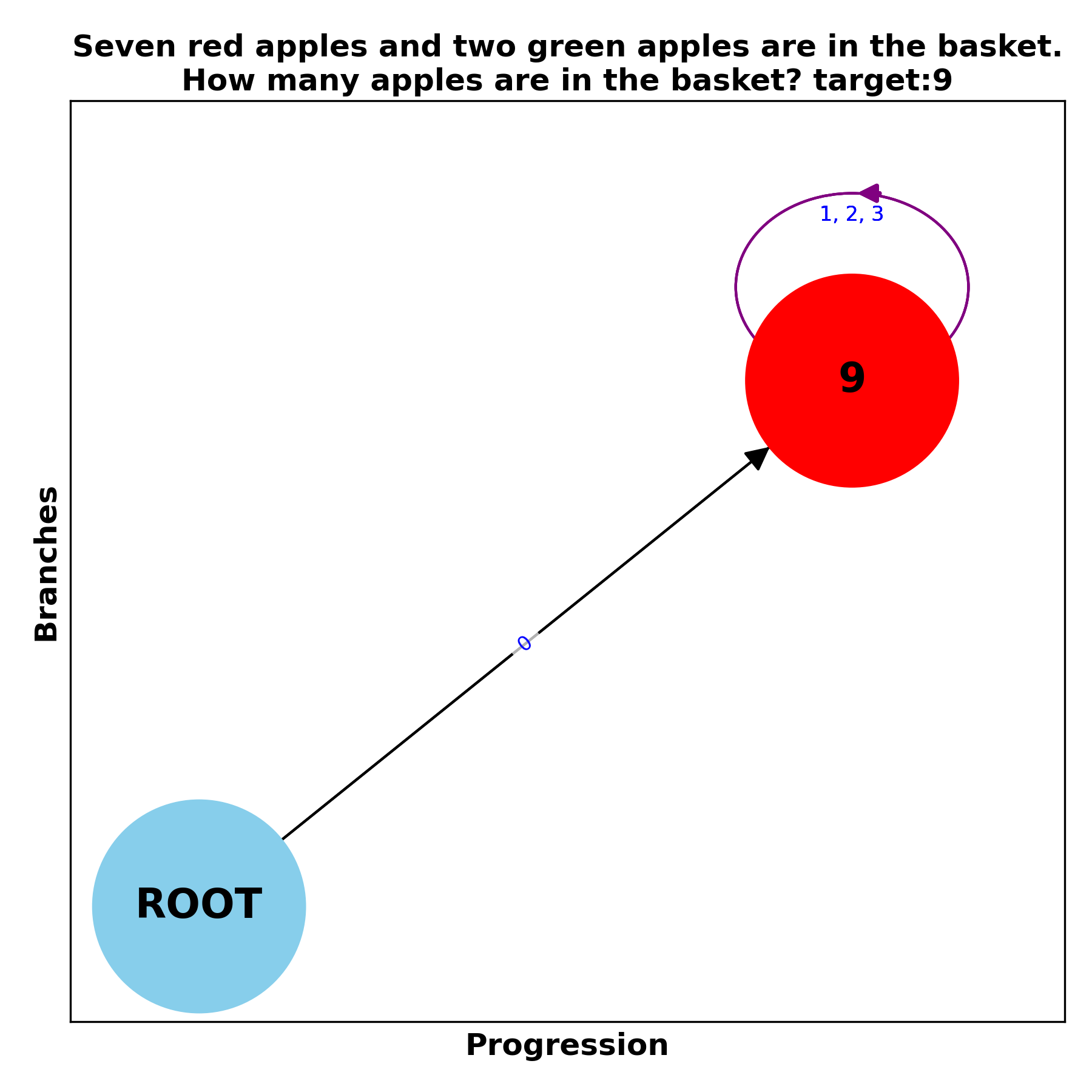}
    \caption{Individual thought progression graph of \texttt{Qwen3-32B} model on a sampled grade-1 arithmetic query.}
    \label{fig:progression_5}
\end{figure}

\Cref{fig:progression_5} displays the thought progression of the \texttt{Qwen3-32B} model on a grade-1 arithmetic query.

\begin{figure}[t]
    \centering
    \includegraphics[width=0.5\textwidth]{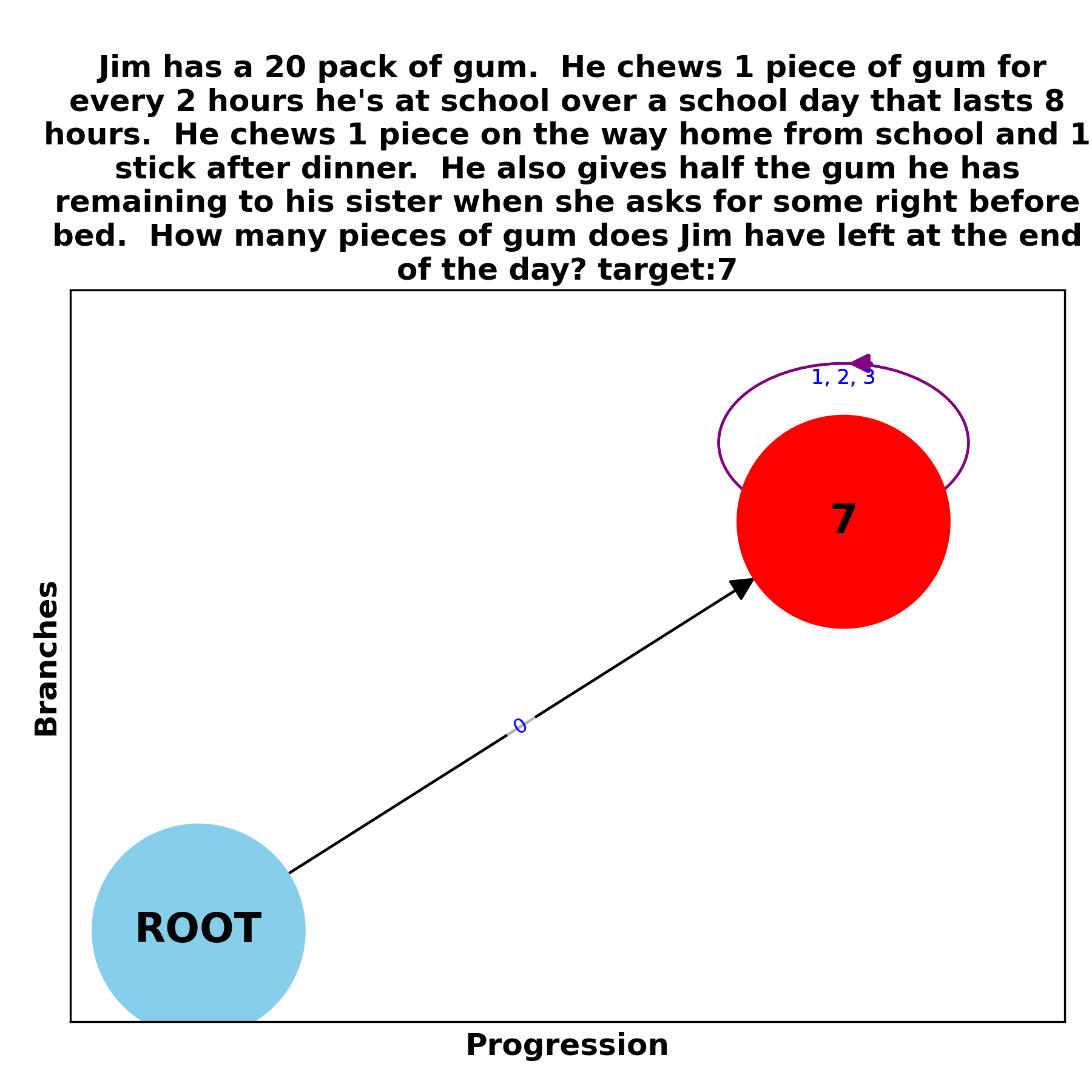}
    \caption{Individual thought progression graph of \texttt{Qwen3-32B} model on a sampled GSM8k query.}
    \label{fig:progression_6}
\end{figure}

\Cref{fig:progression_6} displays the thought progression of the \texttt{Qwen3-32B} model on a GSM8k query.

\begin{figure}[t]
    \centering
    \includegraphics[width=0.5\textwidth]{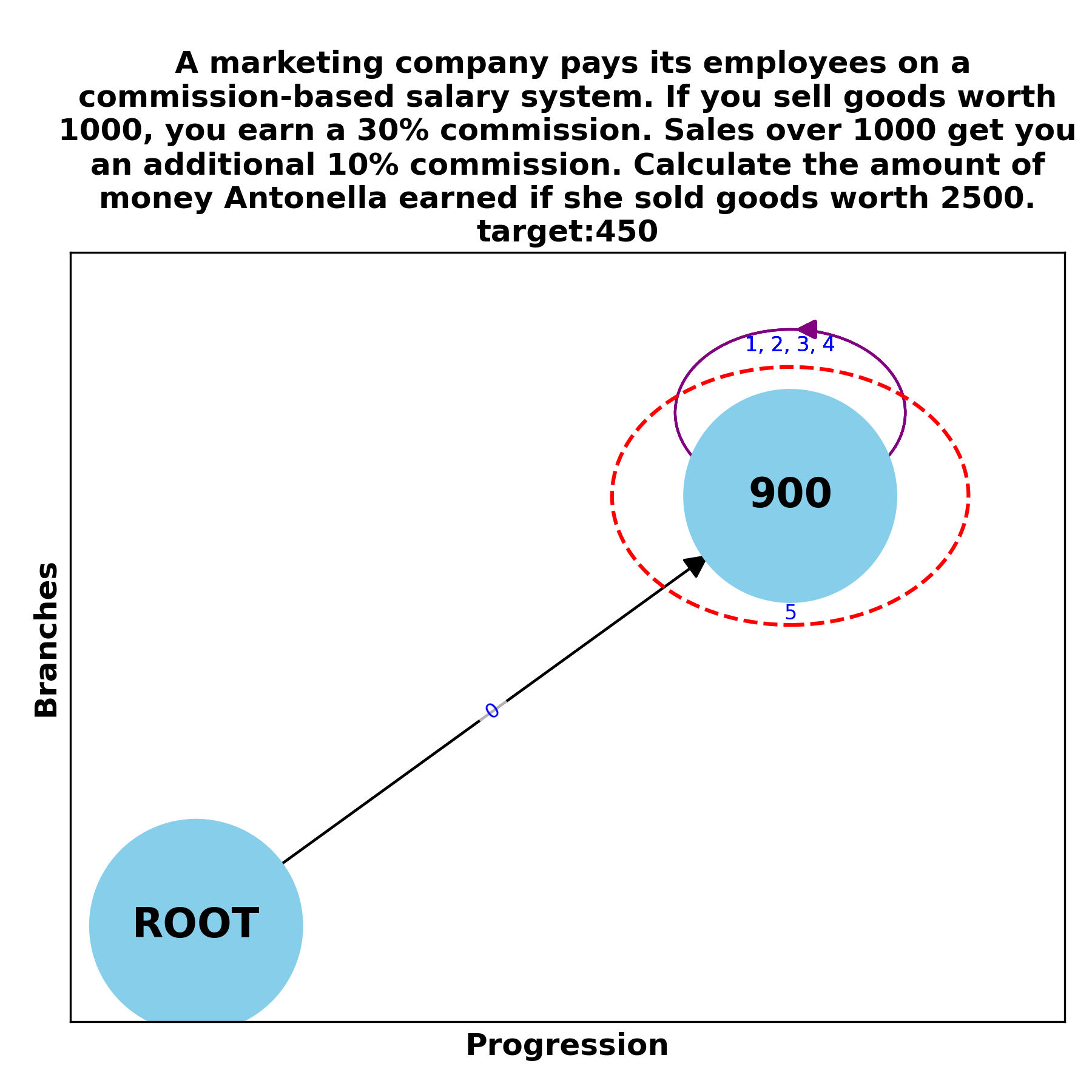}
    \caption{Individual thought progression graph of \texttt{Qwen3-32B} model on a sampled GSM8k query.}
    \label{fig:progression_7}
\end{figure}

\Cref{fig:progression_7} displays the thought progression of the \texttt{Qwen3-32B} model on another GSM8k query.

\begin{figure}[t]
    \centering
    \includegraphics[width=0.5\textwidth]{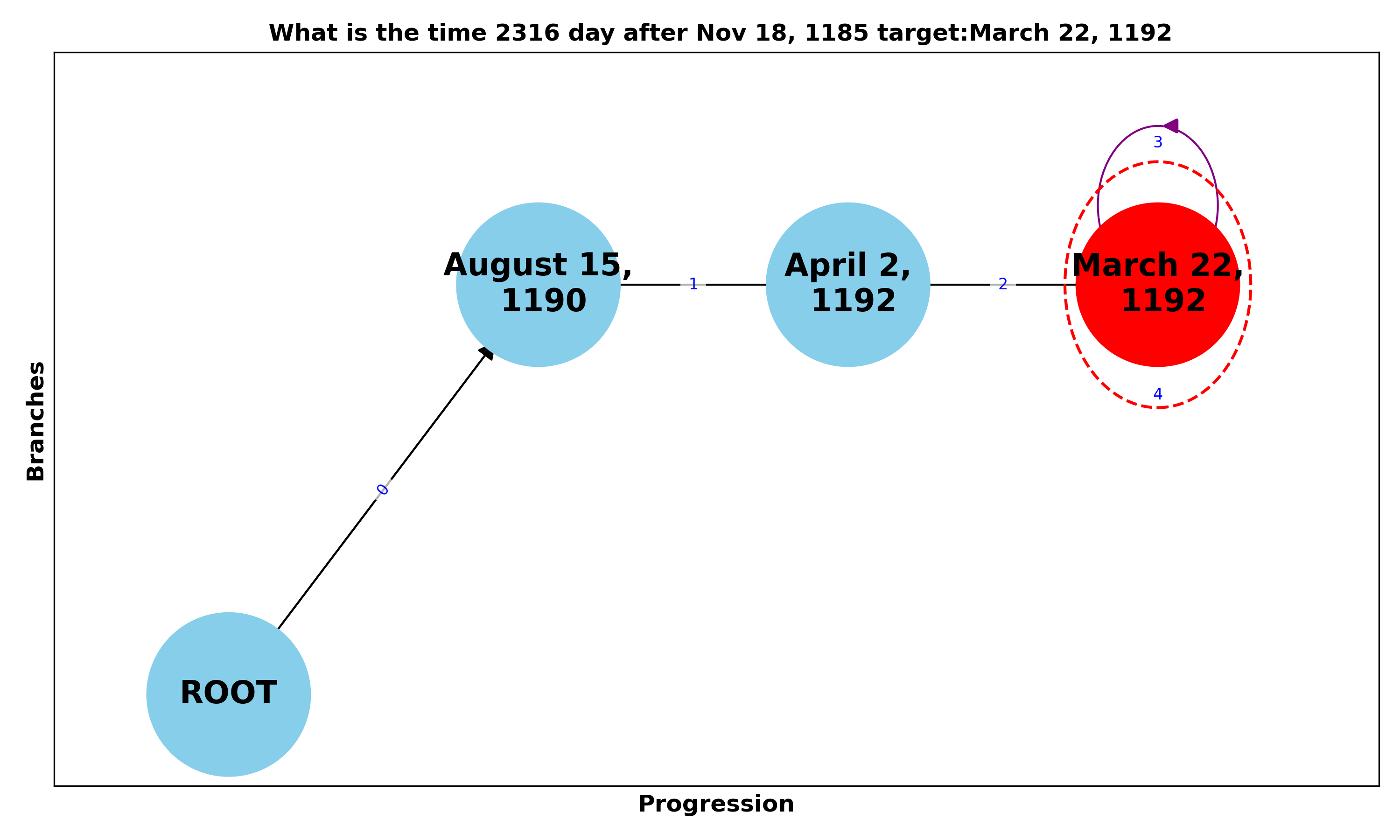}
    \caption{Individual thought progression graph of \texttt{R1-Distill-Llama-70B} model on a sampled date arithmetic (temporal-L3) query.}
    \label{fig:progression_8}
\end{figure}

\Cref{fig:progression_8} displays the thought progression of the \texttt{R1-Distill-Llama-70B} model on a date arithmetic (temporal-L3) query.

\begin{figure}[t]
    \centering
    \includegraphics[width=0.5\textwidth]{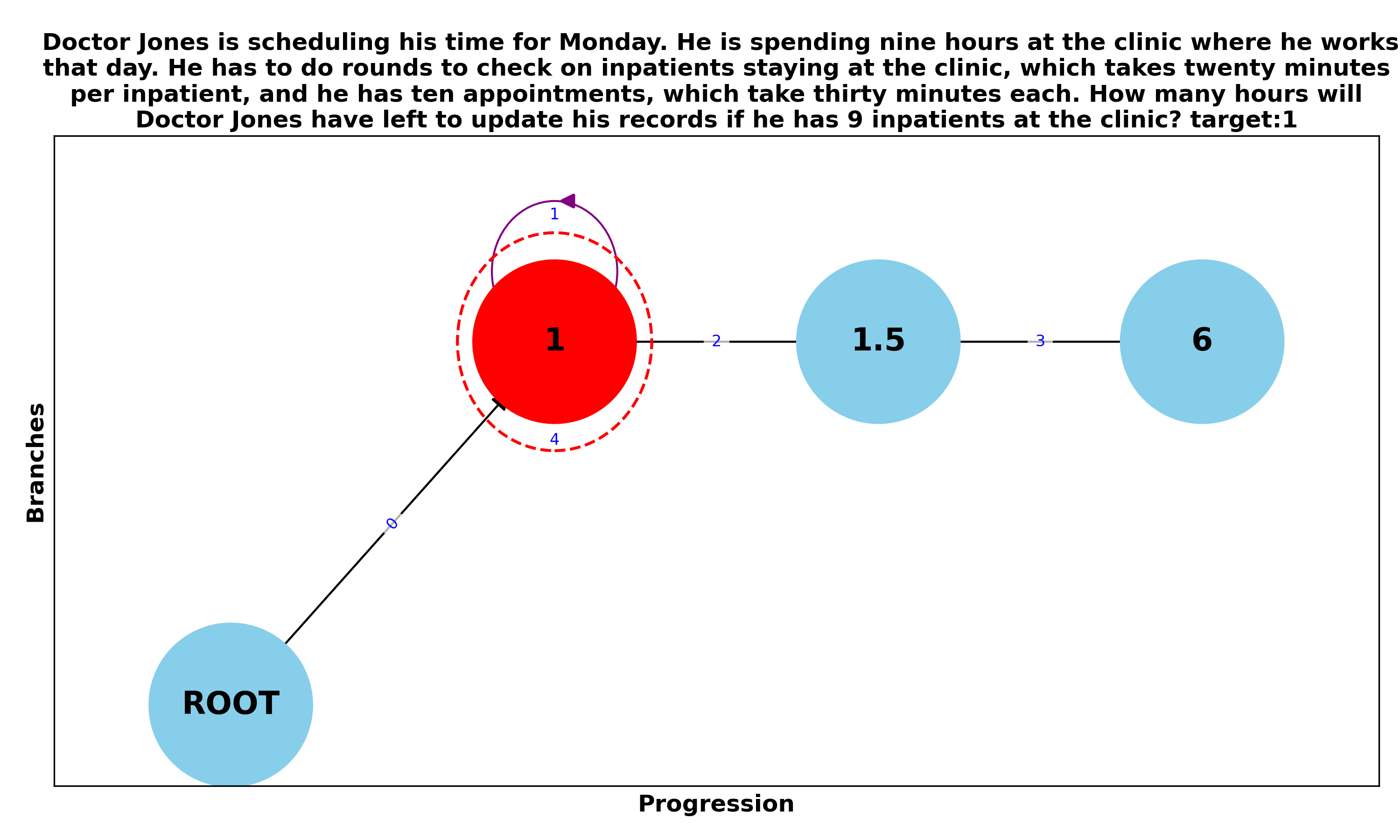}
    \caption{Individual thought progression graph of \texttt{R1-Distill-Llama-70B} model on a sampled GSM8k query.}
    \label{fig:progression_9}
\end{figure}

\Cref{fig:progression_9} displays the thought progression of the \texttt{R1-Distill-Llama-70B} model on a GSM8k query.

\begin{figure}[t]
    \centering
    \includegraphics[width=0.5\textwidth]{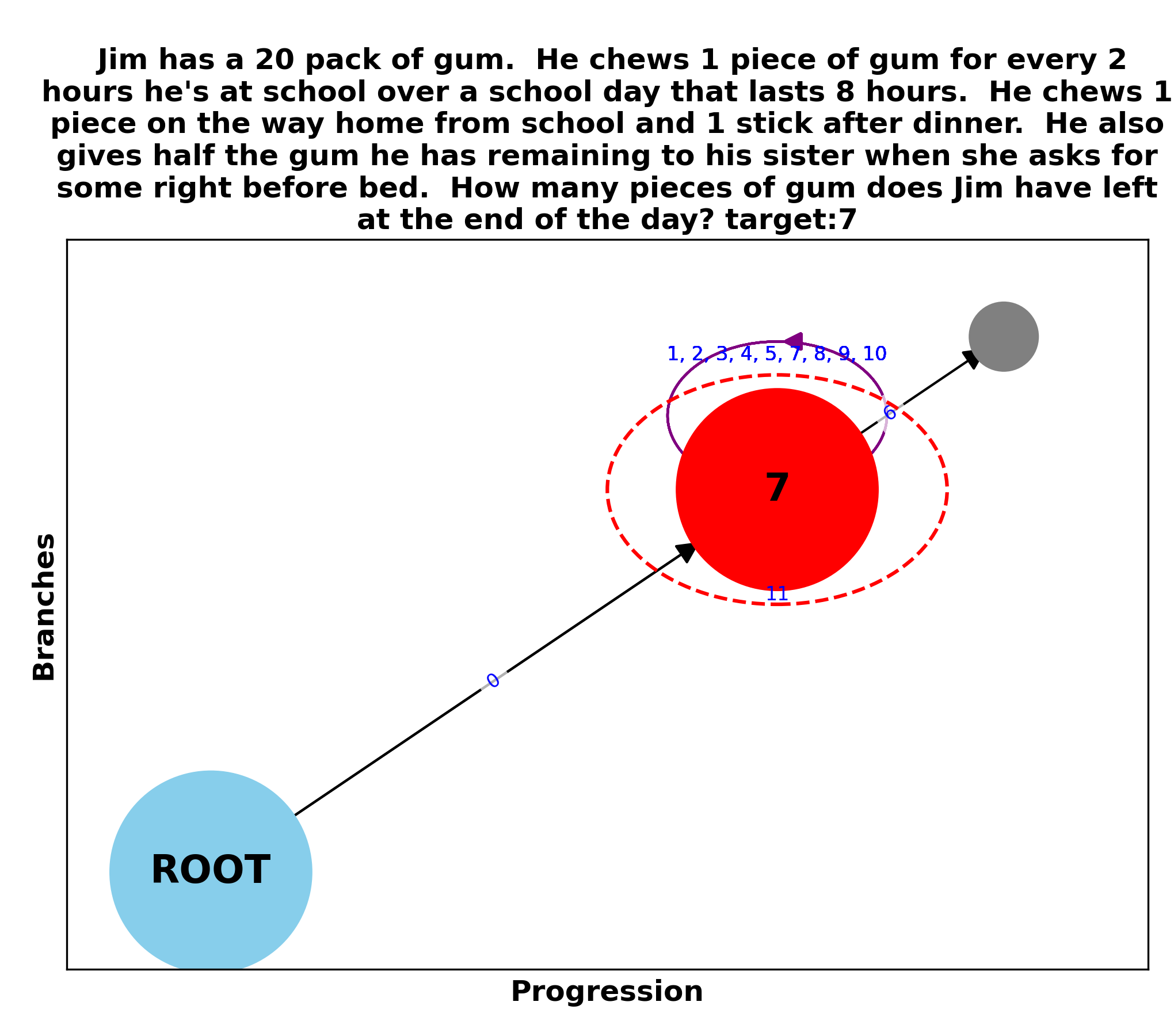}
    \caption{Individual thought progression graph of \texttt{Qwen3-235B-A22B} model on a sampled GSM8k query.}
    \label{fig:progression_10}
\end{figure}

\Cref{fig:progression_10} displays the thought progression of the \texttt{Qwen3-235B-A22B} model on a GSM8k query.

\begin{figure}[t]
    \centering
    \includegraphics[width=0.5\textwidth]{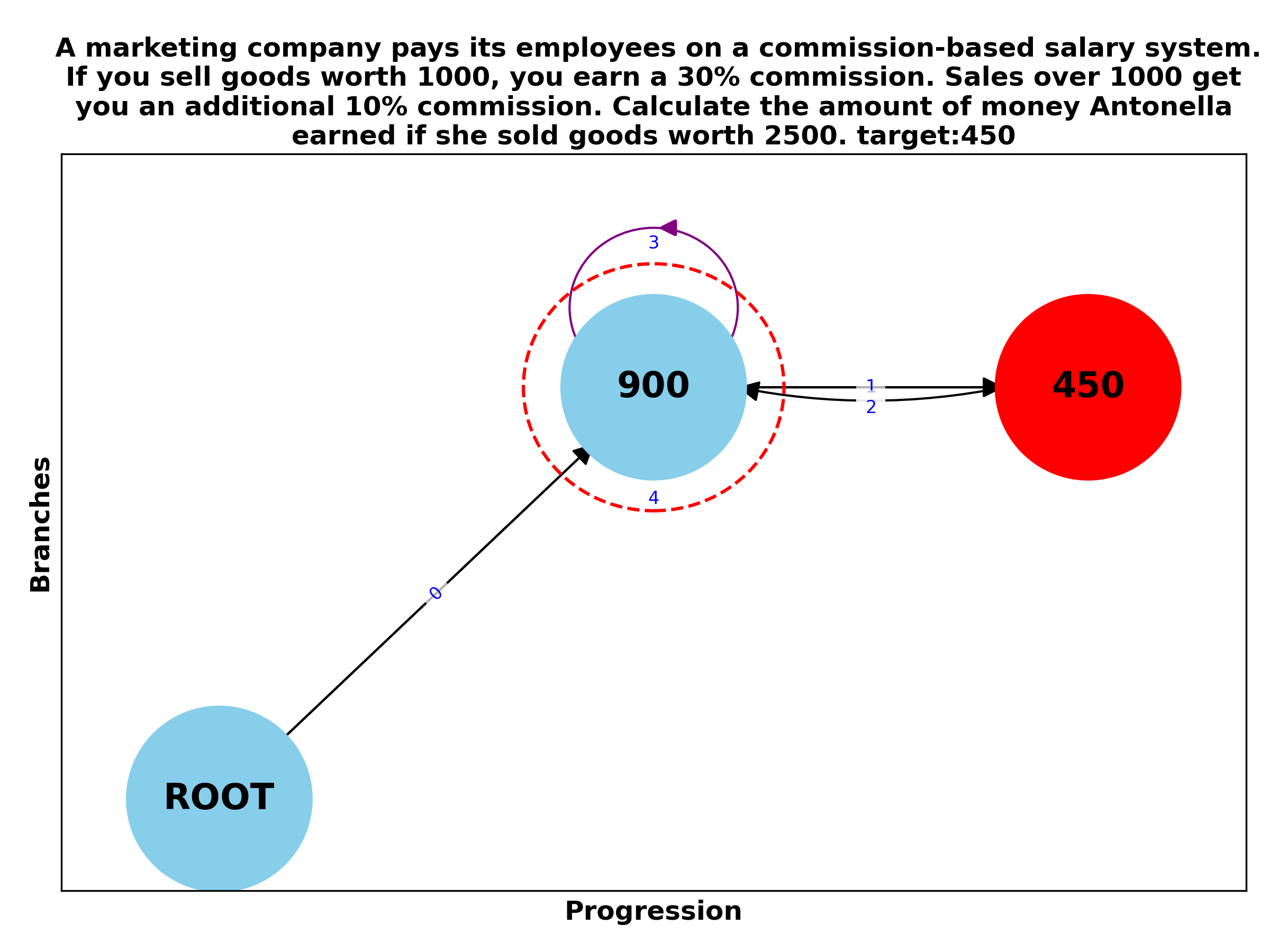}
    \caption{Individual thought progression graph of \texttt{Qwen3-235B-A22B}model on a sampled GSM8k query.}
    \label{fig:progression_11}
\end{figure}

\Cref{fig:progression_11} displays the thought progression of the \texttt{Qwen3-235B-A22B} model on another GSM8k query.

\begin{figure}[t]
    \centering
    \includegraphics[width=0.5\textwidth]{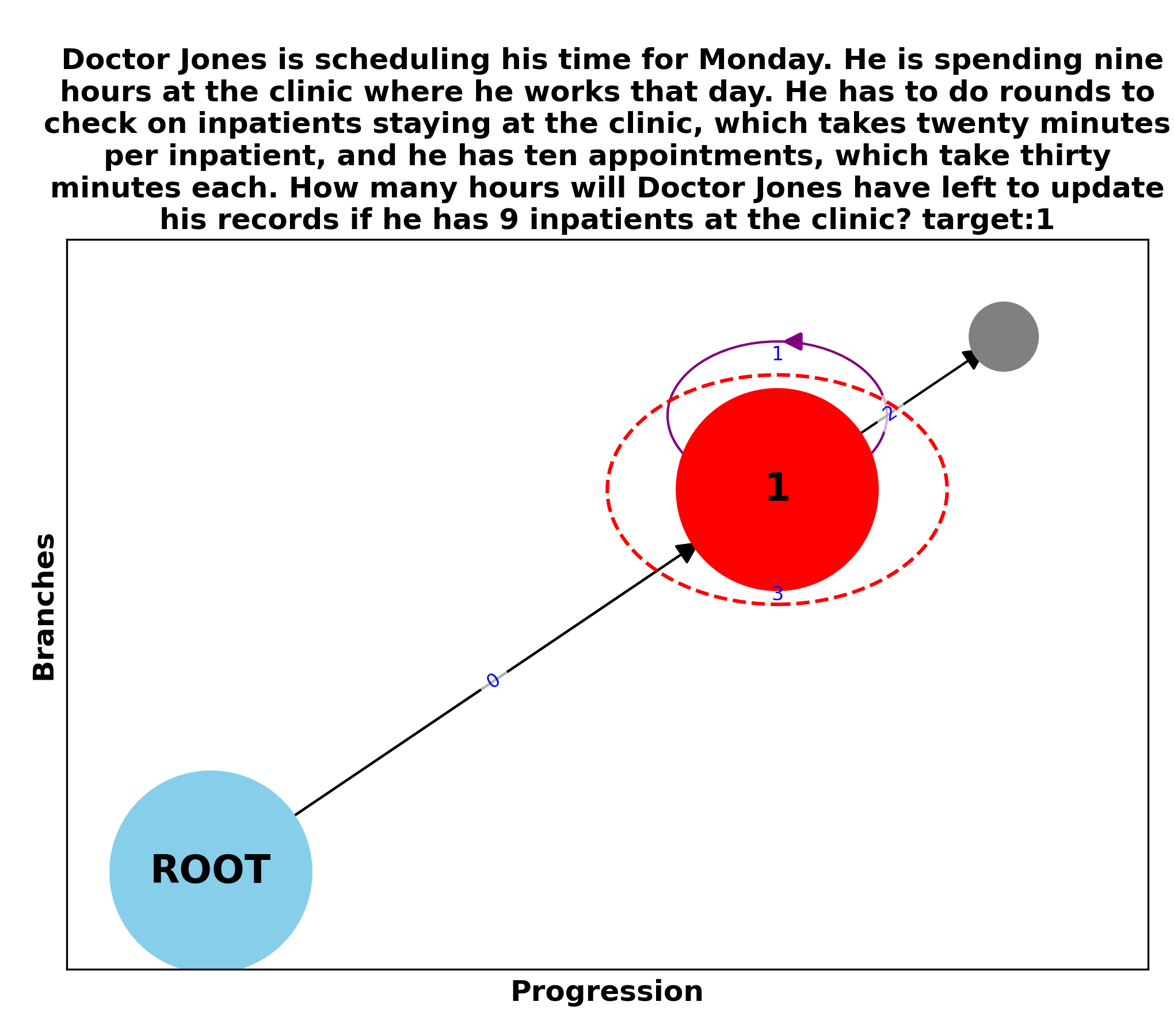}
    \caption{Individual thought progression graph of \texttt{Qwen3-235B-A22B} model on a sampled GSM8k query. This is the graphical representation of~\Cref{fig:gsm8k_parsed_example_part1} and \ref{fig:gsm8k_parsed_example_part2}.}
    \label{fig:progression_12}
\end{figure}

\begin{figure*}[t]
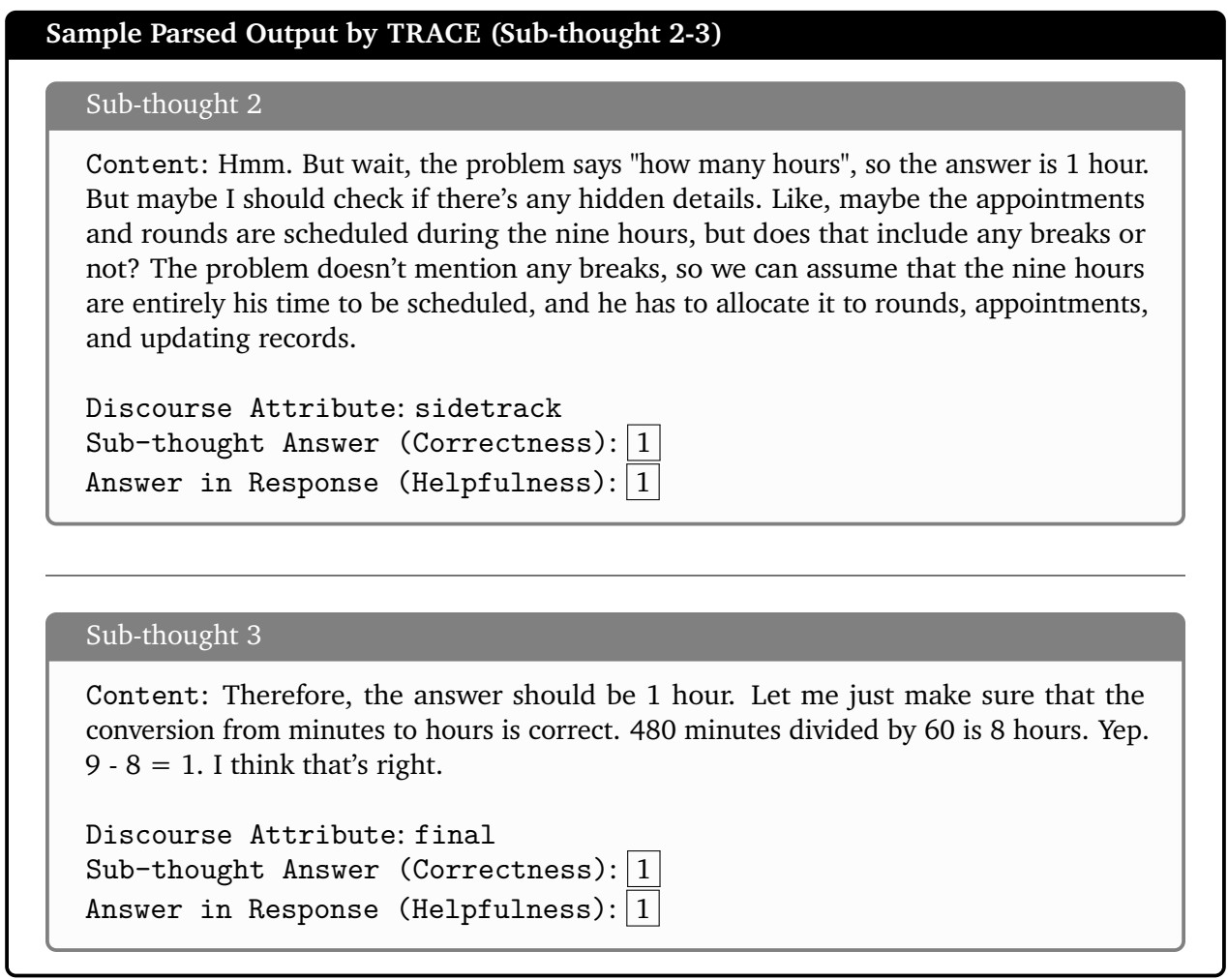

\begin{tcolorbox}[colback=white, colframe=black, fonttitle=\bfseries, title=Sample Parsed Output by \framework (Sub-thought 2-3)]

% ---------- Sub-thought 2 ----------
\begin{tcolorbox}[colback=white!98!gray, colframe=black!50, title=Sub-thought 2]
\texttt{Content}: Hmm. But wait, the problem says "how many hours", so the answer is 1 hour. But maybe I should check if there's any hidden details. Like, maybe the appointments and rounds are scheduled during the nine hours, but does that include any breaks or not? The problem doesn't mention any breaks, so we can assume that the nine hours are entirely his time to be scheduled, and he has to allocate it to rounds, appointments, and updating records. \\

\texttt{Discourse Attribute}: \texttt{sidetrack} \\
\texttt{Sub-thought Answer (Correctness)}: $\boxed{1}$ \\
\texttt{Answer in Response (Helpfulness)}: $\boxed{1}$
\end{tcolorbox}

% ---------- Separator ----------
\vspace{0.3em}
\noindent\hrulefill
\vspace{0.3em}

% ---------- Sub-thought 3 ----------
\begin{tcolorbox}[colback=white!98!gray, colframe=black!50, title=Sub-thought 3]
\texttt{Content}: Therefore, the answer should be 1 hour. Let me just make sure that the conversion from minutes to hours is correct. 480 minutes divided by 60 is 8 hours. Yep. 9 - 8 = 1. I think that's right. \\

\texttt{Discourse Attribute}: \texttt{final} \\
\texttt{Sub-thought Answer (Correctness)}: $\boxed{1}$ \\
\texttt{Answer in Response (Helpfulness)}: $\boxed{1}$
\end{tcolorbox}

\end{tcolorbox}
\caption{Sample parsed output (sub-thoughts 2-3) by \framework on the \texttt{\texttt{Qwen3-235B-A22B}}-generated response to a GSM8K prompt. Sub-thought 2 is marked as \textit{sidetrack} since the model is rambling and exploring supplementary 
information which does not directly contribute to solve the input problem. The graphical representation of \textit{sidetrack} is a tiny, gray bubble as shown in~\Cref{fig:progression_12}.}
\label{fig:gsm8k_parsed_example_part2}
\end{figure*}

\Cref{fig:progression_12} displays the thought progression of the \texttt{Qwen3-235B-A22B} model on a GSM8k query, which is the graphical representation of~\Cref{fig:gsm8k_parsed_example_part1} and \ref{fig:gsm8k_parsed_example_part2}.

\section{More Plots for Induced Thought Progression Patterns}
\subsection{More Plots of Explorer Thought Progression Pattern}
\label{appx:more_plots_explorer}
\begin{figure}[t]
    \centering
    \includegraphics[width=0.5\textwidth]{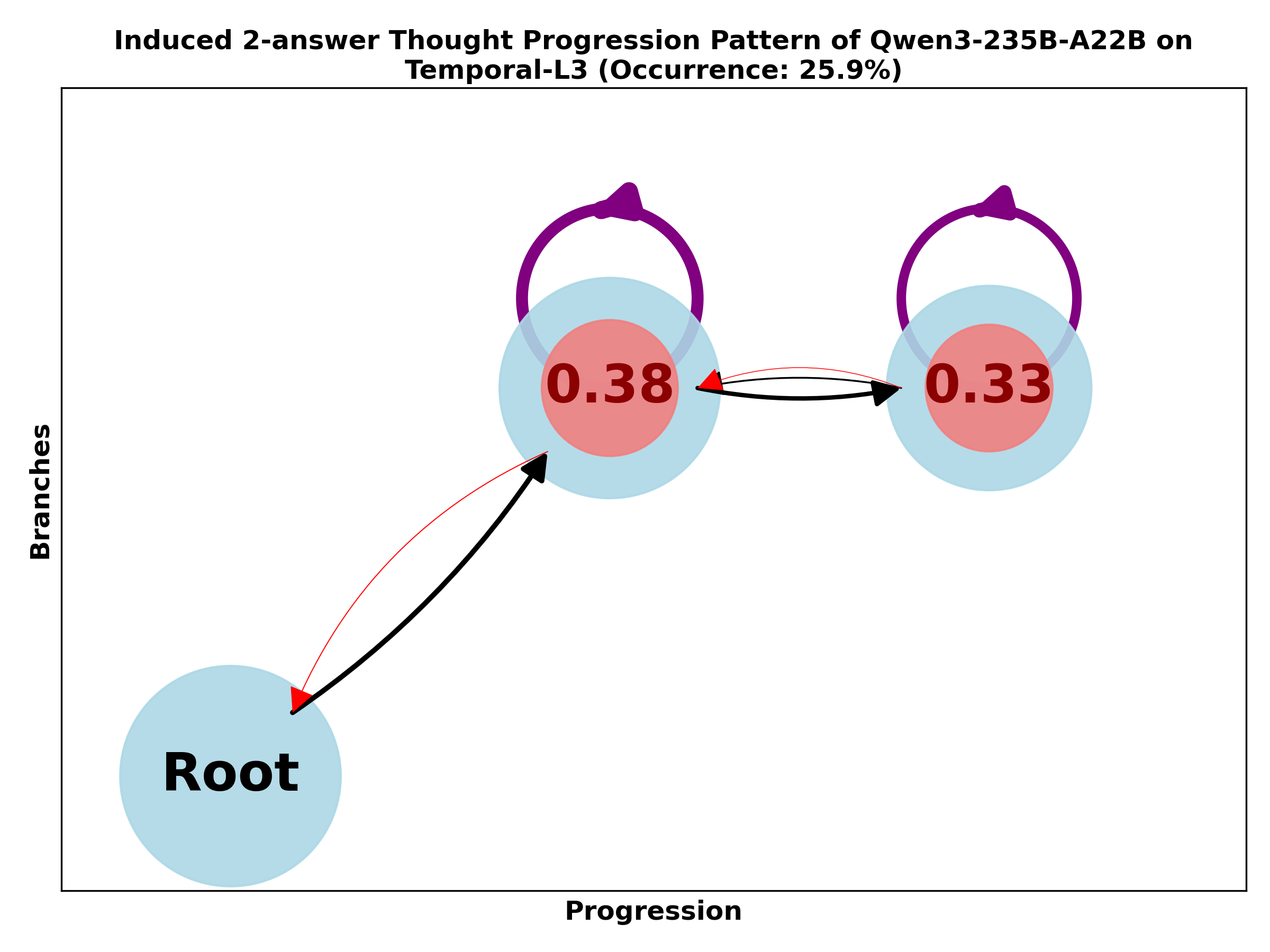}
    \caption{The \textbf{Explorer} thought progression pattern (2 distinct answer case). This is a trivial case.}
    \label{fig:explorer_pattern_group2}
\end{figure}

\begin{figure}[t]
    \centering
    \includegraphics[width=0.5\textwidth]{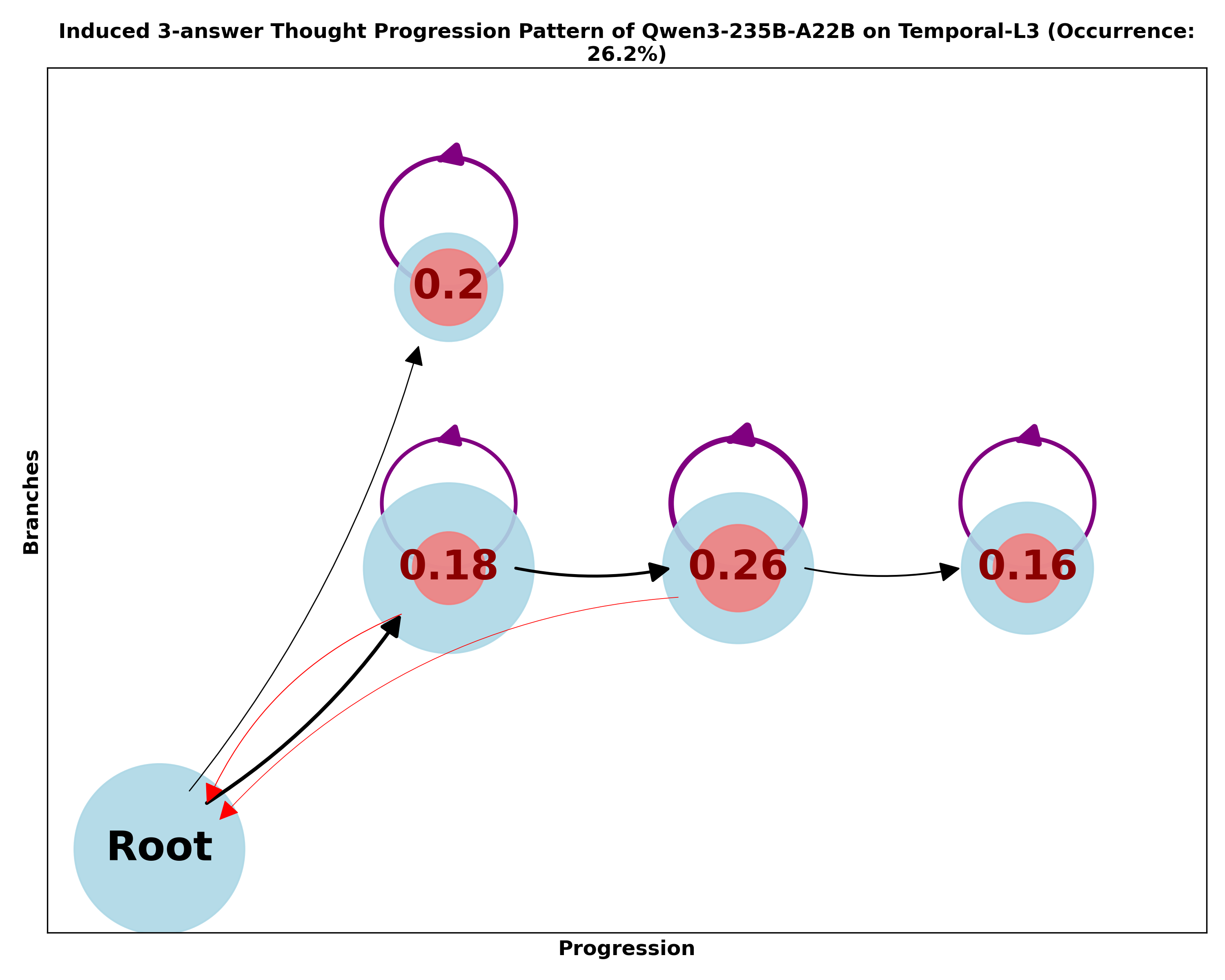}
    \caption{The \textbf{Explorer} thought progression pattern (3 distinct answer case).}
    \label{fig:explorer_pattern_group3}
\end{figure}

\begin{figure}[t]
    \centering
    \includegraphics[width=0.5\textwidth]{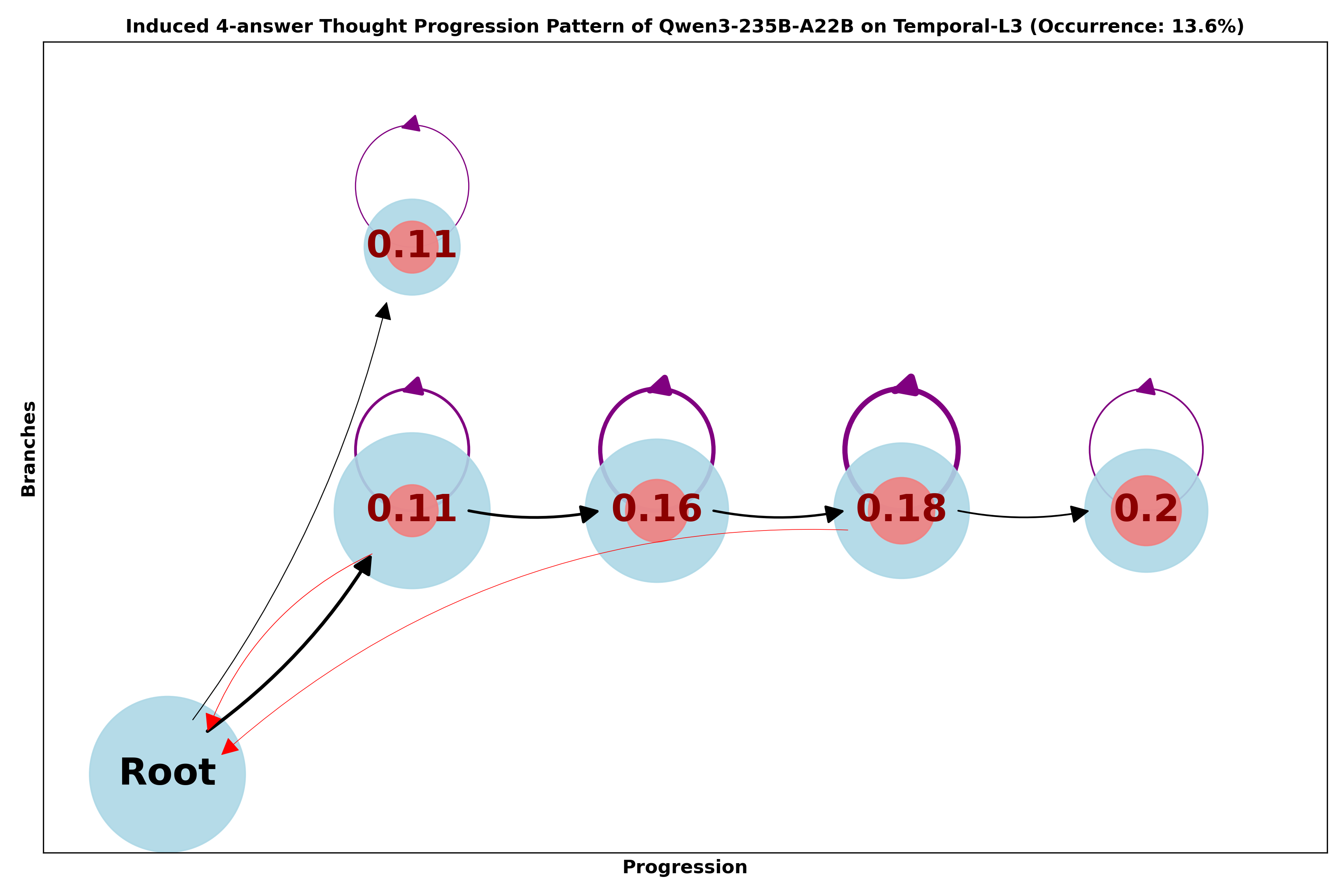}
    \caption{The \textbf{Explorer} thought progression pattern (4 distinct answer case).}
    \label{fig:explorer_pattern_group4}
\end{figure}

\begin{figure}[t]
    \centering
    \includegraphics[width=0.5\textwidth]{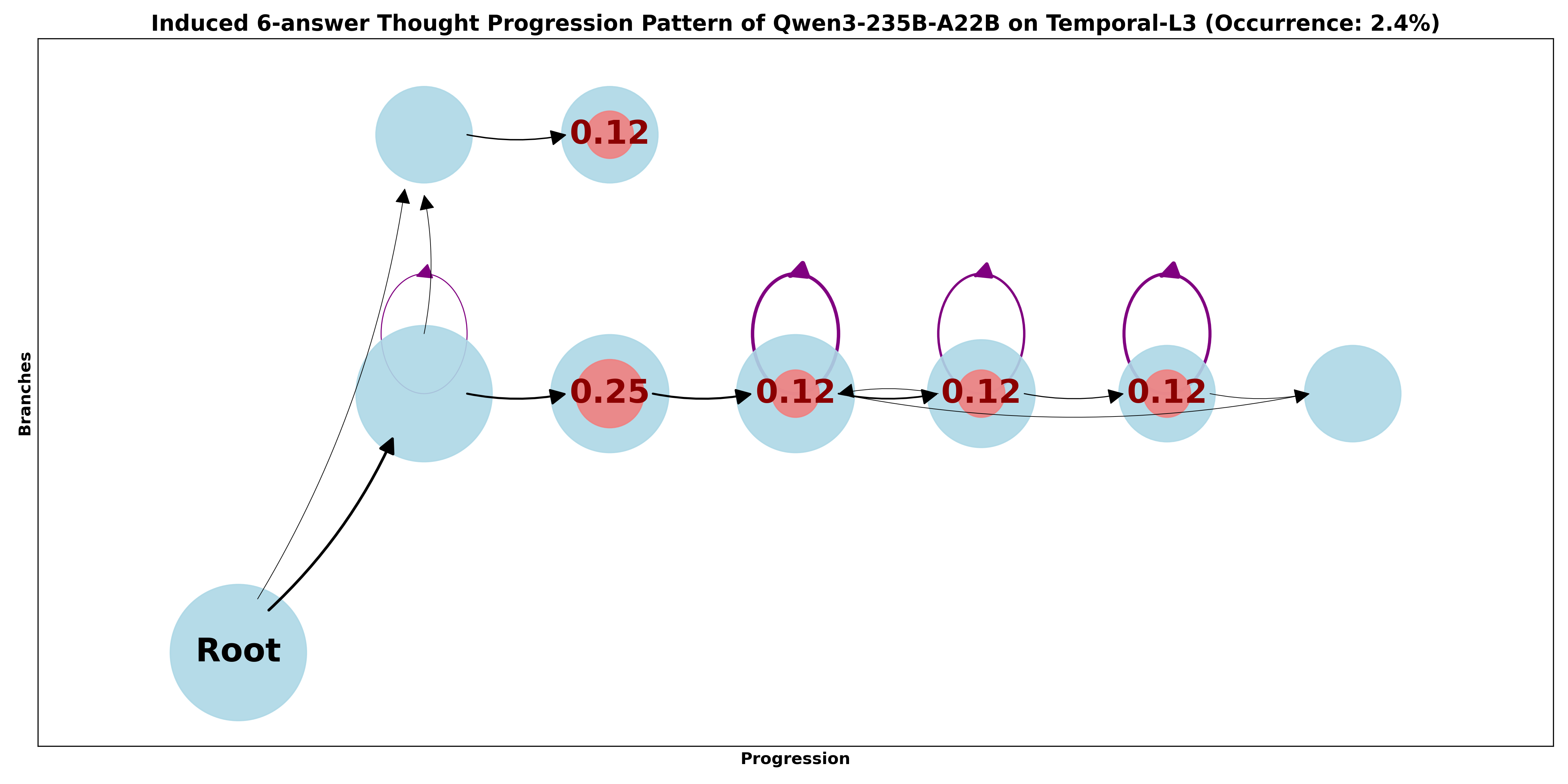}
    \caption{The \textbf{Explorer} thought progression pattern (6 distinct answer case). }
    \label{fig:explorer_pattern_group6}
\end{figure}

\begin{figure}[t]
    \centering
    \includegraphics[width=0.5\textwidth]{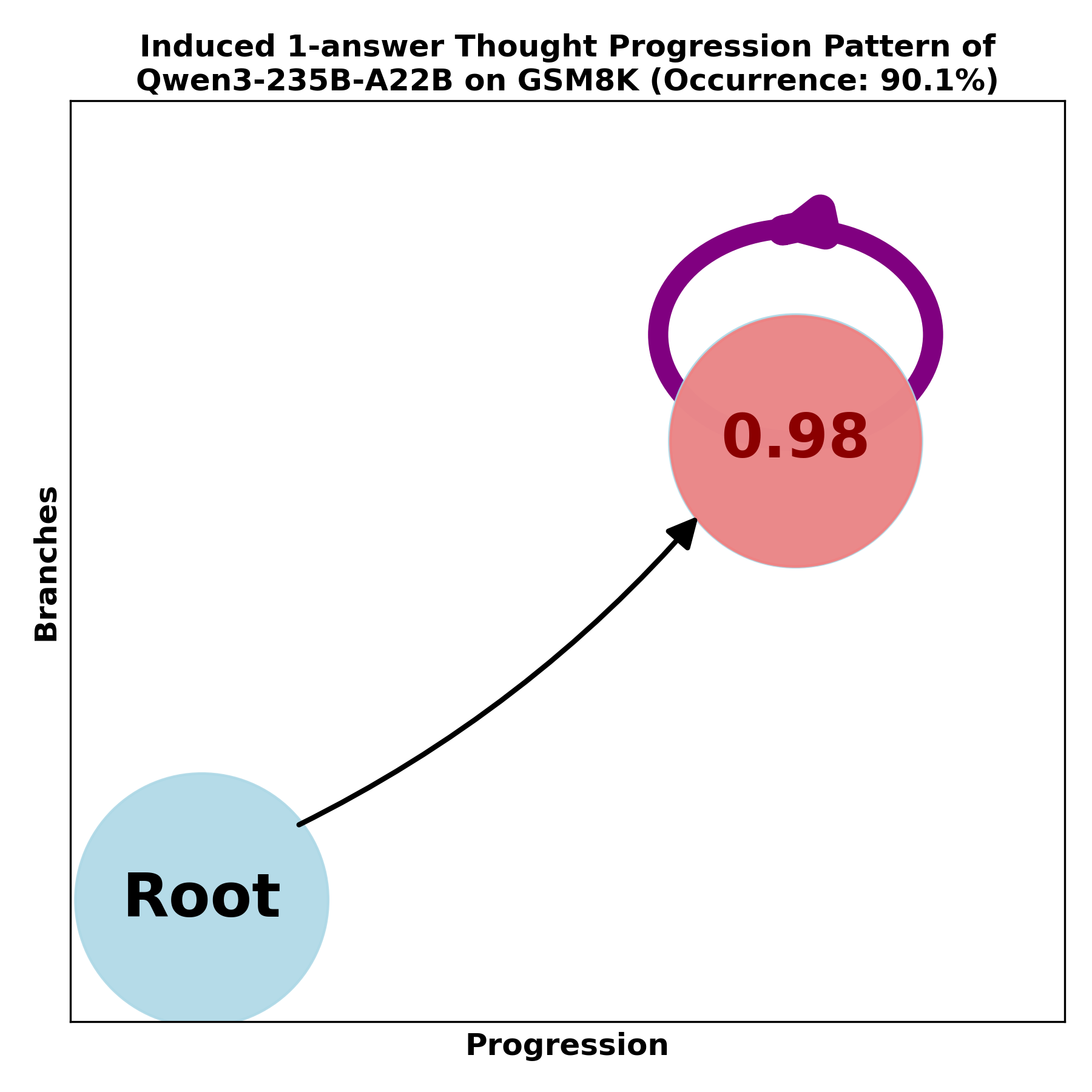}
    \caption{The \textbf{Explorer} thought progression pattern (1 distinct answer case). This is a trivial case.}
    \label{fig:explorer_pattern_group1_gsm8k}
\end{figure}

\begin{figure}[t]
    \centering
    \includegraphics[width=0.5\textwidth]{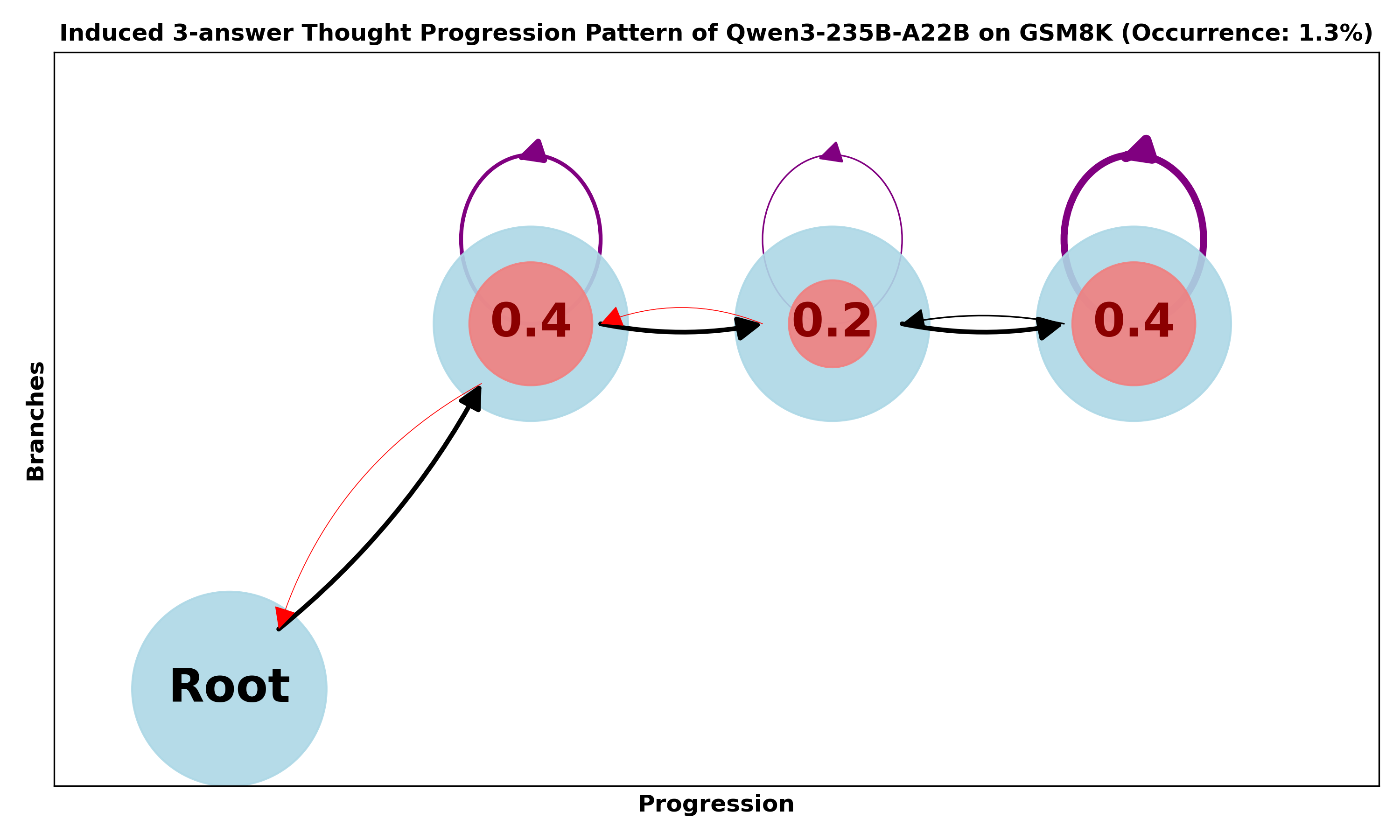}
    \caption{The \textbf{Explorer} thought progression pattern (3 distinct answer case).}
    \label{fig:explorer_pattern_group3_gsm8k}
\end{figure}

\Cref{fig:explorer_pattern_group2} (trivial case),~\ref{fig:explorer_pattern_group3},~\ref{fig:explorer_pattern_group4},~\ref{fig:explorer_pattern_group6},~\ref{fig:explorer_pattern_group1_gsm8k} (trivial case) and~\ref{fig:explorer_pattern_group3_gsm8k} present more plots of the \textbf{Explorer} thought progression pattern. 

\subsection{More Plots of Late Landing Thought Progression Pattern}
\label{appx:more_plots_late_landing}
\begin{figure}[t]
    \centering
    \includegraphics[width=0.5\textwidth]{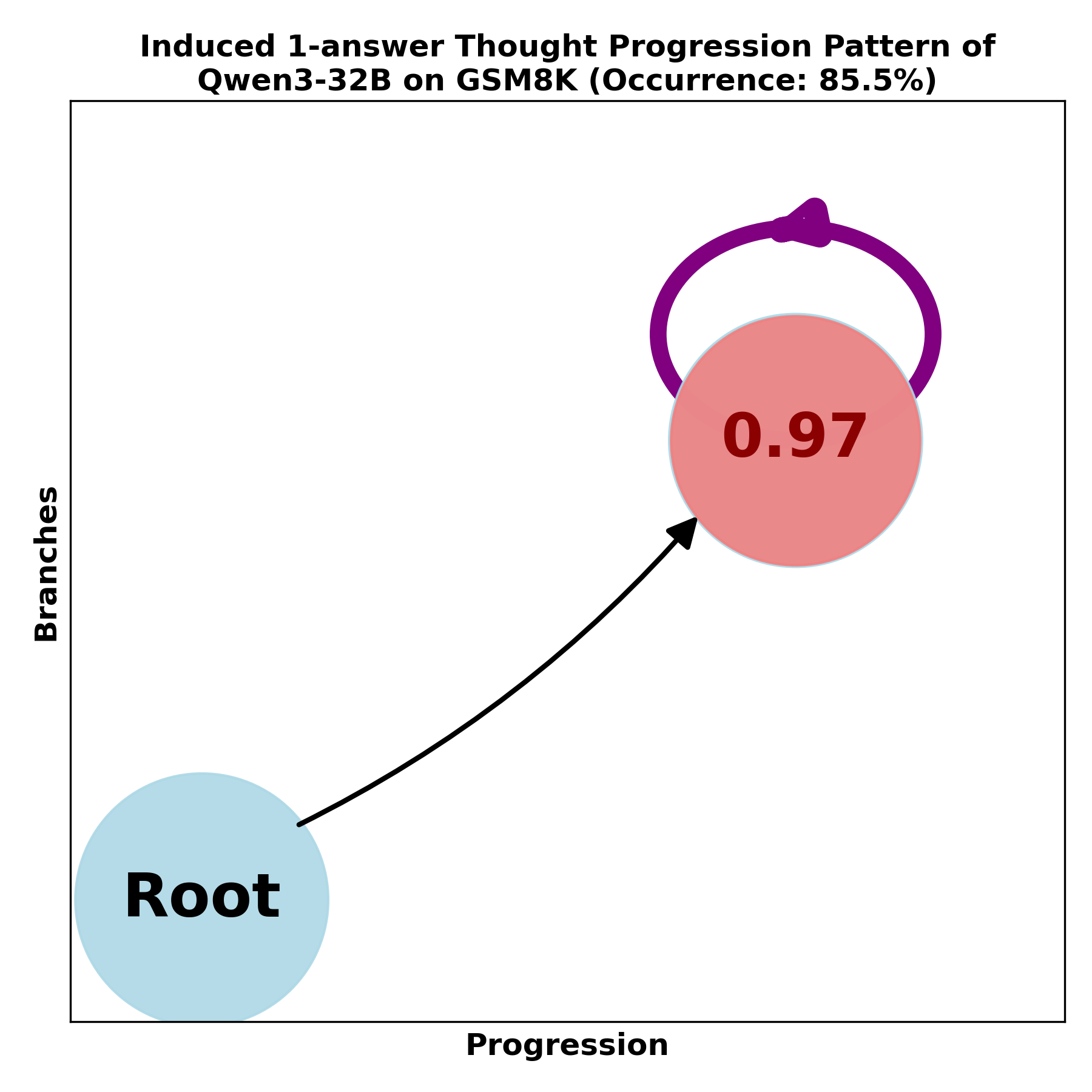}
    \caption{The \textbf{Late Landing} thought progression pattern (1 distinct answer case). This is a trivial case.}
    \label{fig:late_lander_pattern_group1_gsm8k}
\end{figure}

\begin{figure}[t]
    \centering
    \includegraphics[width=0.5\textwidth]{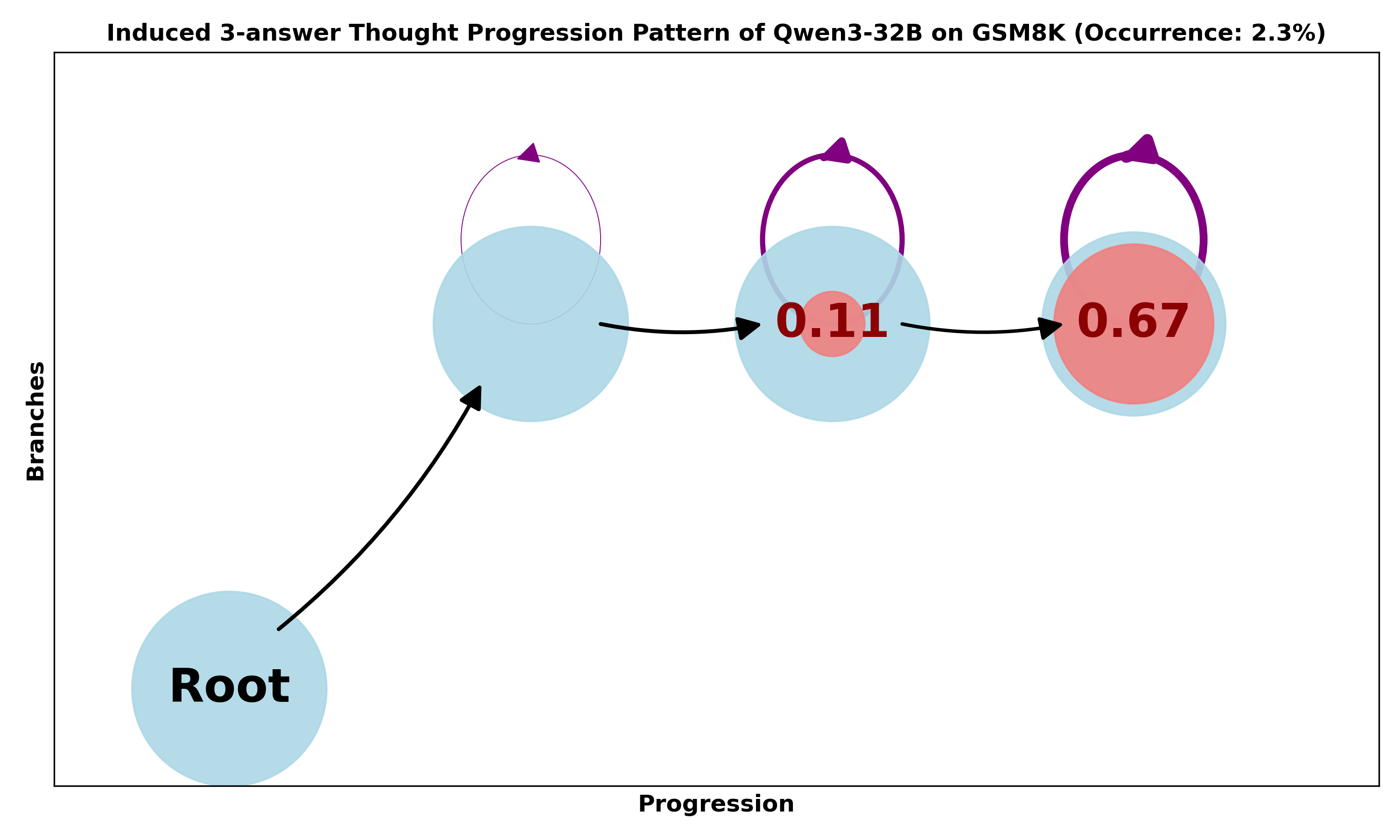}
    \caption{The \textbf{Late Landing} thought progression pattern (3 distinct answer case).}
    \label{fig:late_lander_pattern_group3_gsm8k}
\end{figure}

\begin{figure}[t]
    \centering
    \includegraphics[width=0.5\textwidth]{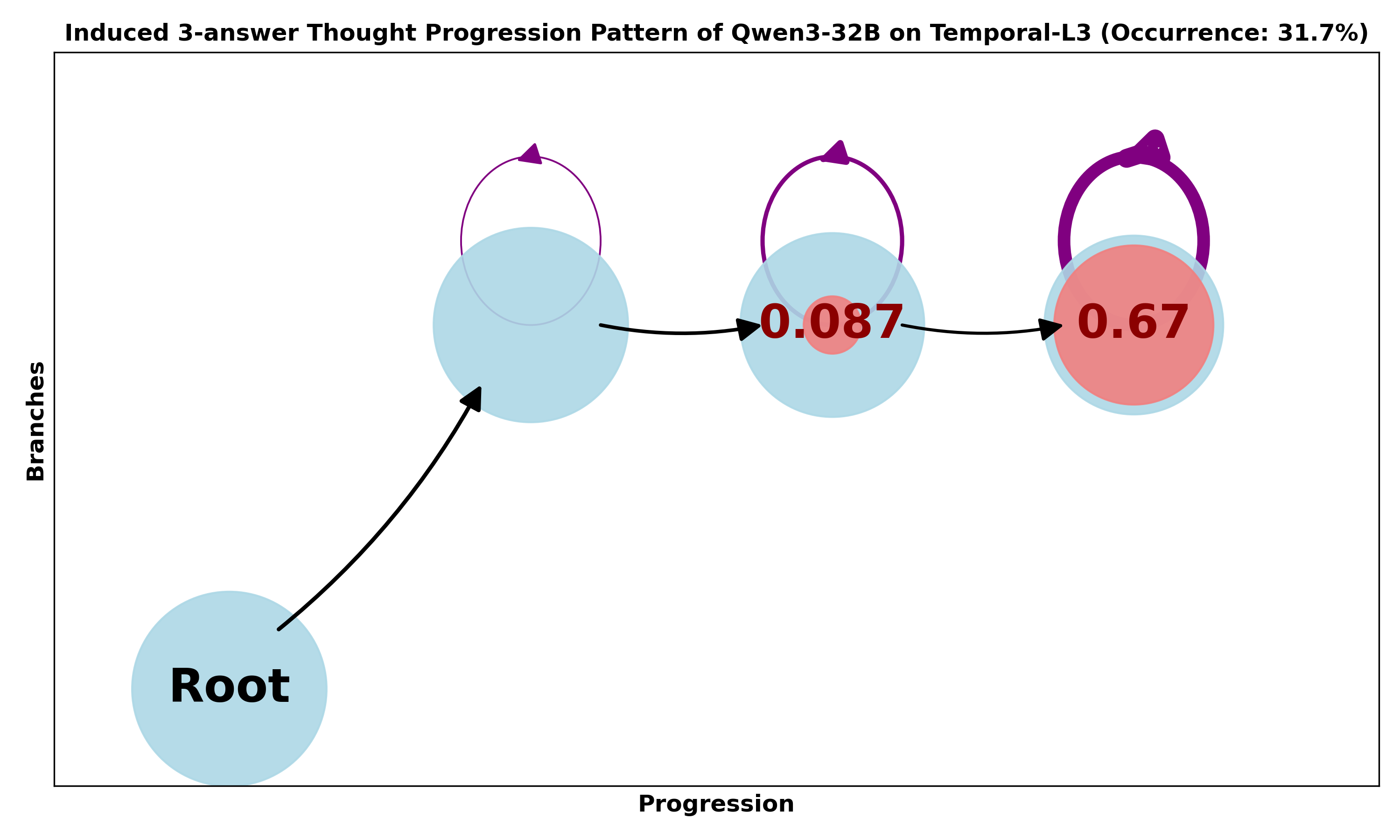}
    \caption{The \textbf{Late Landing} thought progression pattern (3 distinct answer case).}
    \label{fig:late_lander_pattern_group3}
\end{figure}

\begin{figure}[t]
    \centering
    \includegraphics[width=0.5\textwidth]{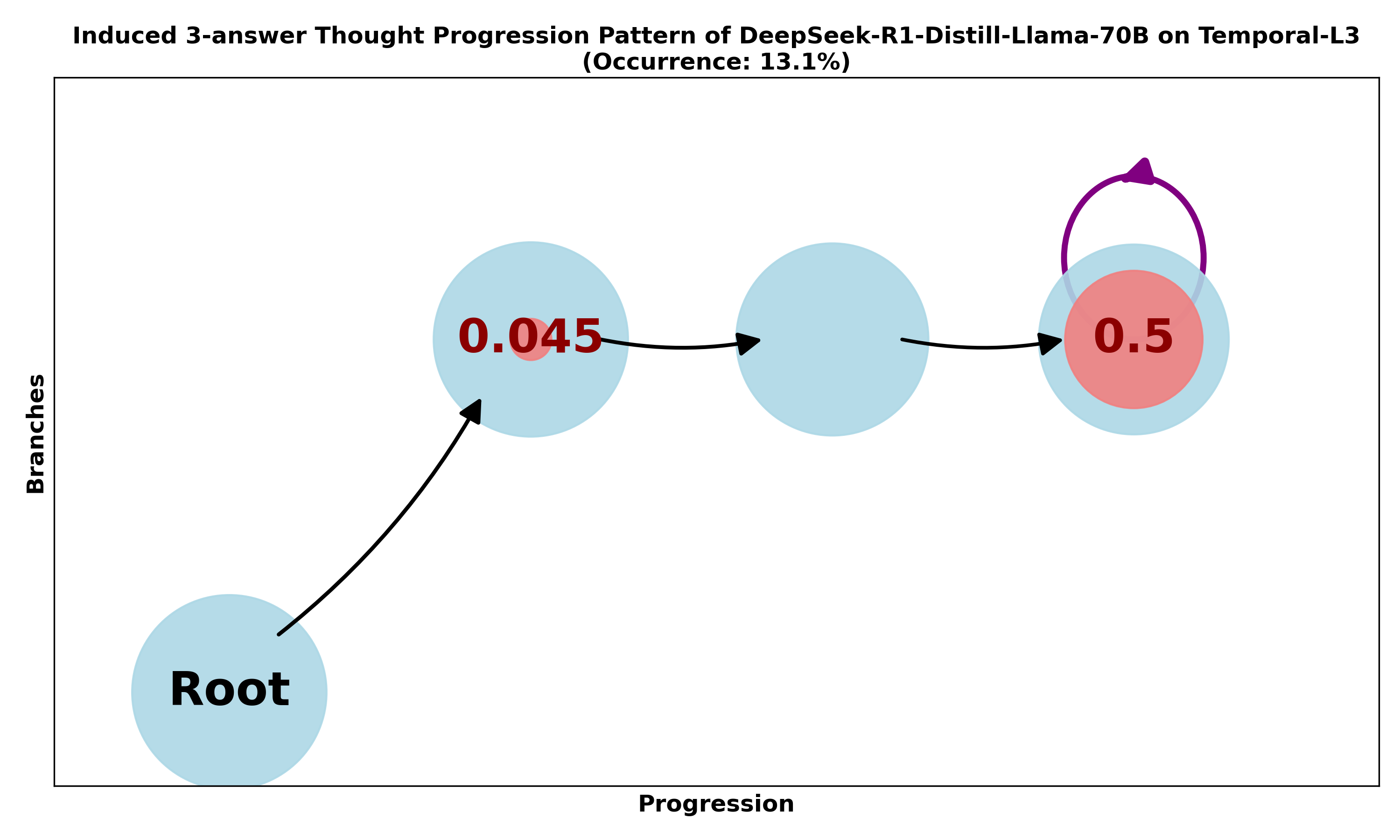}
    \caption{The \textbf{Late Landing} thought progression pattern (3 distinct answer case).}
    \label{fig:late_lander_pattern_group3_70b}
\end{figure}

\begin{figure}[t]
    \centering
    \includegraphics[width=0.5\textwidth]{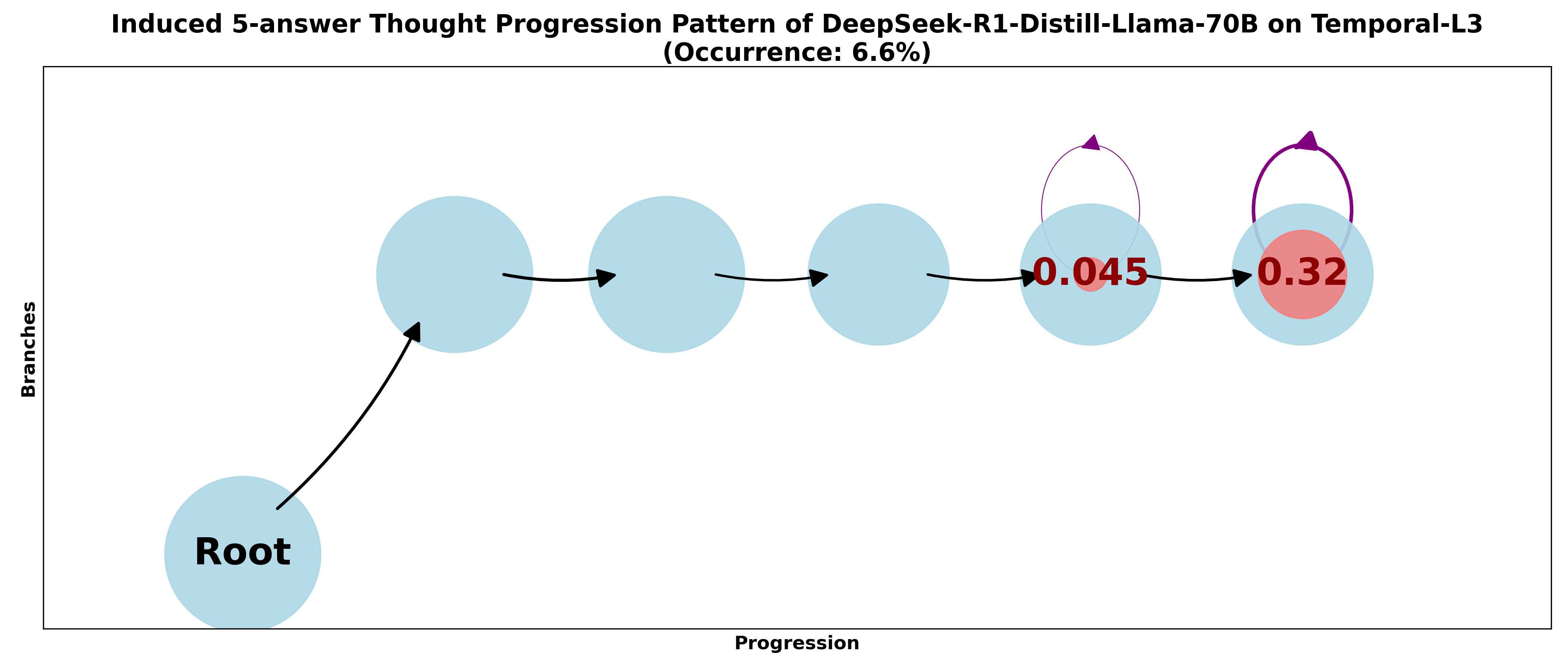}
    \caption{The \textbf{Late Landing} thought progression pattern (5 distinct answer case).}
    \label{fig:late_lander_pattern_group5_70b}
\end{figure}

\Cref{fig:late_lander_pattern_group1_gsm8k} (trivial case),~\ref{fig:late_lander_pattern_group3_gsm8k},~\ref{fig:late_lander_pattern_group3},~\ref{fig:late_lander_pattern_group3_70b} and~\ref{fig:late_lander_pattern_group5_70b} present more plots of the \textbf{Late Landing} thought progression pattern. 

\subsection{Comparison of Induced Thought Progression Patterns with Different Distinct Answers}
\label{appx:pattern_diff_answers}
\begin{figure}[t]
    \centering
    \includegraphics[width=0.5\textwidth]{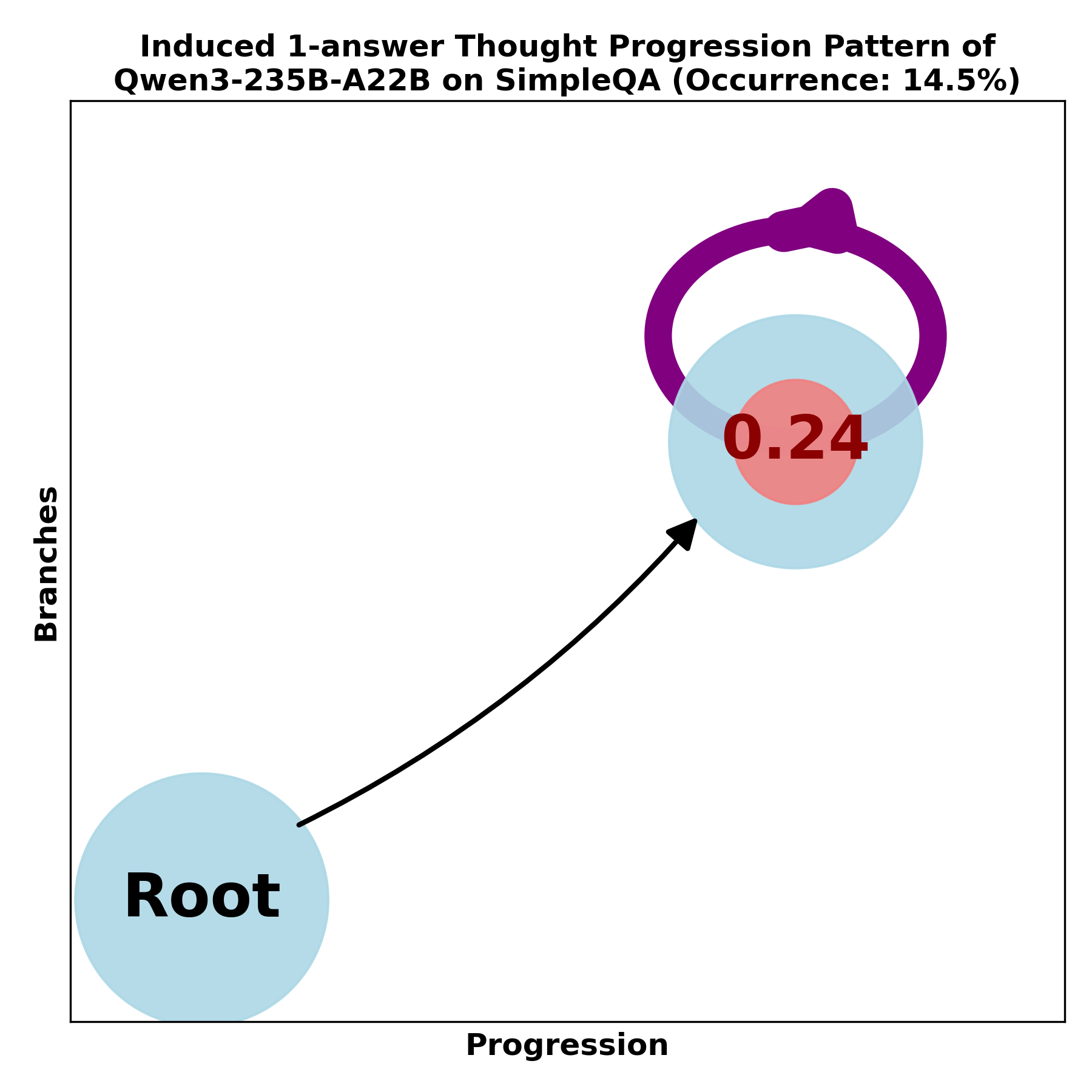}
    \caption{The \textbf{Explorer} thought progression pattern (1 distinct answer case) on the SimpleQA dataset which requires minimal reasoning efforts.}
    \label{fig:explorer_pattern_diff_group1}
\end{figure}

\begin{figure}[t]
    \centering
    \includegraphics[width=0.5\textwidth]{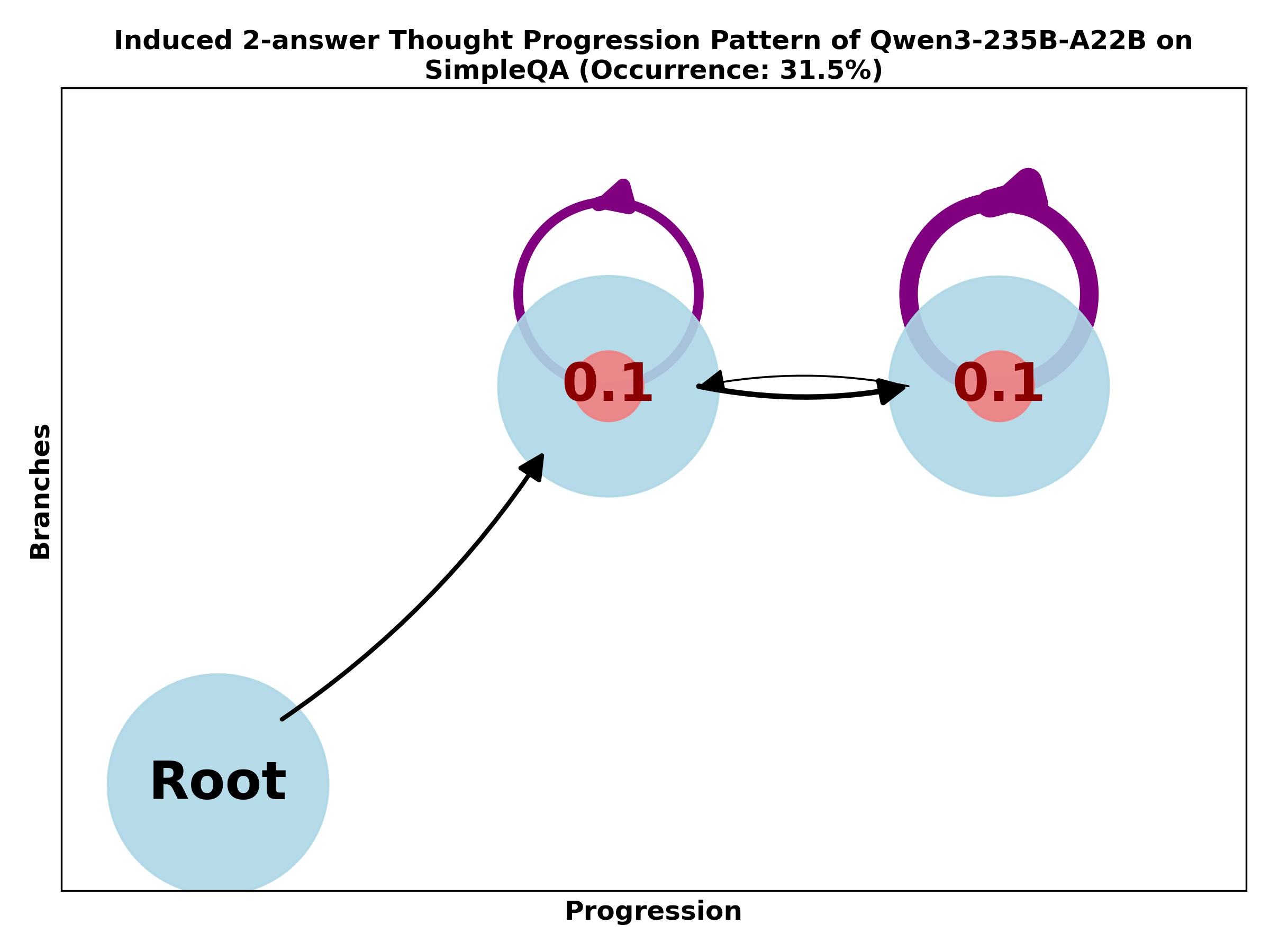}
    \caption{The \textbf{Explorer} thought progression pattern (2 distinct answer case) on the SimpleQA dataset which requires minimal reasoning efforts.}
    \label{fig:explorer_pattern_diff_group2}
\end{figure}

\begin{figure}[t]
    \centering
    \includegraphics[width=0.5\textwidth]{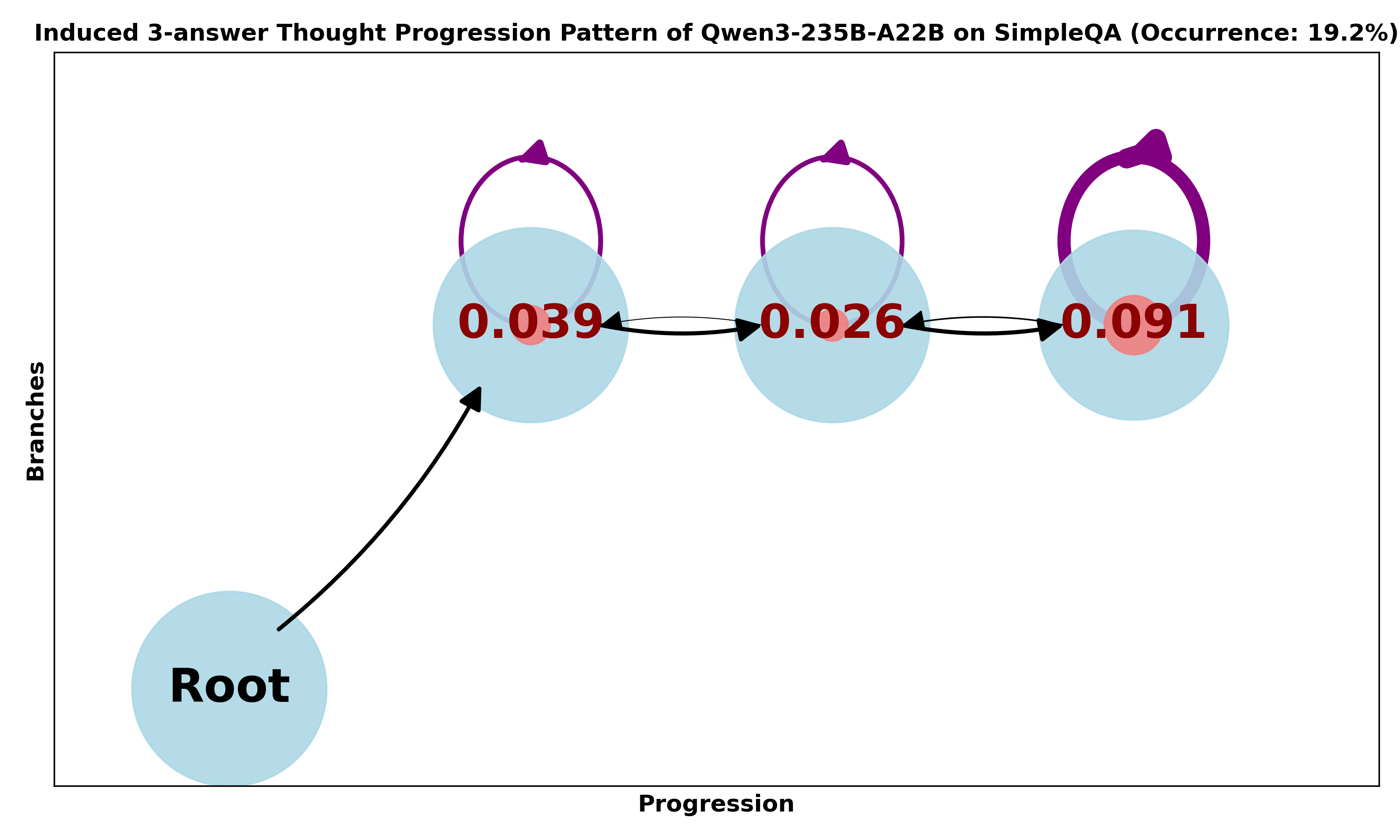}
    \caption{The \textbf{Explorer} thought progression pattern (3 distinct answer case) on the SimpleQA dataset which requires minimal reasoning efforts.}
    \label{fig:explorer_pattern_diff_group3}
\end{figure}

\begin{figure}[t]
    \centering
    \includegraphics[width=0.5\textwidth]{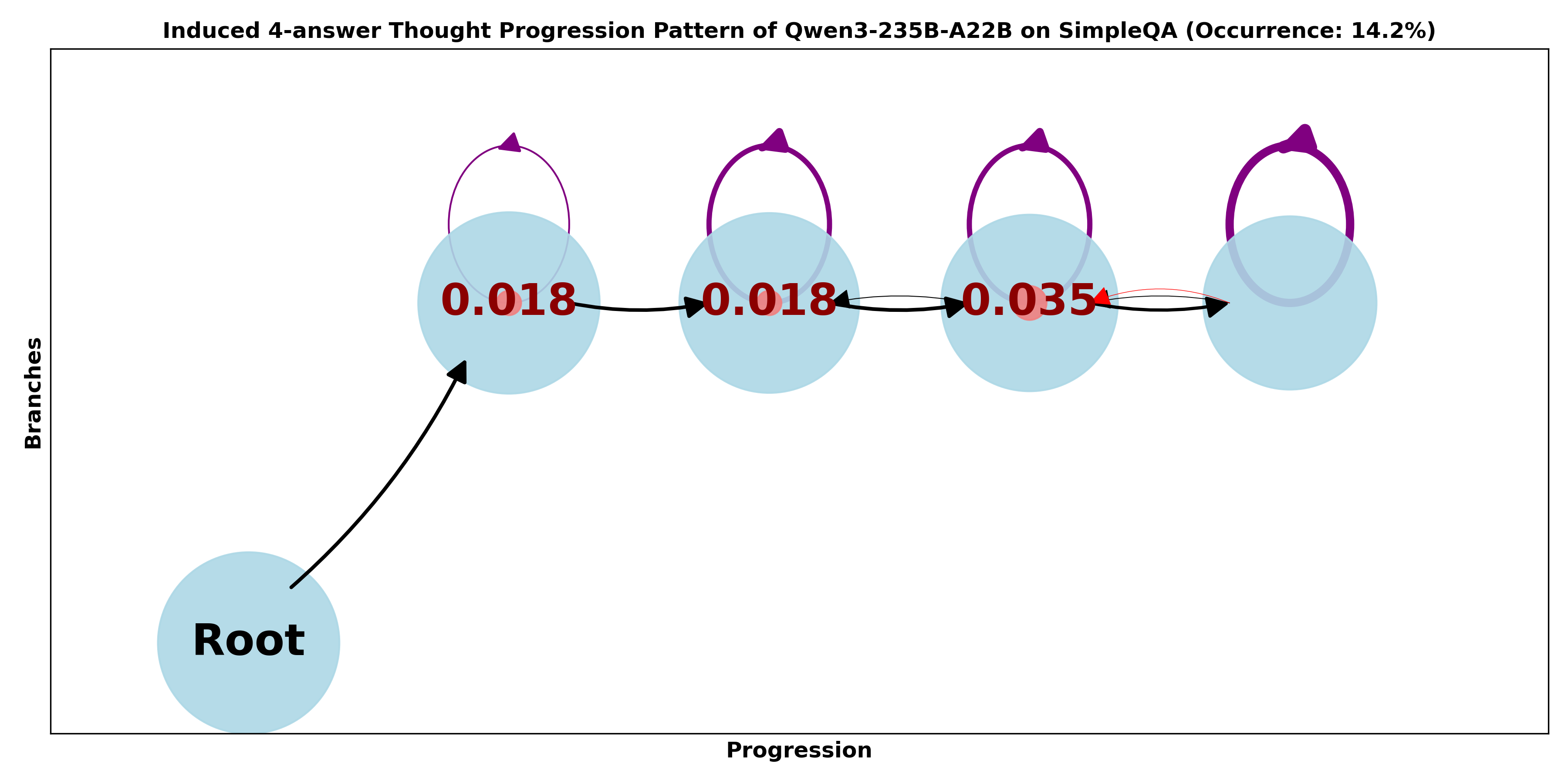}
    \caption{The \textbf{Explorer} thought progression pattern (4 distinct answer case) on the SimpleQA dataset which requires minimal reasoning efforts.}
    \label{fig:explorer_pattern_diff_group4}
\end{figure}

\Cref{fig:explorer_pattern_diff_group1} to \ref{fig:explorer_pattern_diff_group4} show how the progression patterns change with respect to the increasing number of distinct answers. The selected task is \textit{SimpleQA} which is factuality QA task with minimal reasoning efforts. Therefore, the extended reasoning is futile, and the model is trapped by over-verification and frequent revisiting of a previous answer (\textit{marked by the extremely think back edge in \Cref{fig:explorer_pattern_diff_group2} and \Cref{fig:explorer_pattern_diff_group3}}).

\clearpage

\begin{figure*}[t]
\begin{tcolorbox}[colback=white, colframe=black, fonttitle=\bfseries, title=Instructions for Sub-Thought Decomposition and Label Inference (part 1)]
Your goal is to meticulously analyze a machine's thought process, breaking it down into individual sub-thoughts to understand its reasoning flow. This detailed breakdown helps in evaluating and improving AI transparency and logic.

Here's the input format you'll receive:
\begin{verbatim}
Context: {context}
Question: {question}
Gold target: {target}
Thought process: {thought_process}
\end{verbatim}

\hrulefill

\subsubsection*{Your Task: Sub-Thought Decomposition}
\begin{enumerate}
    \item \textbf{Decompose the Thought Process:} Break down the entire machine-generated \texttt{Thought process} into individual sub-thoughts. Each sub-thought must be:
    \begin{itemize}
        \item \textbf{Self-contained:} It must make sense on its own, without heavy reliance on other sub-thoughts.
        \item \textbf{Complete:} A sub-thought must represent a \textbf{complete} sequence of logical steps, generally consiting of intent, action/calculation, and conclusion in that order.
        \item \textbf{Answer-bearing:} A proposed answer must be \textbf{clearly} visible in each sub-thought, which should be in the same format as the \texttt{Gold target}.
    \end{itemize}
    Determine the boundary of each sub-thought by locating \textbf{pivoting phrases}. These are linguistic cues that indicate a \textit{shift} in the thought process. More importantly, pivoting phrases is \textit{always} at the \textbf{beginning} of a sub-thought. \\
    Common pivoting phrases include but are not limited to:
    \begin{itemize}
        \item Wait, Alternatively, Another angle, Another approach, But wait, However, Hold on, Let me double-check, On the other hand
    \end{itemize}
    
    \textbf{Crucial Rule 1: Avoid Isolating Pivoting Phrases} \\
    A pivoting phrase \textbf{cannot be a sub-thought by itself.} It is a signal of a new thought, not the thought itself. The sub-thought \textbf{must include the action or reasoning} that the pivot introduces.
    
    \textbf{Crucial Rule 2: Avoid Fragmented Intents} \\
    A statement of intent (e.g., "I will now calculate...") and its immediate execution must be treated as a single, indivisible sub-thought. The intent and the action are two parts of the same logical step and should not be split.
    
    \textbf{Crucial Rule 3: Avoid Fragmented Conclusions} \\
    A sub-thought must represent a complete logical step. A concluding phrase (e.g., "So, the answer is...", "Therefore, the result is...") should \textbf{NOT} be separated from the immediately preceding calculation, reasoning, or verification that produced it. The reasoning and its conclusion are part of the \textbf{same} sub-thought.

    \item \textbf{Analyze Each Sub-Thought:} For every sub-thought you identify, provide:
    \begin{itemize}
        \item A \textbf{short description} summarizing how the sub-thought approaches the question.
        \item Its \textbf{discourse attribute}, classifying its role in the overall thought process.
    \end{itemize}
\end{enumerate}
\end{tcolorbox}
\caption{Instructions for Sub-Thought Decomposition and Label Inference (part 1). Part 2 is shown in~\Cref{fig:prompt_p2}.}
\label{fig:prompt_p1}
\end{figure*}

\begin{figure*}[t]
\begin{tcolorbox}[colback=white, colframe=black, fonttitle=\bfseries, 
title=Instructions for Sub-Thought Decomposition and Label Inference (part 2)]
\subsubsection*{Discourse Attributes Defined:}
\begin{itemize}[leftmargin=*, nosep]
    \item \textbf{initial:} The very first attempt of solving the question. The initial sub-thought might include an outline of the overall plan. There is only one initial sub-thought.
    \item \textbf{verification:} Verify and confirms the correctness of the \textit{immediately preceding} sub-thought. No changes or corrections are made.
    \item \textbf{correction:} Modifies or rectifies the \textit{immediately preceding} sub-thought.
    \item \textbf{backtrack:} Abandons the current line of reasoning and reverts to an earlier (\textit{NOT} immediately preceding) sub-thought to pursue a different path from that point.
    \item \textbf{sidetrack:} Explores new, supplementary information that isn't directly related to a different approach to the main question.
    \item \textbf{branching out:} Explores alternative ways/methods to answer the user's question or different aspects of it, as opposed to simply adding new information (sidetrack).
    \item \textbf{final:} The final sub-thought that provides the final proposed answer.
\end{itemize}
\bigskip
\textbf{Note:} For attributes that inherently relate to a specific previous sub-thought, use the following format: \texttt{'[Attribute Type]->[Target Index]'}.
\begin{itemize}[nosep]
    \item This \textit{ONLY} applies to \textbf{\texttt{verification}}, \textbf{\texttt{correction}}, and \textbf{\texttt{backtrack}}
    \item The \texttt{[Target Index]} is the 1-indexed position of the sub-thought being referred to.
    \item For \texttt{verification} and \texttt{correction}, the target will \textit{ALWAYS} be the immediately preceding sub-thought (e.g., a \texttt{correction} at index \textbf{4} would be formatted as \texttt{'correction->3'}).
    \item For \texttt{backtrack}, the target can be any earlier, \textit{NON-PRECEDING} sub-thought (e.g., a \texttt{backtrack} at index \textbf{5} that reverts to the thought at index \textbf{1} would be formatted as \texttt{'backtrack->1'}). If there is a "start over" action, mark it as \texttt{'backtrack->ROOT'}.
\end{itemize}

\hrulefill

\subsubsection*{Expected Output Format:}
Your analysis output must be a JSON array structured as follows:
\begin{verbatim}
[
    {{
      "sub_thought_index": "The 0-indexed position of the sub-thought",
      "sub_thought_description": "A concise summary of what this 
      sub-thought is about",
      "sub_thought_answer": "The machine-generated answer extracted  
      from this *CURRENT* sub-thought. It should be in the same format  
      as the ``Gold target''",
      "discourse_attribute": "The classified role of this sub-thought  
      in the entire thought process. Each sub-thought can only have 
      one discourse attribute. Use your best judgement to 
      determine the most appropriate attribute.",
      "sub_thought_content": "The exact text segment from the original 
      thought process that constitutes this sub-thought"
    }}
]
\end{verbatim}
\end{tcolorbox}
\caption{Instructions for Sub-Thought Decomposition and Label Inference (part 2).}
\label{fig:prompt_p2}
\end{figure*}

% ================= Figure 1 =================
\begin{figure*}[t]
\begin{tcolorbox}[colback=white, colframe=black, fonttitle=\bfseries, title=Sample Parsed Output by \framework (Sub-thoughts 0-1)]

% ---------- Sub-thought 0 ----------
\begin{tcolorbox}[colback=white!98!gray, colframe=black!50, title=Sub-thought 0]
\texttt{Content}: Okay, let's see. Doctor Jones has a total of nine hours at the clinic on Monday. He needs to spend time on rounds for inpatients and ten appointments. Then he has to update his records. The question is asking how many hours he has left for updating records.\\

First, I need to calculate the total time he spends on rounds and appointments. Then subtract that from the nine hours to find the remaining time for records.\\

Let me start with the rounds. He has 9 inpatients, and each takes 20 minutes. So 9 times 20 minutes. Let me compute that: 9 * 20 = 180 minutes.\\

Then the appointments. He has ten appointments, each 30 minutes. So 10 * 30 = 300 minutes.\\

Total time spent on both rounds and appointments is 180 + 300 minutes. That's 480 minutes. Now, convert that to hours because the total time he has is in hours. Since 60 minutes = 1 hour, divide 480 by 60. 480 ÷ 60 = 8 hours.\\

Wait, so he spends 8 hours on those two tasks. He has nine hours total. So 9 - 8 = 1 hour left for updating records. That seems straightforward. \\

\texttt{Discourse Attribute}: \texttt{initial} \\
\texttt{Sub-thought Answer}: $\boxed{1}$ \\
\texttt{Generated Answer}: $\boxed{1}$
\end{tcolorbox}

% ---------- Separator ----------
\vspace{0.3em}
\noindent\hrulefill
\vspace{0.3em}

% ---------- Sub-thought 1 ----------
\begin{tcolorbox}[colback=white!98!gray, colframe=black!50, title=Sub-thought 1]
\texttt{Content}: Wait, but let me double-check. Maybe I made a mistake in the calculation. Let me go through again.\\

Inpatient rounds: 9 patients * 20 minutes each. 9*20 is 180. Correct. Appointments: 10 * 30 = 300. Correct. Total 180 + 300 = 480. 480 divided by 60 is 8. So yes, 8 hours. Subtract from 9 hours, 1 hour left. So the answer should be 1 hour. \\

\texttt{Discourse Attribute}: \texttt{verification} \\
\texttt{Sub-thought Answer (Correctness)}: $\boxed{1}$ \\
\texttt{Answer in Response (Helpfulness)}: $\boxed{1}$
\end{tcolorbox}

\end{tcolorbox}
\caption{Sample parsed output (sub-thoughts 0-1) by \framework on the \texttt{\texttt{Qwen3-235B-A22B}}-generated response to a GSM8K prompt. The prompt is ``Doctor Jones is scheduling his time for Monday. He is spending nine hours at the clinic where he works that day. He has to do rounds to check on inpatients staying at the clinic, which takes twenty minutes per inpatient, and he has ten appointments, which take thirty minutes each. How many hours will Doctor Jones have left to update his records if he has 9 inpatients at the clinic?''. The ground-truth answer is 1. The constructed thought progression graph for this thought process is shown in~\Cref{fig:progression_12}.}
\label{fig:gsm8k_parsed_example_part1}
\end{figure*}

% ================= Figure 2 =================

\end{document}